\documentclass[journal]{IEEEtran}

\usepackage{amsmath}
\usepackage{graphicx}
\usepackage{hyperref}
\usepackage{comment}
\usepackage{subfigure}
\usepackage[percent]{overpic}
\usepackage{multirow}
\usepackage{booktabs}
\usepackage{color}
\usepackage{amssymb}
\usepackage{breqn}
\usepackage{enumitem}
\usepackage{lscape}
\usepackage[para]{threeparttable}

\begin{document}

\title{Temporally Consistent Depth Prediction with Flow-Guided Memory Units}

\author{Chanho~Eom, Hyunjong~Park, and~Bumsub~Ham,~\IEEEmembership{Member,~IEEE}
\thanks{This work was supported by Institute for Information \& communications Technology Promotion~(IITP) grant funded by the Korea government~(MSIP) (No.2016-0- 00197, Development of the high-precision natural 3D view generation technology using smart-car multi sensors and deep learning). \newline\indent C. Eom, H. Park, B. Ham are with the School of Electrical and Electronic Engineering, Yonsei University, Seoul 03722, South Korea (e-mail: cheom@yonsei.ac.kr; hyunpark@yonsei.ac.kr; bumsub.ham@yonsei.ac.kr). Corresponding author: Bumsub Ham.}}

\maketitle

\begin{abstract}
Predicting depth from a monocular video sequence is an important task for autonomous driving. Although it has advanced considerably in the past few years, recent methods based on convolutional neural networks~(CNNs) discard temporal coherence in the video sequence and estimate depth independently for each frame, which often leads to undesired inconsistent results over time. To address this problem, we propose to memorize temporal consistency in the video sequence, and leverage it for the task of depth prediction. To this end, we introduce a two-stream CNN with a flow-guided memory module, where each stream encodes visual and temporal features, respectively. The memory module, implemented using convolutional gated recurrent units~(ConvGRUs), inputs visual and temporal features sequentially together with optical flow tailored to our task. It memorizes trajectories of individual features selectively and propagates spatial information over time, enforcing a long-term temporal consistency to prediction results. We evaluate our method on the KITTI benchmark dataset in terms of depth prediction accuracy, temporal consistency and runtime, and achieve a new state of the art. We also provide an extensive experimental analysis, clearly demonstrating the effectiveness of our approach to memorizing temporal consistency for depth prediction.
\end{abstract}


\IEEEpeerreviewmaketitle

\section{ Introduction }
\IEEEPARstart {D}{epth} prediction from images plays a significant role in autonomous driving and advanced driver assistance systems, which helps understanding a geometric layout in a scene, and can be leveraged to solve other tasks, including  vehicle/pedestrian detection~\cite{chen2017coherent,keller2011benefits}, traffic scene segmentation~\cite{li2018traffic}, and 3D reconstruction~\cite{wu2017geometry}. Stereo matching is a typical approach to recovering depth that finds dense correspondences between a pair of stereo images~\cite{nguyen2017robust,van2006real,li2008binocular}. Stereo matching methods compute similarities between local patches~\cite{muresan2017mutlipatch} or optimize global objective functions to consider smoothness priors penalizing large derivatives of depth~\cite{hirschmuller2007stereo,miclea2018real,spangenberg2014large}. These approaches show state-of-the-art performance, but capturing pairs of stereo images requires multiple cameras calibrated, making it difficult to apply them in practice. An alternative is to predict depth from a monocular video sequence, and it is of great interests in recent years~\cite{eigen2014depth,liu2014discrete,zhou2017unsupervised,wang2018learning,godard2017unsupervised,kuznietsov2017semi,cs2018depthnet,fu2018deep}. This approach builds upon the insight that human can perceive depth using monocular depth cues~(\emph{e.g.}, occlusion, perspective, motion parallax) only~\cite{rogers1979motion}. Eigen~\emph{et al.}~\cite{eigen2014depth} first propose a supervised learning method for predicting depth from a single still image using CNNs. Zhou \emph{et al.}~\cite{zhou2017unsupervised} and Wang \emph{et al.}~\cite{wang2018learning} recently propose CNN architectures for predicting depth from a monocular video, where two networks are trained separately to estimate depth and camera pose. These methods are limited in that they predict depth independently for each frame, discarding temporal coherence in the video sequence. That is, they give temporally inconsistent results, causing serious temporal flickering artifacts. Recurrent neural networks~(RNNs) have been widely used to model temporal dependency across sequential data~(\emph{e.g.},~video and text), and they have shown the effectiveness in various applications including action recognition~\cite{du2018recurrent} and machine translation~\cite{zhang2017context}. They, however, still show a limited capability of handing the flickering artifacts~\cite{cs2018depthnet,shi2017deep}.

\begin{figure}[t]
  \centering
  \renewcommand*{\thesubfigure}{}
  \subfigure[(a) Depth at time $t$]{
    \begin{minipage}[t]{0.457\linewidth}
      \begin{overpic}[width=\linewidth]{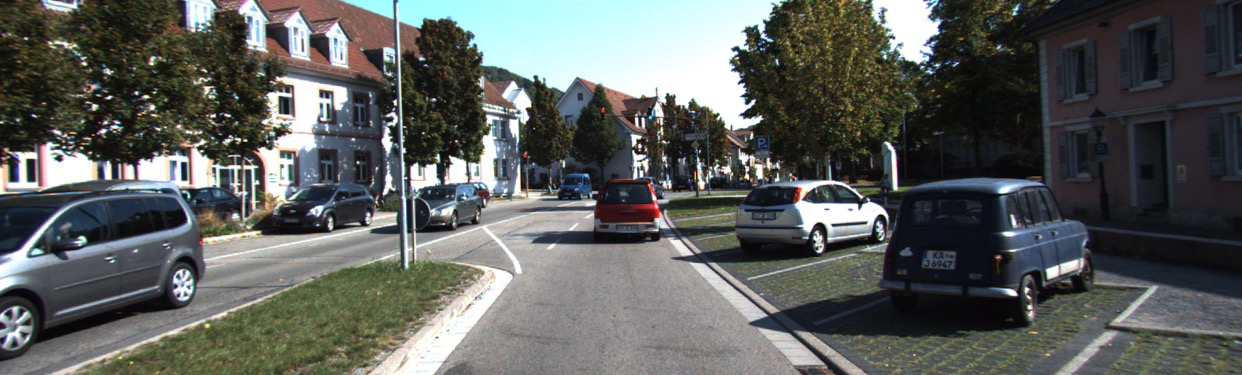}
      \end{overpic}\vspace{0.073cm}
      \begin{overpic}[width=\linewidth]{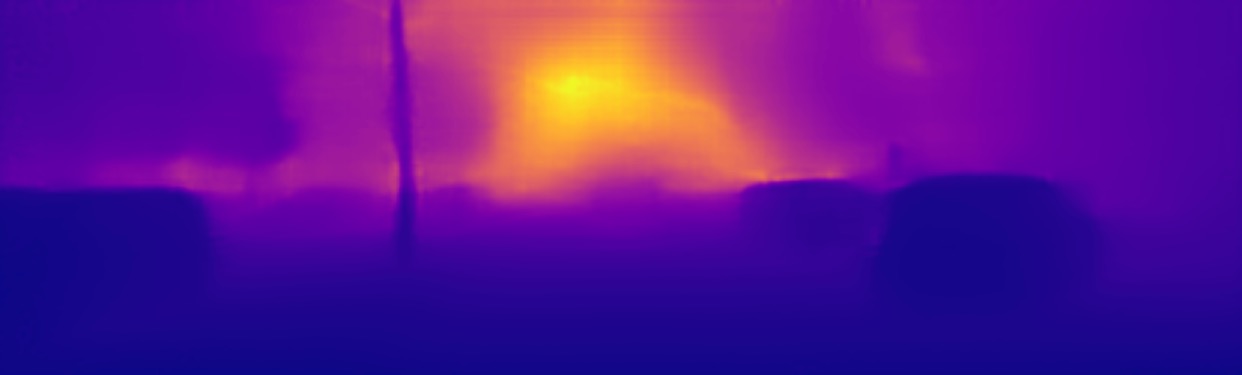}
      \put(2,24){\color{white}{\scriptsize Wang~\emph{et al.}}}
      \end{overpic}\vspace{0.073cm}
      \begin{overpic}[width=\linewidth]{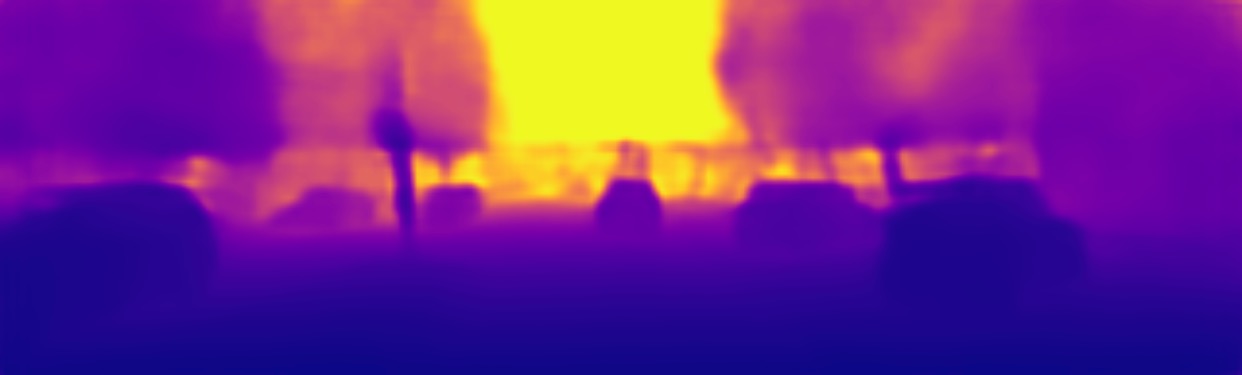}
      \put(2,24){\color{white}{\scriptsize Kuznietsov~\emph{et al.}}}
      \end{overpic}\vspace{0.073cm}
      \begin{overpic}[width=\linewidth]{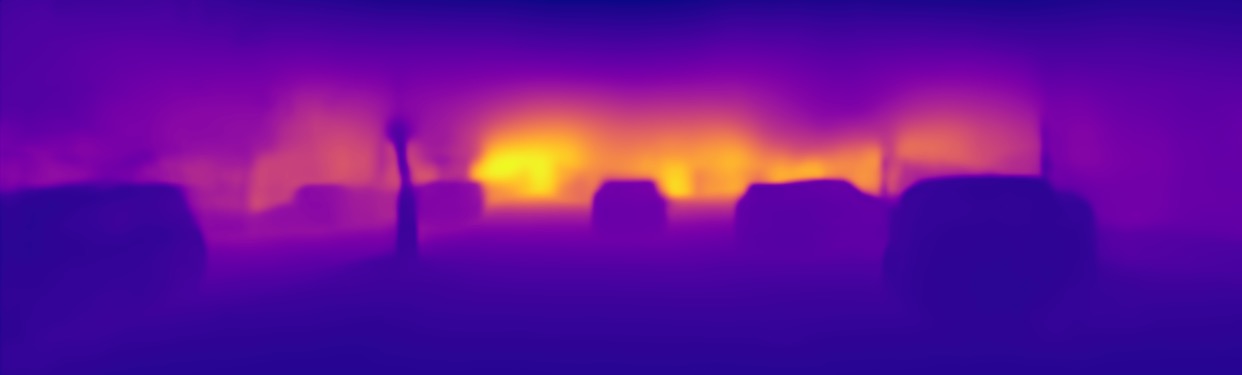}
      \put(2,24){\color{white}{\scriptsize Ours}}
      \end{overpic}\vspace{0.073cm}
      \begin{overpic}[width=\linewidth]{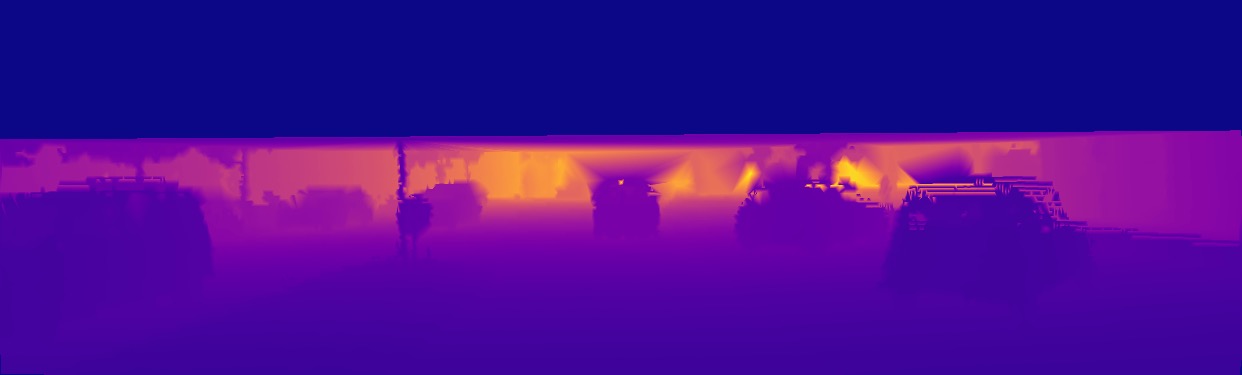}
      \put(2,24){\color{white}{\scriptsize Ground truth}}
      \end{overpic}\vspace{0.173cm}
    \end{minipage}
    }\hspace{-0.270cm}
  \subfigure[(b) Depth at time $t+1$]{
    \begin{minipage}[t]{0.457\linewidth}
      \begin{overpic}[width=\linewidth]{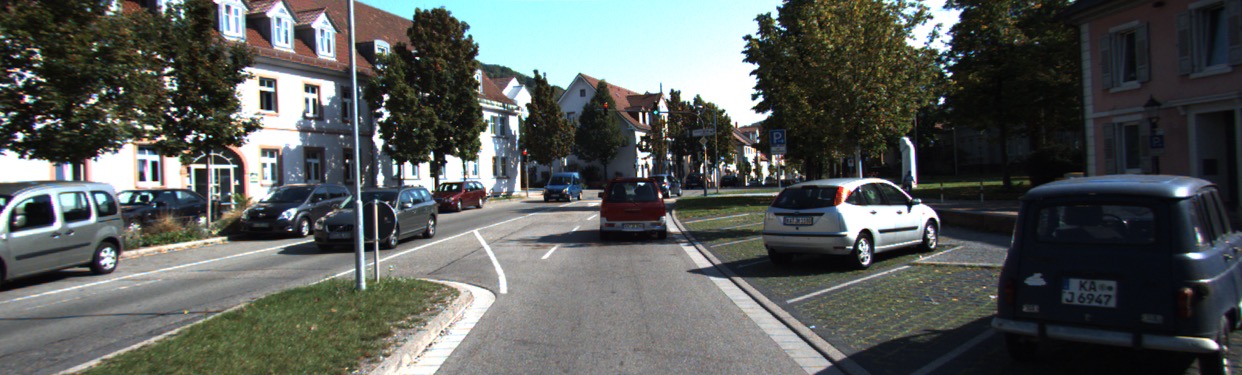}
      \put(2,24){\color{white}{}}
      \end{overpic}\vspace{0.073cm}
      \begin{overpic}[width=\linewidth]{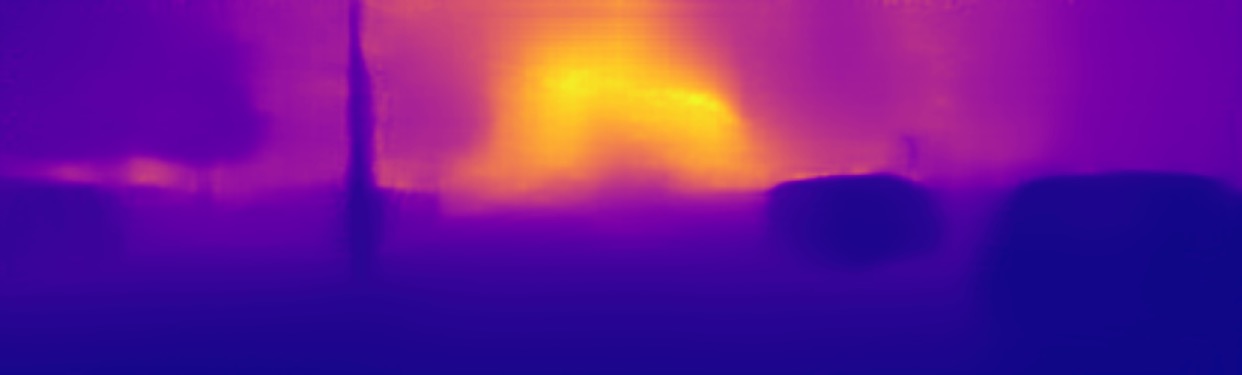}
      \put(2,24){\color{white}{}}
      \end{overpic}\vspace{0.073cm}
      \begin{overpic}[width=\linewidth]{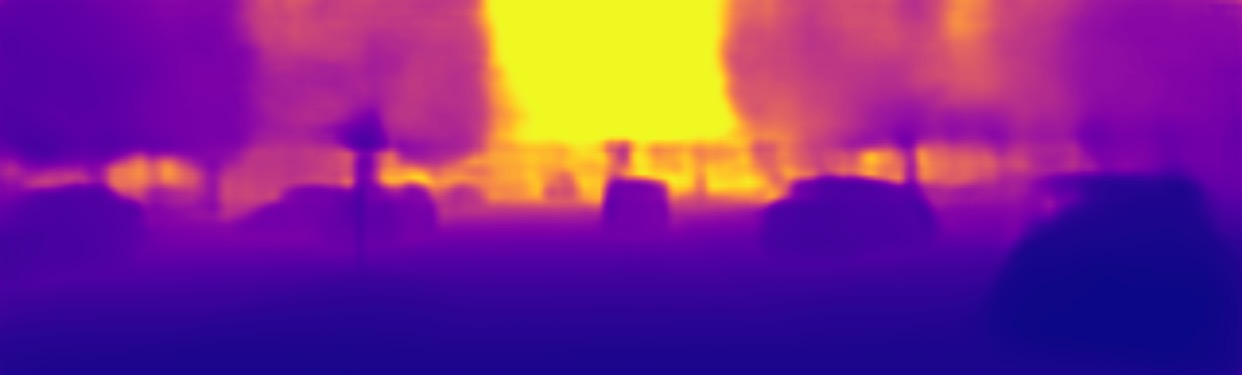}
      \put(2,24){\color{white}{}}
      \end{overpic}\vspace{0.073cm}
      \begin{overpic}[width=\linewidth]{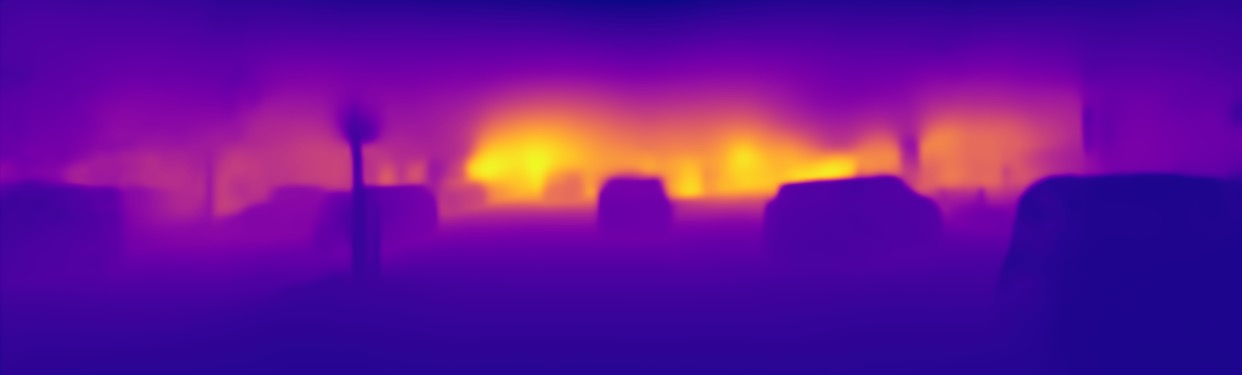}
      \put(2,24){\color{white}{}}
      \end{overpic}\vspace{0.073cm}
      \begin{overpic}[width=\linewidth]{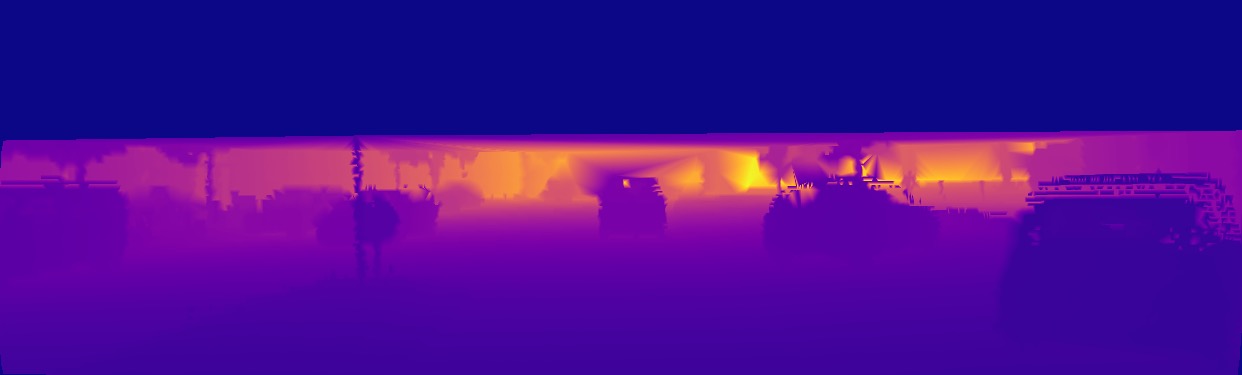}
      \put(2,24){\color{white}{}}
      \end{overpic}\vspace{0.173cm}
    \end{minipage}
    }\hspace{-0.3cm}
  \subfigure[(c) Temporal differences]{
    \begin{minipage}[t]{0.93\linewidth}
      \begin{overpic}[width=\linewidth]{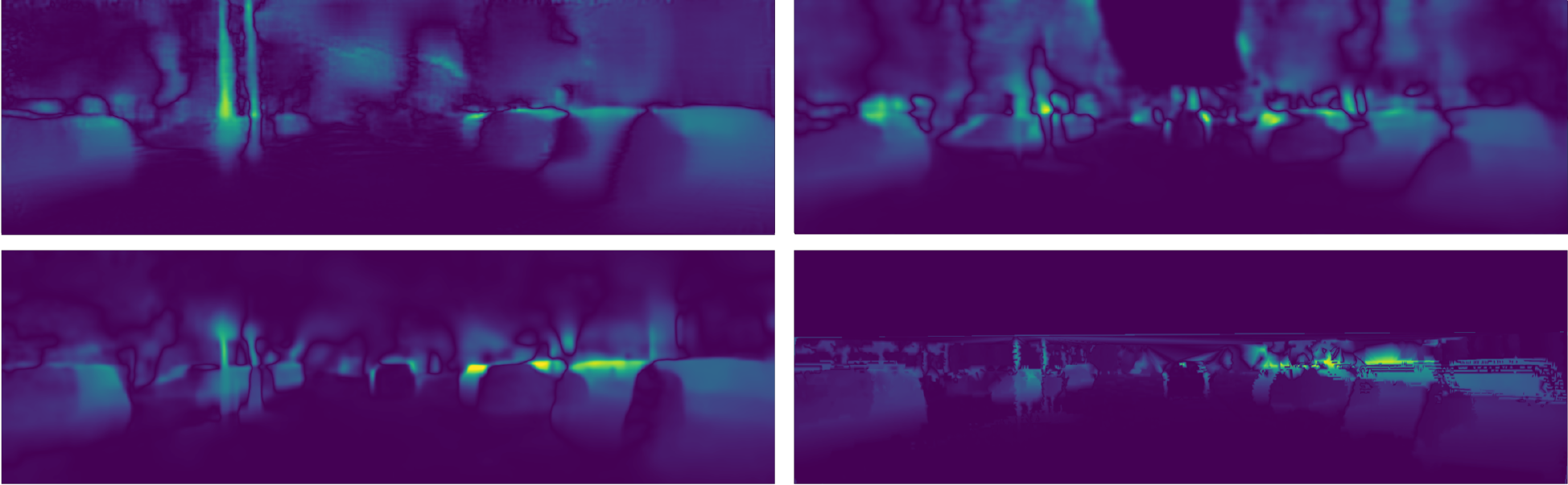}\vspace{0.065cm}
      \put(1,28){\color{white}{\scriptsize Wang~\emph{et al.}}}
      \put(52,28){\color{white}{\scriptsize Kuznietsov~\emph{et al.}}}
      \put(1,12){\color{white}{\scriptsize Ours}}
      \put(52,12){\color{white}{\scriptsize Ground truth}}
      \end{overpic}

    \end{minipage}
    }
\vspace{-0.2cm}
  \caption{Visual comparison of the state of the art and our model on monocular depth prediction. (a-b) Top to bottom: Video frames, depth maps obtained by Wang~\emph{et al.}~\cite{wang2018learning}, Kuznietsov~\emph{et al.}~\cite{kuznietsov2017semi} and our model, and ground-truth depth at time~$t$ and $t+1$, respectively. (c) Absolute differences between the depth maps at time~$t$ and $t+1$. Compared to other methods, our model gives a temporally consistent result similar to ground truth while providing a sharp depth transition (yellow: high, blue: low). (Best viewed in color.)}
  \vspace{-0.2cm}  
  \label{fig:teaser}
\end{figure}

In this paper, we present a simple yet effective method for a temporally consistent depth prediction from a monocular video sequence~(Fig.~\ref{fig:teaser}). We transfer temporal consistency in the video to RNNs explicitly, particularly using convolutional gated recurrent units~(ConvGRUs)~\cite{ballas2015delving}. To implement this idea, we propose a flow-guided memory unit using optical flow specific to our task, maintaining a long-term temporal consistency in depth prediction results. Our module uses spatial and temporal features extracted by a two-stream CNN. We have two main reasons for decoupling these features.  First, it has been proven that learning spatiotemporal features jointly from a stack of frames does not capture the motion well~\cite{simonyan2014two}. Second, optical flow itself provides an important clue for motion parallax, which is helpful to infer depth from a monocular video sequence. For example, objects closer to a camera move faster than distant ones. We show that our method outperforms the state of the art in terms of temporal consistency, and shows a good trade-off between depth prediction accuracy and runtime. The main contributions of this paper can be summarized as follows:

\begin{itemize}[leftmargin=*]
\item We present an effective ConvGRU encoder-decoder module for a temporally consistent depth prediction from a monocular video sequence. To our knowledge, this is the first approach based on convolutional/recurrent networks to considering temporal consistency in depth prediction.
\item We propose a flow-guided memory unit that retains a long-term temporal consistency explicitly for individual pixels.
\item We present state-of-the-art results on the KITTI~\cite{geiger2013vision} benchmark. We additionally provide an extensive experimental analysis, clearly demonstrating the effectiveness of our approach to memorizing temporal consistency for depth prediction.
\end{itemize}

To encourage comparison and future work, we release our code and models online: \url{https://cvlab-yonsei.github.io/projects/FlowGRU/}. 

\section{ Related work }
  In this section, we briefly review representative works related to ours.
\vspace{-0.3cm}

  \subsection{Monocular depth prediction}
The problem of predicting depth from monocular images or video sequences has significant attention in recent years. Early works exploit hand-crafted features such as SIFT~\cite{lowe2004distinctive} and HOG~\cite{dalal2005histograms} together with graphical models~\cite{saxena2006learning,liu2014discrete} or nonparametric sampling~\cite{karsch2014depth}. Saxena~\emph{et al}.~\cite{saxena2006learning} estimate depth from monocular images using Markov random fields~(MRFs) where they incorporate multi-scale features. Liu~\emph{et al}.~\cite{liu2014discrete} extend this idea by using a discrete-continuous graphic model. Karsch~\emph{et al.}~\cite{karsch2014depth} introduce a nonparametric approach to depth prediction from monocular images and videos. They transfer depth labels from a large-scale RGB/D dataset using dense correspondences established by a SIFT flow method~\cite{liu2011sift}. CNNs have allowed remarkable advances in depth prediction in the past few years. Eigen \emph{et al.}~\cite{eigen2014depth} first leverage CNNs to predict depth from monocular images in a coarse-to-fine manner. In particular, they introduce a scale-invariant loss function that alleviates ambiguity in scale. Liu \emph{et al.}~\cite{liu2015deep} combine CNNs with conditional random fields~(CRFs) for structured prediction. Kuznietsov~\emph{et al.}~\cite{kuznietsov2017semi} propose to use additional stereo images at training time. They predict depth from left images and synthesize novel views by warping right ones using estimated depth. The differences between left and synthesized images are then used as a supervisory signal for training. Unlike the aforementioned methods using monocular images, recent works~\cite{zhou2017unsupervised,wang2018learning,yin2018geonet} have shown success in learning depth from a monocular video sequence. Zhou~\emph{et al.}~\cite{zhou2017unsupervised} present an approach to estimating  depth and camera pose simultaneously from a video sequence. Similar to~\cite{kuznietsov2017semi}, they synthesize adjacent frames using estimated depth and camera pose, and use the discrepancy between the synthesized and original ones as a supervisory signal. This approach, however, requires ground-truth parameters for camera pose. Wang~\emph{et al.}~\cite{wang2018learning} propose an unsupervised learning approach to estimating pose parameters using a differential version of the direct visual odometry~(DVO) method~\cite{steinbrucker2011real}, commonly employed in a SLAM community, and leverage it to depth prediction from a monocular video. Yin~\emph{et al.}~\cite{yin2018geonet} propose an unsupervised learning framework that learns depth, camera pose, and optical flow jointly. They first estimate depth and camera pose to obtain rigid flow, and then use them to compute optical flow. These approaches using CNNs outperform traditional methods by large margins, but none of them consider temporal coherence in a video sequence. They give temporally inconsistent results, resulting in temporal flickering artifacts. On the contrary, our method memorizes temporal coherence in a video sequence, enabling a temporally consistent depth prediction.

\begin{figure*}[!t]
  \centering
  \renewcommand*{\thesubfigure}{}
  \subfigure[]{
    \begin{minipage}[t]{0.9\linewidth}
      \includegraphics[width=\linewidth]{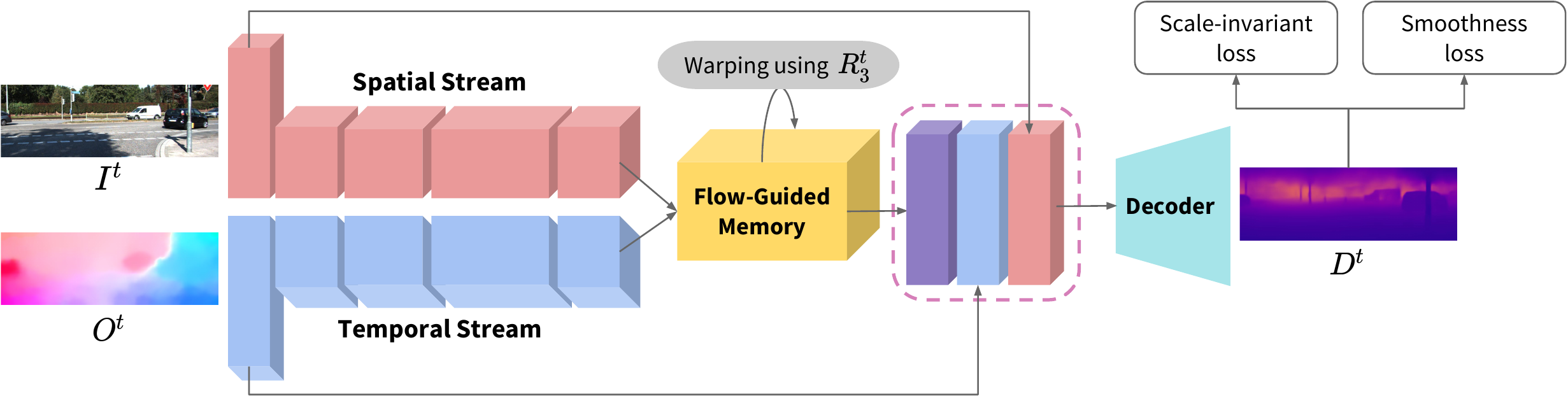}\vspace{0.065cm}
    \end{minipage}
    }
  \caption{Overview of our framework. Our model inputs a video frame~$I^t$ and optical flow~$O^t$ at time~$t$, and extracts spatial and temporal features from each input, respectively, using a two-stream encoder. A flow-guided memory module takes these features concatenated and memorizes temporal coherence in the video using trajectories of individual pixels. Specifically, it aligns hidden states over time using a refined flow~$R_3^t$ specific to our task~(Fig.~\ref{fig:frn}), while partially retaining or filtering out the hidden states in the visual memory. A decoder reconstructs a depth map~$D^t$ at time~$t$ from the output of the memory module together with spatial and temporal features. For depth prediction, we use scale-invariant and smoothness terms, and train the whole network end-to-end. See Table~\ref{tabl:network} for the detailed description of the network structure. (Best viewed in color.)}
\vspace{-0.2cm}
  \label{fig:timestep}
\end{figure*}

\vspace{-0.3cm}
  \subsection{Recurrent models }
  RNNs have been widely used to capture temporal dependency in sequential data~\cite{hopfield1982neural,rumelhart1986learning}. Representative models include GRUs~\cite{cho2014learning} and long short term memory~(LSTM)~\cite{hochreiter1997long}, and they have been adopted successfully to various tasks such as video representation~\cite{srivastava2015unsupervised}, image captioning~\cite{donahue2015long} and car-following modeling~\cite{wang2018capturing}. LSTM and GRUs, typically using fully-connected layers, do not maintain spatial information and require a lot of network parameters. This is problematic especially for high-dimensional data~(\emph{e.g.},~video sequences). ConvLSTM~\cite{xingjian2015convolutional} and ConvGRUs~\cite{ballas2015delving} replace the fully-connected layers with convolutional ones, preserving spatial information while reducing the number of parameters drastically. They have been widely exploited in computer vision and image processing tasks including video recognition~\cite{tokmakov2017learning,ballas2015delving}, depth prediction~\cite{jie2018left,cs2018depthnet}, and precipitation nowcasting~\cite{xingjian2015convolutional,shi2017deep}. Particularly, Shi~\emph{et al.}~\cite{shi2017deep} introduce a variant of ConvGRUs, trajectory GRU~(TrajGRU), and apply it to predict a future rainfall intensity. Similar to the deformable convolutional network~\cite{dai2017deformable}, TrajGRU learns local offsets~(\emph{i.e.}, where to aggregate) and adds them to a regular grid in a standard convolution, which has an effect of using spatially-variant convolutional kernels. In contrast to this, our model memorizes spatiotemporal features aligned along the path guided by optical flow. Most similar to ours is a ConvLSTM-based framework for depth prediction~\cite{cs2018depthnet}. It uses ConvLSTMs to exploit spatiotemporal features from a video sequence, but does not give temporally coherent results. We also use a recurrent network to exploit spatiotemporal features, but a flow-guided memory unit in our model retains a long-term temporal consistency in a video sequence explicitly. Note that the ability to capture temporal dependency in ConvLSTMs or ConvGRUs does not guarantee to obtain temporally consistent results~\cite{shi2017deep}.

  \vspace{-0.3cm}
  \subsection{Temporal coherency }
Recently, several approaches have been introuced to model temporal coherence in a video sequence. They typically use optical flow to smooth results along dense trajectories~\cite{lang2012practical,aydin2014temporally,bonneel2013example}, to construct loss functions penalizing the difference between current and synthesized frames~\cite{chen2017coherent,lai2018learning} or to align features from current and previous ones~\cite{chen2017coherent,gadde2017semantic}. Different from the first and second approaches, we focus on designing a recurrent model itself that transfers temporal coherence in the video sequence to depth prediction results, without using temporal filtering techniques or corresponding loss functions.  Our method is similar to the last approach in that we use aligned features to obtain temporally consistent results. In contrast to~\cite{chen2017coherent,gadde2017semantic}, we use a memory unit that filters out or retains spatiotemporal features aligned along dense trajectories. More recently, V. Miclea~\emph{et al.} propose to exploit temporal cues~\cite{miclea2018real} for depth prediction. They use a previous frame and corresponding depth and segmentation results to refine incorrect depth values. Compared to this work, our method maintains a long-term temporal consistency by using the memory unit.

\section{ Proposed solution }
In this section, we describe a recurrent model with a flow-guided memory module for a temporally consistent depth prediction~(Section~\ref{sec:Network Architecture}). We then present loss functions for learning depth and refined flow~(Section~\ref{sec:Loss Functions}). The entire network is trained end-to-end. 

\vspace{-0.3cm}
\subsection{ Network architecture } \label{sec:Network Architecture}
Our network mainly consists of three parts~(Fig.~\ref{fig:timestep}): A two-stream encoder extracts spatial and temporal features from a video frame~$I^t$ and optical flow~$O^t$ in a backward direction~(\textit{i.e},~a dense flow field from~$I^{t}$ to $I^{t-1}$), respectively, where $t$ represents a time step. A flow-guided memory module inputs both features, and retains parts of them along trajectories of individual pixels to memorize temporal coherence in a video sequence. A decoder takes an output of the flow-guided memory module and outputs a depth map~$D^t$. In the following, we present the detailed description of each part.

\subsubsection{ Two-stream encoder } \label{sec:Two-Stream Encoder}

  A video sequence allows to leverage spatial and temporal information for depth prediction. Motivated by the works~\cite{simonyan2014two,wang2016temporal} for action recognition, we use a two-stream encoder where each stream has the same CNN architecture~(but different parameters), takes the video frame~$I^t$ and optical flow~$O^t$, and then extracts spatial and temporal features, respectively. They are complementary each other. The spatial features capture appearance of objects and scene layout within each frame while the temporal ones encode trajectories of individual pixels~(\emph{i.e.},~motion) across frames. Monocular depth prediction using CNNs typically requires large receptive fields to extract monocular depth cues including motion parallax and perspective~\cite{eigen2014depth}. We can enlarge the receptive fields by using convolutions with multiple strides or pooling methods, but they lead to loss of spatial resolution and scene details such as small and thin structures~\cite{yu2017dilated}. We instead implement the two-stream encoder using a series of dilated convolutions~\cite{yu2015multi} with which we can adjust size of receptive fields by changing a dilation rate without loss of resolution. Note that the dilated version with rate of~1 corresponds to a standard convolution.

	\begin{figure*}[!t]
  \centering
  \includegraphics[width=0.8\linewidth]{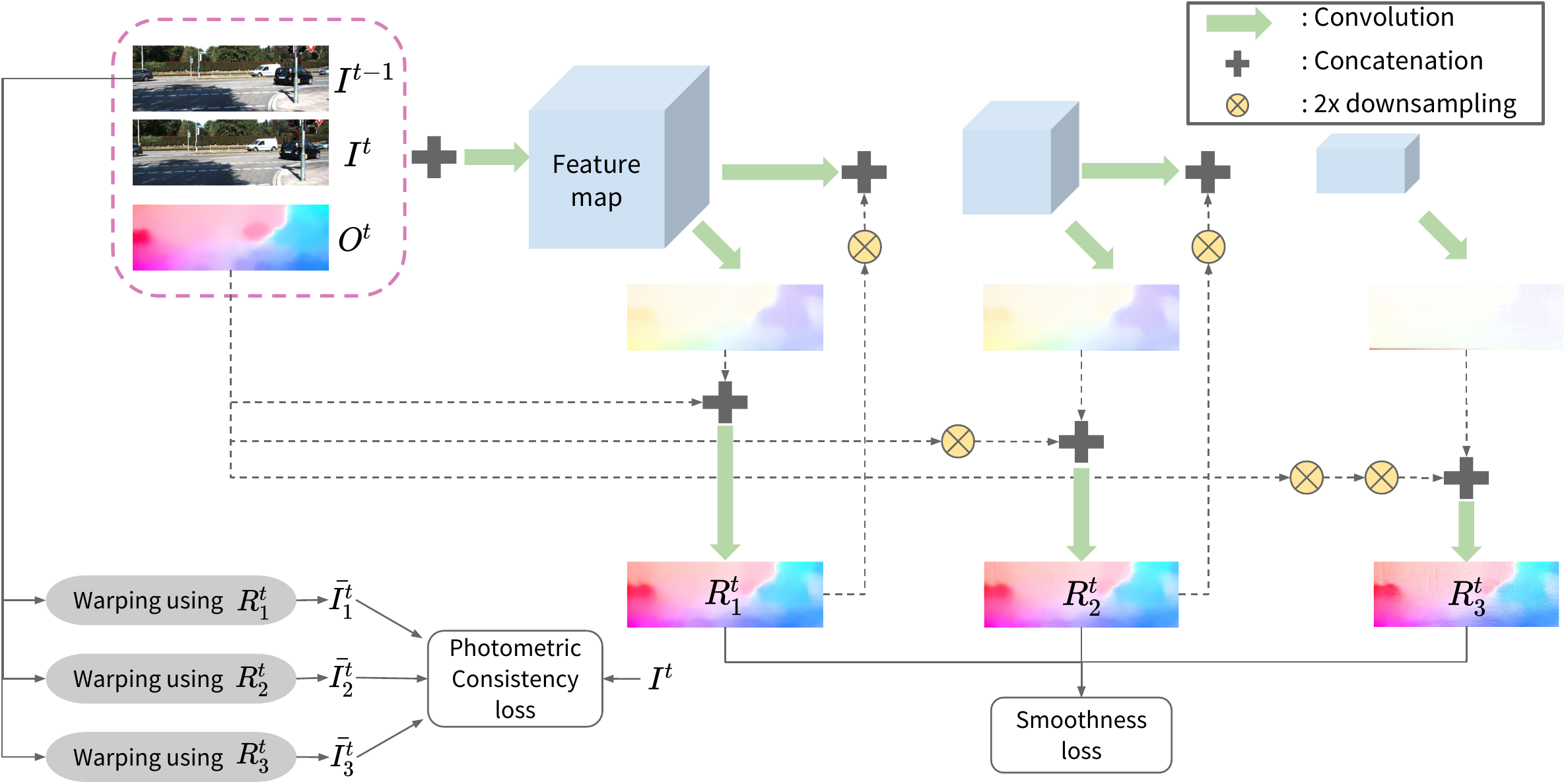}
  \caption{Flow refine network. The flow refine network inputs video frames, $I^{t-1}$ and $I^t$, and optical flow~$O^t$. Using features extracted from concatenated inputs, the network estimates residuals with different scales. They are concatenated to the pre-computed optical flow~$O^t$, and then passed on to additional convolutional layers to obtain the refined ones,~$R_1^t$, $R_2^t$, and $R_3^t$. The network maintains the spatial resolution of~$R_1^t$ to be the same as that of optical flow~$O^t$ while reducing~$R_2^t$ and~$R_3^t$ by a factor of 2 and 4 in each dimension, respectively. To compute the refined flows, we use a multi-scale loss using photometric consistency and smoothness terms. The photometric consistency term computes the differences between video frames and warped ones using the refined flows. The smoothness term encourages the refined flows to be smooth while preserving flow discontinuities. See Table~\ref{tabl:network} for the detailed description of the network structure. (Best viewed in color.)}\label{fig:frn}
\vspace{-0.2cm}
  \end{figure*}

	\subsubsection{ Flow refine network } \label{sec:Flow Refine Network}
  In addition to the extraction of the temporal features from the encoder, we also use optical flow to align video frames and hidden states in a flow-guided memory module over time. Although CNN-based optical flow methods give state-of-the-art results, they are still not accurate enough to propagate fine-grained information across the video frames or hidden states, especially for motion boundaries. To address this problem, we use an additional network for refining the pre-computed optical flow~(Fig.~\ref{fig:frn}). In particular, we learn residuals between the pre-computed optical flow and refined one~\cite{he2016deep}, built upon the assumption that they are similar and the initial flow does not change drastically. Similar to~\cite{gadde2017semantic}, we use an early fusion approach, directly concatenating video frames and optical flow, to transfer the low-level information in the frames to the initial flow effectively. On the contrary, we compute the refined flow using a multi-scale architecture~(Fig.~\ref{fig:frn}) and use it to align both video frames and hidden states in the memory module for the task of depth prediction. Specifically, the network extracts spatiotemporal features from video frames, $I^t$ and $I^{t-1}$, and optical flow $O^t$, and computes the residuals through convolutional layers. They are concatenated to the pre-computed optical flow, and the results are then passed on to additional convolutional layers, resulting in a refined flow~$R_1^t$~specific to aligning video frames. The spatial resolution of~$R_1^t$ is the same as that of the pre-computed optical flow~$O^t$. Other refined flow fields, $R_2^t$ and $R_3^t$, are similarly computed by applying convolutional layers, while reducing the spatial resolution of the pre-computed optical flow~$O^t$ by factor of 2 and 4 in each dimension, respectively.

	\subsubsection{ Flow-guided memory } \label{sec:Visual Memory} 

\begin{figure}[!t]
	\centering
	\includegraphics[width=0.8\linewidth]{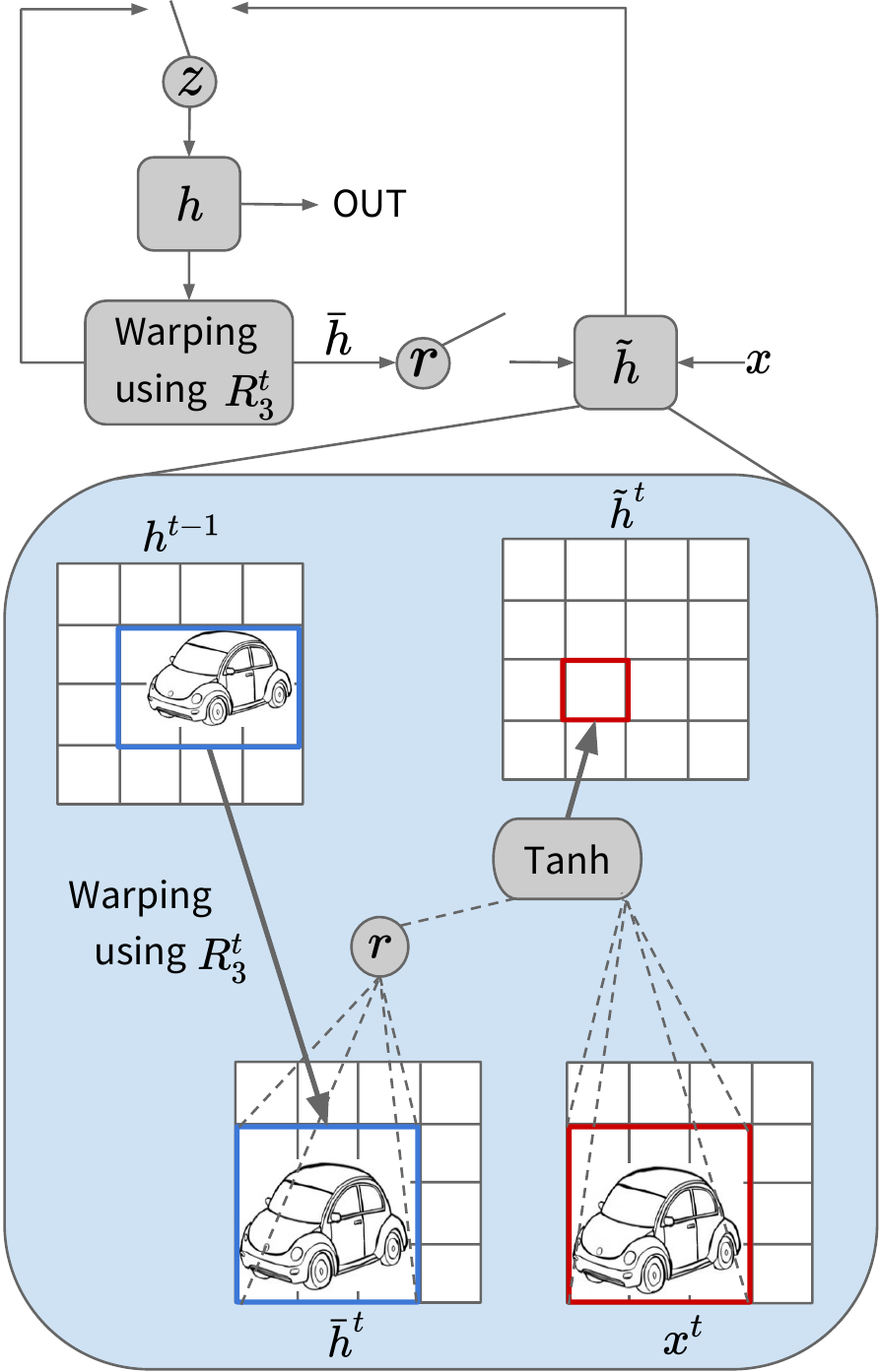}
	\caption{Illustration of the flow-guided memory module. It uses the refined flow~$R_3^t$ to align the hidden state~$h^{t-1}$ at time~$t-1$ to the input feature $x^t$ at time $t$, making a new candidate state~$\tilde h^t$. On the contrary, conventional GRUs compute the candidate state~$\tilde h^t$ using~$h^{t-1}$ directly. This is problematic when the previous hidden state~$h^{t-1}$ and the current input~$x^t$ are not spatially aligned, due to~\emph{e.g.,} viewpoint changes and moving objects. See Table~\ref{tabl:network} for the detailed description of the network structure. (Best viewed in color.)}
	\label{fig:flowgru}
	\vspace{-0.2cm}
	\end{figure}

    Our memory module exploits trajectories of individual pixels using the refined optical flow to align hidden states selectively across frames~(Fig.~\ref{fig:flowgru}). This allows to transfer a long-term temporal consistency in a video sequence to depth prediction results. We implement the memory module with a ConvGRU~\cite{ballas2015delving}, since it does not suffer from spatial resolution loss and is more efficient in terms of memory, compared to vanilla RNNs and ConvLSTMs~\cite{jaderberg2015spatial}, respectively. The flow-guided memory module is defined as follows:
    \begin{align}
		  \bar{h}^{t} &= M^t \odot \mathcal W ( h^{t-1}; R_3^t) \label{eq:Ht} \\
		  z^t &= \sigma( W_{xz} * x^t + W_{\bar h z} * \bar{h}^{t} + b_z ) \label{eq:Z} \\
		  r^t &= \sigma( W_{xr} * x^t + W_{\bar hr} * \bar{h}^{t} + b_r ) \label{eq:R} \\
		  \tilde{h^t} &= \tanh( W_{x\tilde{h}} * x^t + r^t \odot (W_{\bar h\tilde{h}} * \bar{h}^{t}) + b_{\tilde{h}} ) \label{eq:Htil}
		\end{align}
	\begin{align}
		  h^t &= (1-z^t) \odot \bar{h}^{t} + z^t \odot \tilde{h^t} \label{eq:H},     
		\end{align}
where~$\odot$ and~$*$ are element-wise multiplication and convolution, respectively. Here, we denote by $\mathcal{W}$ a warping operator using a flow field,~\emph{e.g.},~$\mathcal{W}(h^{t-1}; R^t_3)(p) = h^{t-1}(p+R^t_3(p))$ at position~$p$. $W$ and $b$ are weight and bias terms, respectively. $\sigma$ is the sigmoid function.

    The flow-guided memory module inputs the feature~$x^t$ obtained from the two-stream encoder and a previous hidden state~$\bar{h}^{t}$ acting as a visual memory, and outputs a new state~$h^t$ by combining~$\bar{h}^{t}$ and a candidate state~$\tilde{h^t}$ weighted by an output of an update gate~$z^t$. The update and reset gates, $z^t$ and $r^t$, selectively choose and discard information, respectively, from the input feature~$x^t$ and the previous hidden state~$\bar{h}^{t}$. Conventional GRUs aggregate features from the hidden state~$h^{t-1}$ at time~$t-1$ directly to compute the current one~$h^t$. This is problematic especially when input features from previous and current frames, $x^{t-1}$ and $x^t$, are not aligned with each other. Examples of this issue are cases when objects move across video frames or the viewpoint is changed due to camera motion. Mixing features from different locations leads to temporally inconsistent results. To address this problem, we instead use a flow-guided memory~$\bar{h}^{t}$ where the feature from the previous state~$h^{t-1}$ are aligned to the current input feature~$x^t$ by warping using the refined flow~$R_3^t$. We implement this with a differential warping operator using bilinear interpolation~\cite{jaderberg2015spatial}. We additionally use a matching confidence~$M^t$ to consider reliability of the refined optical flow~$R_3^t$ as follows.
    \begin{equation}
	     M^t(p) = \exp(- \epsilon {\lVert I_3^t(p) - \bar I_3^t(p) \rVert}_1 ),
    \end{equation}
where $\bar I_i^t =\mathcal W( I_i^{t-1}; R_i^t)$ and $\epsilon$ is a bandwidth parameter. We denote by~$I_i^t$ a resized video frame at time~$t$ that has the same spatial resolution as~$R_i^t$. We use the same matching confidence in each channel of the hidden state~$h^{t-1}$.

    Recently, Shi~\emph{et al}. introduce the TrajGRU~\cite{shi2017deep} for precipitation nowcasting. Our model is closely related to the TrajGRU in that both consider temporally aligned hidden states to compute a new one. The TrajGRU learns offsets for sampling locations~\cite{dai2017deformable}, typically defined on a regular grid in the standard convolution, to fetch information from the previous frame. Although this can be seen as an \emph{implicit} feature alignment, the TrajGRU is not designed to enforce temporal consistency and does not consider large displacements. It may provide temporally inconsistent results when the learned offsets are wrong or displacements between video frames are large. The TrajGRU is also computationally inefficient, since it applies a warping operator for each offset. Compared to this work, we align hidden states in the memory module \emph{explicitly} using the refined optical flow together with a matching confidence. This considers large motion and prevents aggregating the hidden states for unreliable correspondences, making it possible to obtain temporally consistent results.

	\subsubsection{ Decoder } \label{sec:Decoder}
    The decoder inputs the hidden state in the flow-guided memory module and gives depth maps that have the same resolution as input images. In order to consider fine details~(\emph{e.g.}, depth boundaries), we use additional low-level features from spatial and temporal streams by skip connections~(Fig.~\ref{fig:timestep}).

\vspace{-0.3cm}
\subsection{ Training loss } \label{sec:Loss Functions}
We use three types of losses for training: First, a scale-invariant term is used to alleviate scale ambiguity in predicted depth.  Second, we use a photometric consistency term to learn the refined flow, making the pre-computed optical flow specific to aligning video frames. Finally, smoothness terms regularize depth and flow fields. Our final loss is a linear combination of them, balanced by the parameter~$\lambda$ as
  \begin{equation}
      \mathcal{L}= \mathcal{L}^D + {\lambda} \sum_{i} {\mathcal{L}_i^O},
  \end{equation}
where $i \in \{1,2,3\}$. $\mathcal{L}^D$ and $\mathcal{L}_i^O$ are losses for depth prediction and flow refinement, respectively. In the following, we describe each term in detail. 

	\subsubsection{$\mathcal{L}^D$ for depth prediction} \label{Loss Functions for Depth}
Motivated by the work~\cite{eigen2014depth}, we define the scale-invariant loss as 
    \begin{equation}
       \mathcal{L}^{SI} = \frac{1}{N} \sum_{p} s^2(p) - \frac{\alpha}{N^2}  \sum_{p,q} s(p)s(q), \label{eq:L_si}
    \end{equation}
where $s(p) = \log D^t(p) - \log G^t(p)$ is the difference between the predicted depth~$D^t$ and ground truth~$G^t$ at position~$p$~in log space, and $N$ is the total number of pixels. The first term encourages predicted depth to be similar to ground truth. Estimating absolute scale of depth is, however, extremely hard especially from monocular video sequences. The second term alleviates this problem by comparing relationships between pairs of pixels~$p$, $q$ in~$s$. It encourages them to have the same direction, and gives lower error when both $s(p)$ and $s(q)$ are positive or negative values. The first and second terms are balanced by $\alpha \in [0,1]$. As $\alpha$ approaches to one, predicted depth becomes robust to scale variations. We also use the smoothness term that regularizes a prediction result, while preserving depth discontinuities, defined as 
    \begin{equation}
       \mathcal{L}^{DS} =\frac{1}{N}\sum_{p} {\|\nabla^2 D^t(p)\|_1 \ e^{-\gamma\|\nabla^2 I^t (p) \|_1}}, \label{eq:L_ds}
    \end{equation}
where $\nabla^2$ is a Laplace operator and $\gamma$ is the smoothness bandwidth. We compute the second-order derivative of a predicted depth map weighted using the magnitude of image discontinuities, with an assumption that depth boundaries are aligned well to image discontinuities. We define a total loss for depth prediction as
 \begin{equation}
       \mathcal L^D = L^{SI} + \lambda_{D} L^{DS}, \label{eq:L_D}
    \end{equation}
where $\lambda_{D}$ balances the scale-invariant and smoothness terms.

  \subsubsection{$\mathcal{L}_i^O$ for flow refinement } \label{Loss Functions for Flow}
We use the photometric consistency loss to refine the pre-computed optical flow. This term encourages the refined flow to be specific to aligning video frames over time. Motivated by the works~\cite{zhao2017loss,godard2017unsupervised}, we define the consistency term but in a multi-scale manner as
\begin{dmath}
      \mathcal L^{PH}_i = \frac{1}{N_i} \sum_{p} \left( \beta \ \frac{1 - \text{SSIM}({I_i^t(p)}, \bar{I}_i^t(p))}{2} \\
      + (1-\beta) {\lVert I_i^t(p) - \bar{I}_i^t (p) \rVert}_1 \right),
\end{dmath} \label{eq:L_ph}
where $N_i$ is the total number of pixels in the image~$I_i^t$. The first and second terms, balanced by $\beta\in[0,1]$, compute the differences and structural similarity~(SSIM) between original images~$I_i^t$ and synthesized ones~$\bar I_i^t$ from~$I_i^{t-1}$ using the corresponding refined flow~$R_i^t$, respectively. Similar to depth prediction, we define the smoothness term for the refined flow as
    \begin{equation}
      \mathcal{L}_i^{OS} = \frac{1}{N_i}\sum_{p} {\|\nabla^2 R_i^t(p)\|_1 \ e^{-\gamma\|\nabla^2 I_i^t(p) \|_1}} \label{eq:L_os},
    \end{equation}
and use a sum of photometric consistency and smoothness terms, balanced by a regularization parameter $\lambda_O$, as a total loss:
 \begin{equation}
       \mathcal{L}_{i}^{O} = \mathcal{L}_{i}^{PH} + \lambda_{O}  \mathcal{L}_{i}^{OS} \label{eq:L_O}.
    \end{equation}

\section{ Experimental results }

\begin{table}[t]
\renewcommand{\arraystretch}{1.3}
\caption{Network architecture details.}

\label{tabl:network}
\centering
\resizebox{\linewidth}{!}{%
	\begin{threeparttable}
	\begin{tabular}{c c c c c c c}
	\hline
	\hline
	\textbf{Layer} & \textbf{Type} & \textbf{K} & \textbf{S} & \textbf{I/O ch} & \textbf{I/O rs} & \textbf{Input} \\
	\hline
	\multicolumn{7}{c}{\textbf{Encoder (Spatial \& temporal)}}\\
	\hline
	Econv1a        & c          & 3          & 2            & 3/32          & 1/2          & $I^t$ or $O^t$ \\
	Econv1b        & c          & 3          & 1            & 32/32         & 2/2          & Econv1a\\
	Econv2a        & c          & 3          & 2            & 32/64         & 2/4          & Econv1b \\
	Econv2b        & c          & 3          & 1            & 64/64         & 4/4          & Econv2a \\
	Econv3a        & d          & 3          & 2            & 64/64         & 4/4          & Econv2b \\
	Econv3b        & c          & 3          & 1            & 64/64         & 4/4          & Econv3a \\
	Econv4a        & d          & 3          & 4            & 64/64         & 4/4          & Econv3b \\
	Econv4b        & c          & 3          & 1            & 64/64         & 4/4          & Econv4a \\
	Econv5a        & d          & 3          & 8            & 64/128        & 4/4          & Econv4b \\
	Econv5b        & c          & 3          & 1            & 128/128       & 4/4          & Econv5a \\
	Econv6a        & d          & 3          & 16           & 128/128       & 4/4          & Econv5b \\
	Econv6b        & c          & 3          & 1            & 128/128       & 4/4          & Econv6a \\
	Econv7a        & d          & 3          & 16           & 128/256       & 4/4          & Econv6b \\
	Econv7b        & c          & 3          & 1            & 256/256       & 4/4          & Econv7a \\
	Econv8a        & d          & 3          & 1            & 256/256       & 4/4          & Econv7b \\
	Econv8b        & d          & 3          & 1            & 256/64        & 4/4          & Econv8a \\
	\hline
	\multicolumn{7}{c}{\textbf{Flow-guided memory}} \\
	\hline
	Gxz            & c          & 5          & 1            & 128/64        & 4/4          & Econv8b~($I^t$) + Econv8b~($O^t$) \\
	Ghz            & c          & 5          & 1            & 64/64         & 4/4          & $\bar h^{t}$ \\
	Gz             & s          & -          & -            & 64/64         & 4/4          & Gxz + Ghz \\
	Gxr            & c          & 5          & 1            & 128/64        & 4/4          & Econv8b~($I^t$) + Econv8b~($O^t$) \\
	Ghr            & c          & 5          & 1            & 64/64         & 4/4          & $\bar h^{t}$\\
	Gr             & s          & -          & -            & 64/64         & 4/4          & Gxr + Ghr \\
	Gxh            & c          & 5          & 1            & 128/64        & 4/4          & Econv8b~($I^t$) + Econv8b~($O^t$) \\
	Ghh            & c          & 5          & 1            & 64/64         & 4/4          & Gr $\odot \hspace{0.1cm} \bar h^{t} $\\
	Gh             & t          & -          & -            & 64/64         & 4/4          & Gxh + Ghh \\
	$h^t$          & -          & -          & -            & 64/64         & 4/4          & ($1-$Gz) $\odot \hspace{0.1cm} \bar h^{t-1}$ + Gz $\odot $ Gh \\ 
	$\bar h^{t+1}$ & - & - & - & 64/64 & 4/4 & $M^{t+1} \odot \mathcal{W}(h^{t}, R^{t+1}_3)$\\ 
	\hline
	\multicolumn{7}{c}{\textbf{Decoder}} \\
	\hline
	Dconv1a        & u          & 5          & 2            & 64/32         & 4/2          & $h^t$ \\
	Dconv1b        & c          & 5          & 1            & 96/32        & 2/2          &
	\begin{tabular}[c]{@{}c@{}}Dconv1a + Econv1b~($I^{t}$) \\ + Econv1b~($O^{t}$) \end{tabular} \\
	Dconv2a        & u          & 5          & 2            & 32/16         & 2/1          & Dconv1b \\
	Dconv2b        & c          & 5          & 1            & 16/16         & 1/1          & Dconv2a \\
	Output         & c          & 5          & 1            & 16/1          & 1/1          & Dconv2b \\
	\hline
	\multicolumn{7}{c}{\textbf{Flow refine network}} \\
	\hline
	Fconv1a        & c          & 3          & 1            & 8/32          & 1/1          & $I^t + I^{t-1} + O^t$ \\
	Fconv1b        & c          & 3          & 1            & 32/2          & 1/1          & Fconv1a \\
	$R^t_1$        & c          & 3          & 1            & 4/2           & 1/1          & Fconv1b + $O^t$ \\
	Fconv2a        & c          & 3          & 2            & 32/32         & 1/2          & Fconv1a \\
	Fconv2b        & c          & 3          & 1            & 34/2          & 2/2          & Fconv2a + $\mathcal{D}(R^t_1)$ \\
	$R^t_2$        & c          & 3          & 1            & 4/2           & 2/2          & Fconv2b + $\mathcal{D}({O^t})$ \\
	Fconv3a        & c          & 3          & 2            & 32/32         & 2/4          & Fconv1a \\
	Fconv3b        & c          & 3          & 1            & 34/2          & 4/4          & Fconv3a + $\mathcal{D}(R^t_2)$ \\
	$R^t_3$        & c          & 3          & 1            & 4/2           & 4/4          & Fconv3b + $\mathcal{D}(\mathcal{D}({O^t}))$\\
	\hline
	\hline
	\end{tabular}
	\begin{tablenotes}
        {\textbf{Type}: A type of operations; \textbf{K}: Kernel size; \textbf{S}: Strides; \textbf{I/O ch}: The number of channels for the input/output; \textbf{I/O rs}: A downsampling factor for the input/output relative to the input image. c:~Convolution; d:~Dilated convolution; u:~Up-convolution; s:~Sigmoid;~t: Hyperbolic tangent.}
	\end{tablenotes}
	\end{threeparttable}
	}
\vspace{-0.28cm}
\end{table}

 In this section we present a detailed analysis and evaluation of our approach. Our code and more results including depth videos are available at our project webpage:~\url{https://cvlab-yonsei.github.io/projects/FlowGRU/}
  
 \begin{table*}[ht]
  \renewcommand{\arraystretch}{1.0}
\caption{Quantitive comparison with the state of the art on monocular depth prediction with the test split provided by~\cite{eigen2014depth}.}

  \label{tabl:benchmark}
  \centering
  \resizebox{\linewidth}{!}{
  \begin{threeparttable}
  \begin{tabular}{l c c c c c c c c c c c}
  \hline
  \hline
   & & & & \multicolumn{4}{c}{\centering lower is better} &  \multicolumn{3}{c}{\centering higher is better} \\
   \cmidrule(lr){5-8}  \cmidrule(lr){9-11}
  \parbox{7em}{\centering Method} & \parbox{5em}{\centering Dataset} & \parbox{5em}{\centering Supervision} & \parbox{3em}{\centering cap} & Abs Rel & Sq Rel & RMSE & RMSE~(log) & $\delta<1.25$ & $\delta<1.25^2$ & $\delta<1.25^3$ & Runtime(s) \\ 
  \hline
  Eigen~\emph{et al.}~\cite{eigen2014depth} & K & D & 0-80m & 0.215 & 1.515 & 7.156 & 0.270 & 0.692 & 0.899 & 0.967 & - \\
  Liu~\emph{et al.}~\cite{liu2014discrete} & K & D & 0-80m & 0.217 & 1.841 & 6.986 & 0.289 & 0.647 & 0.882 & 0.961 & - \\
  Godard~\emph{et al.}~\cite{godard2017unsupervised} & K & S & 0-80m & 0.148 & 1.344 & 5.927 & 0.247 & 0.803 & 0.922 & 0.964 & \underline{0.04} \\
  Zhou~\emph{et al.}~\cite{zhou2017unsupervised} & K & M & 0-80m & 0.208 & 1.768 & 6.856 & 0.283 & 0.678 & 0.885 & 0.957 & \textbf{0.03} \\
  Wang~\emph{et al.}~\cite{wang2018learning} & K & M & 0-80m & 0.151 & 1.257 & 5.583 & 0.228 & 0.810 & 0.936 & 0.974 & \textbf{0.03} \\
  Yin~\emph{et al.}~\cite{yin2018geonet} & K & M & 0-80m & 0.155 & 1.296 & 5.857 & 0.233 & 0.793 & 0.931 & 0.973 & \underline{0.04} \\
  Kuznietsov~\emph{et al.}~\cite{kuznietsov2017semi} & I+K & D+S & 0-80m & 0.113 & 0.741 & 4.621 & 0.189 & 0.862 & 0.960 & \textbf{0.986} & 0.06 \\
  Kumar~\emph{et al.}~\cite{cs2018depthnet} & K & D+M & 0-80m & 0.137 & 1.019 & 5.187 & 0.218 & 0.809 & 0.928 & 0.971 & - \\
  Fu~\emph{et al.}~\cite{fu2018deep} & I+K & D+M & 0-80m & \textbf{0.102} & \textbf{0.617} & \textbf{3.859} & \textbf{0.165} & \textbf{0.890} & \textbf{0.964} & \underline{0.985} & 1.08 \\
  Ours & K & D+M & 0-80m & 0.117 & 0.726 & 4.537 & 0.192 & 0.865 & 0.958 & 0.983 & 0.13 \\
  Ours-CS+ft-K & CS+K & D+M & 0-80m & \underline{0.112} & \underline{0.700} & \underline{4.260} & \underline{0.184} & \underline{0.881} & \underline{0.962} & 0.983 & 0.13 \\
  \cmidrule(lr){1-11}
  Kuznietsov~\emph{et al.}~\cite{kuznietsov2017semi} & I+K & D+S & 1-50m & \textbf{0.108} & \underline{0.595} & 3.518 & \underline{0.179} & 0.875 & \underline{0.964} & \textbf{0.988} & 0.06 \\
  Garg~\emph{et al.}~\cite{garg2016unsupervised} & K & S & 1-50m & 0.169 & 1.080 & 5.104 & 0.273 & 0.740 & 0.904 & 0.962 & \underline{0.04} \\
  Godard~\emph{et al.}~\cite{godard2017unsupervised} & K & S & 1-50m & \textbf{0.108} & 0.657 & 3.729 & 0.194 & 0.873 & 0.954 & 0.979 & \underline{0.04} \\
  Ours & K & D+M & 1-50m & 0.113 & \textbf{0.580} & \underline{3.493} & 0.181 & \underline{0.877} & 0.963 & \underline{0.985} & 0.13 \\
  Ours-CS+ft-K & CS+K & D+M & 1-50m & \underline{0.109} & \textbf{0.580} & \textbf{3.359} & \textbf{0.176} & \textbf{0.891} & \textbf{0.965} & \underline{0.985} & 0.13 \\
  \hline
  \hline
  \end{tabular}
  \begin{tablenotes}
        {Abs Rel: Absolute relative difference; Sq Rel: Square relative difference; RMSE: Root Mean Square Error; RMSE~(log): RMSE in log scale; $\delta<\tau$: The percentage of pixels where the ratio of estimated depth and ground truth is within a range in the threshold~$\tau$. D:~Ground-truth depth; S:~Rectified stereo pairs; M:~Monocular video sequences.}
  \end{tablenotes}
  \end{threeparttable}
  }
	
\vspace{-0.2cm}
  \end{table*}

\begin{figure*}[h]
	  \centering
	  \renewcommand*{\thesubfigure}{}
	  \subfigure[]{
	    \begin{minipage}[t]{\linewidth}
	    \centering
	    \includegraphics[height=0.18\linewidth, width=0.92\linewidth]{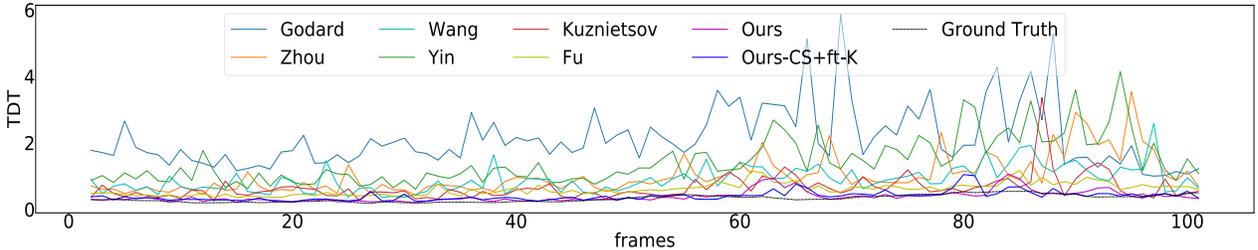}
	    \end{minipage}
	    }
	  \vspace{-0.8cm}
\caption{Examples of TDT variations over time on the KITTI dataset~\cite{geiger2013vision}. Compared to the state of the art~\cite{godard2017unsupervised,zhou2017unsupervised,wang2018learning,yin2018geonet,kuznietsov2017semi,fu2018deep}, our models give lower errors during whole frames and show analogous patterns with ground truth. (Best viewed in color.)}
	  \label{fig:temp}
\vspace{-0.2cm}
	\end{figure*}

\vspace{-0.3cm}
  \subsection{ Training  }

  We train our model from scratch with the KITTI raw dataset~\cite{geiger2013vision} that provides pairs of stereo images for 61 scenes together with 3D points and camera parameters. In particular, we use the split provided by~\cite{eigen2014depth}, where it contains 35,600 and 697 images for training and test, respectively. We consider each view in stereo image pairs as an individual monocular sequence. We also train our model with the Cityscapes dataset~\cite{Cordts2016Cityscapes} that consists of 89k, 15k and 45k images for training, validation and test, respectively. We split the training sets into a chunk of frames, each of which contains 50 and 30 successive frames for the KITTI and Cityscapes datasets, respectively. We choose 20 and 5 nearby frames randomly for the KITTI and Cityscapes dataset, respectively, and augment the datasets by randomly cropping training samples to the size of~$960 \times 320$. We use a batch size of 16 for 200 epochs which corresponds to about 450k iterations for the KITTI dataset. For the Cityscapes dataset, the same batch size of 16 is used with 200 epochs~(about 600k iterations), and the trained model is then fine-tuned with additional 100 epochs with the train split provided by~\cite{eigen2014depth}. We use the Adam optimizer~\cite{kingma2014adam} with $\beta_1=0.9$ and $\beta_2=0.999$. As learning rate, we use 1e-4 at first 100 epochs and gradually reduce it during training. We use a grid search to set the balance parameters, $\lambda_D$, $\lambda_O$ and $\lambda$, to $0.1$, $0.1$ and $0.05$, respectively. We follow the experimental setting in~\cite{eigen2014depth,godard2017unsupervised,lai2018learning,wang2018occlusion} to set other parameters, and fix them in all experiment: $\alpha = 0.5$, $\beta=0.85$, $\gamma=10$, and $\epsilon=1$. We compute optical flow using the DIS-Flow method~\cite{kroeger2016fast} that offers a good compromise in terms of runtime and accuracy. For example, it requires 0.1 seconds for images of size $1392 \times 512$ with an Intel i5 3.3Ghz CPU. All networks are trained end-to-end using~\texttt{TensorFlow}~\cite{abadi2016tensorflow}. With two Nvidia GTX Titan Xs, training our model takes about 10 and 15 days for the KITTI and Cityscapes datasets, respectively, including fine-tuning.

\vspace{-0.4cm}
\subsection{Network architecture details}
We show a detailed description of the network architecture in Table~\ref{tabl:network}. We denote by ``+", ``$\odot$", and ``$\mathcal{D}(\cdot)$" concatenation, element-wise multiplication, and 2$\times$ downsampling, respectively. We use the ReLU~\cite{krizhevsky2012imagenet} as an activation function except for the last layer. Each sub-network in the encoder consists of 9 convolutional and 7 dilated convolutional layers. A dilated convolution~\cite{yu2015multi} enables covering large receptive fields using small-size convolutions and maintaining the spatial resolution of feature maps, but it typically causes grid artifacts~\cite{yu2017dilated}. To alleviate this problem, we add a convolutional layer followed by the dilated one, except the last two layers. The flow-guided memory module has an architecture similar to the ConvGRU~\cite{ballas2015delving} consisting of reset and update gates. Differently, we align the previous hidden state w.r.t. the current input feature using the refined flow. The decoder has 2 up-convolutional and 3 convolutional layers. Following~\cite{mayer2016large}, we add a convolutional layer after applying an up-convolutional operator, which gives smooth prediction results. We use skip connections from the encoder to leverage low-level but fine-grained features for depth prediction. The spatial resolution of predicted depth is the same as that of an input frame. The flow refine network computes three residuals with different scales. The residual for each scale is computed through 3 convolutional layers. We use the ReLU~\cite{krizhevsky2012imagenet} as an activation function except for the last layer. 

\vspace{-0.3cm}
\subsection{ Evaluation }
Depth predicted by our model is defined up to a scale factor. Following the experimental protocol in~\cite{zhou2017unsupervised,wang2018learning}, we multiply a predicted depth map by a constant in order to make median values of predicted depth and ground truth the same. To evaluate our model in terms of temporal consistency, we measure temporal differences along dense trajectories. To this end, we synthesize a depth map~$\bar{D}^{t}$~at time~$t$ by warping~$D^{t-1}$ using optical flow. For fair comparison, we use an optical flow method~\cite{sun2018pwc} different from the one~\cite{kroeger2016fast} used in our model. We then compute the differences between $\bar{D}^{t}$ and ${D}^{t}$ over time. That is, we compute temporal differences along trajectories~(TDT) as follows.
  \begin{equation}\label{eq:TDT}
      \text{TDT}(t) = {1 \over N} \sum_{p}  C^t(p)  \lVert D^t(p) - \bar{D}^t(p) \rVert_1,
    \end{equation}
where a binary confidence map~$C^t$ represents reliability of optical flow, defined as 
\begin{equation}
C^t(p) = 
\begin{cases}
  1, & \text{if } \exp(- \epsilon_1 {\lVert I^t(p)}- \bar{I}^t(p) \rVert_1 )> th.\\
  0, & \text{otherwise}.
\end{cases}
\end{equation}
We set $\epsilon_1$~and~$th$ to 0.5 and 0.05, respectively. We also compute the percentage of erroneous pixels, denoted by TDT $<1$, TDT $<2$, and TDT $<3$, where a point is considered to be erroneous when the differences are more than 1, 2, and 3, respectively.

\begin{figure*}[t]
  \centering
  \renewcommand*{\thesubfigure}{}
  \hspace{-0.3cm}
  \subfigure[]{
    \begin{minipage}[t]{0.23\linewidth}
      \includegraphics[width=\linewidth]{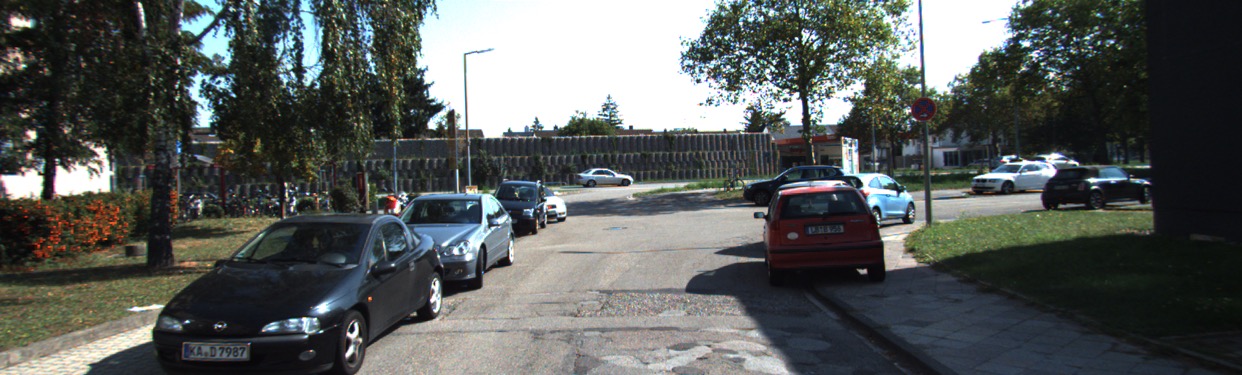}\vspace{0.11cm}
      \begin{overpic}[width=\linewidth]{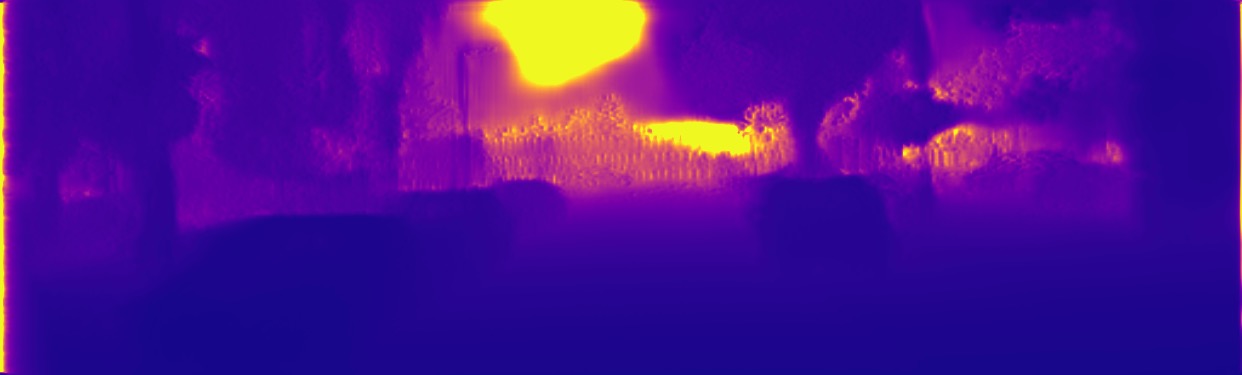}\vspace{0.1cm}
      \put(2,24){\color{white}{\scriptsize Godard~\emph{et al.}}}
      \end{overpic}\vspace{0.108cm}
      \begin{overpic}[width=\linewidth]{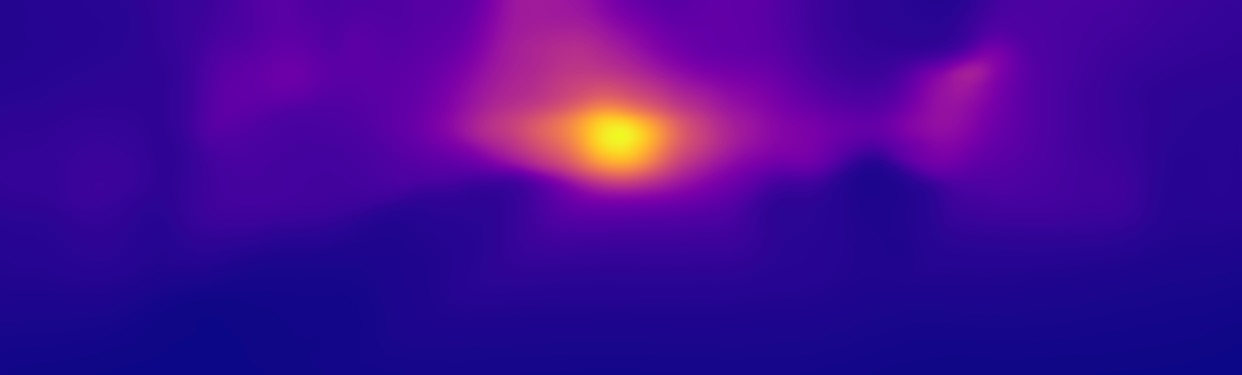}\vspace{0.1cm}
      \put(2,24){\color{white}{\scriptsize Zhou~\emph{et al.}}}
      \end{overpic}\vspace{0.108cm}
      \begin{overpic}[width=\linewidth]{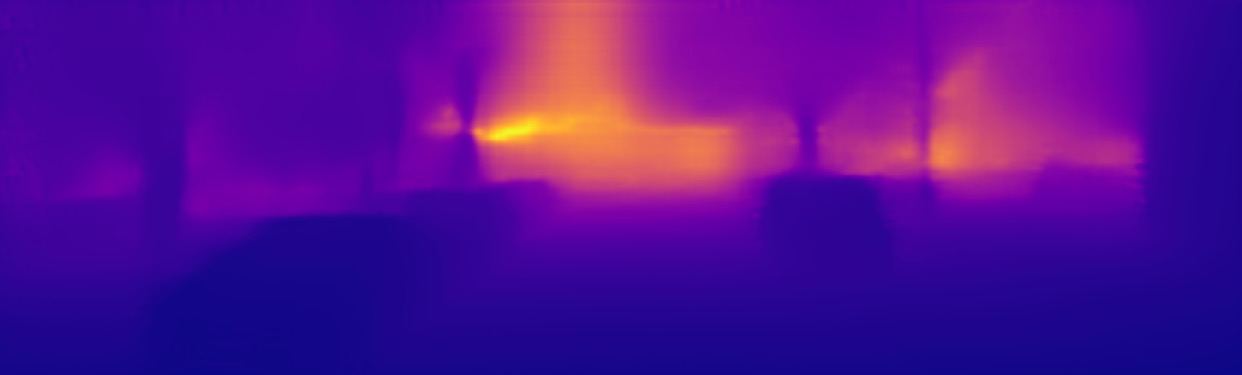}\vspace{0.1cm}
      \put(2,24){\color{white}{\scriptsize Wang~\emph{et al.}}}
      \end{overpic}\vspace{0.108cm}
      \begin{overpic}[width=\linewidth]{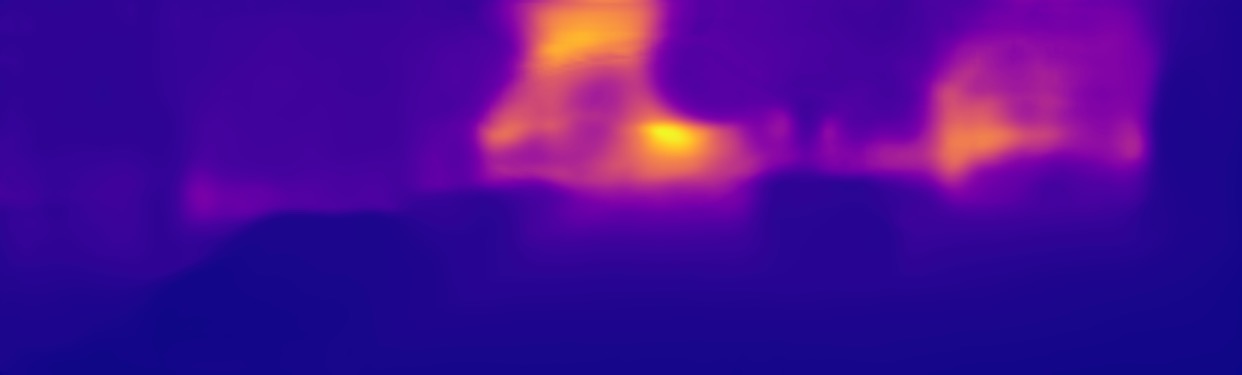}\vspace{0.1cm}
      \put(2,24){\color{white}{\scriptsize Yin~\emph{et al.}}}
      \end{overpic}\vspace{0.108cm}
      \begin{overpic}[width=\linewidth]{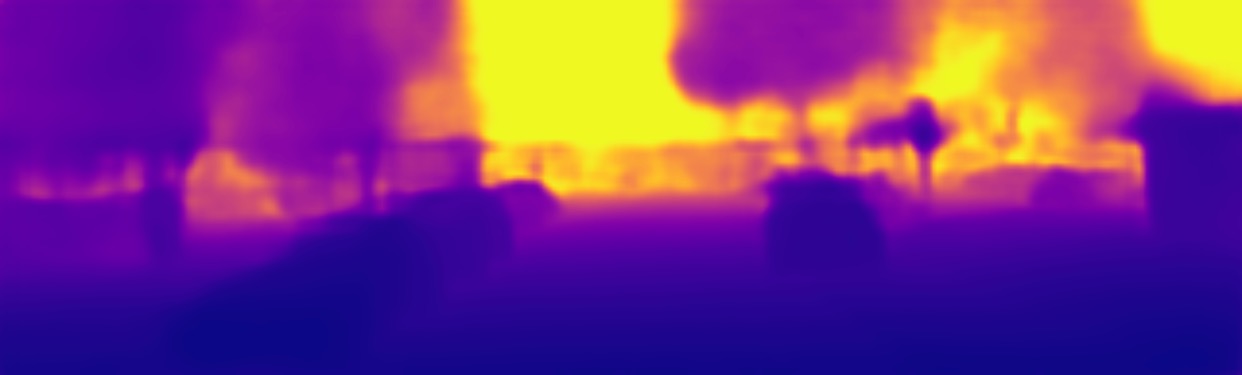}\vspace{0.1cm}
      \put(2,24){\color{white}{\scriptsize Kuznietsov~\emph{et al.}}}
      \end{overpic}\vspace{0.108cm}
      \begin{overpic}[width=\linewidth]{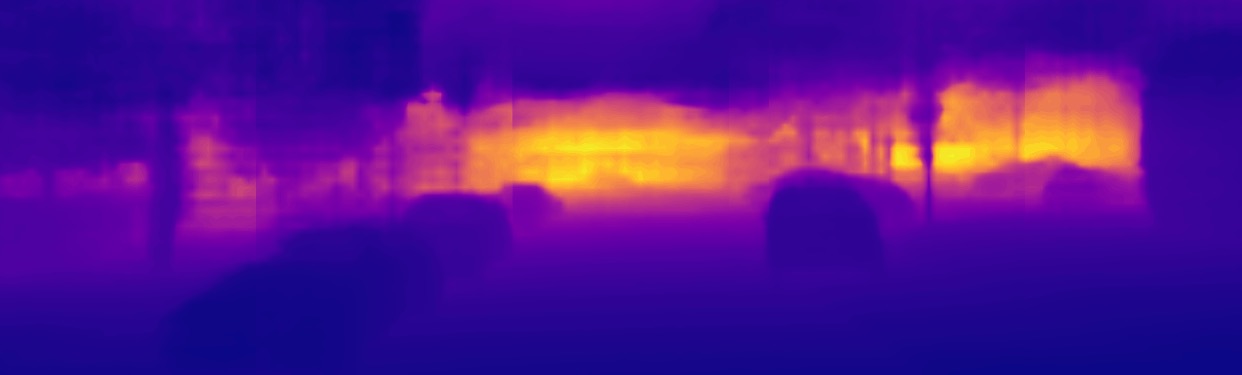}\vspace{0.1cm}
      \put(2,24){\color{white}{\scriptsize Fu~\emph{et al.}}}
      \end{overpic}\vspace{0.108cm}
      \begin{overpic}[width=\linewidth]{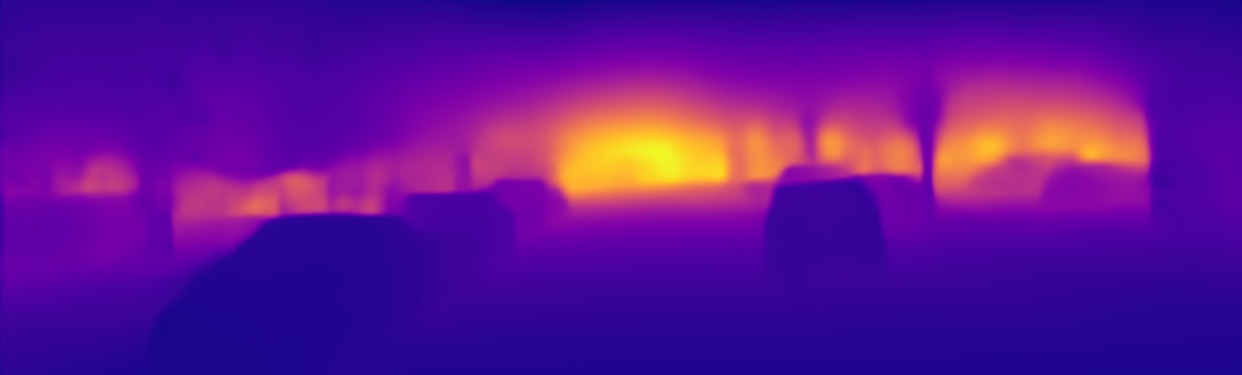}\vspace{0.1cm}
      \put(2,24){\color{white}{\scriptsize Ours}}
      \end{overpic}\vspace{0.108cm}
      \begin{overpic}[width=\linewidth]{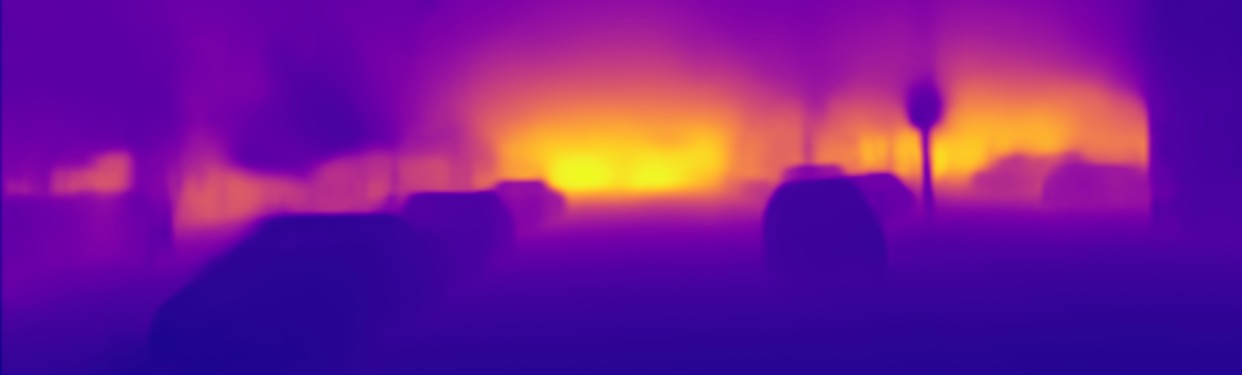}\vspace{0.1cm}
      \put(2,24){\color{white}{\scriptsize Ours-CS+ft-K}}
      \end{overpic}\vspace{0.108cm}
      \begin{overpic}[width=\linewidth]{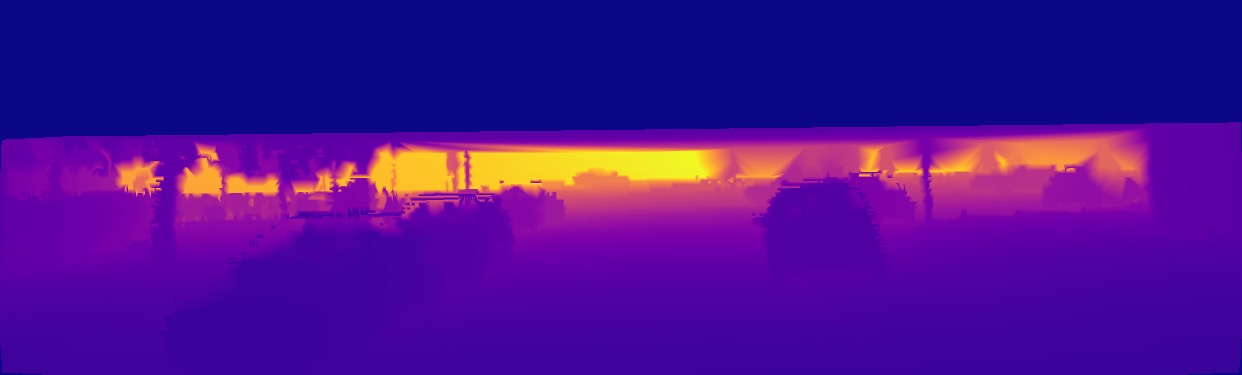}\vspace{0.1cm}
      \put(2,24){\color{white}{\scriptsize Ground truth}}
      \end{overpic}\vspace{0.108cm}
    \end{minipage}
    }\hspace{-0.3cm}
  \subfigure[]{
    \begin{minipage}[t]{0.23\linewidth}
      \includegraphics[width=\linewidth]{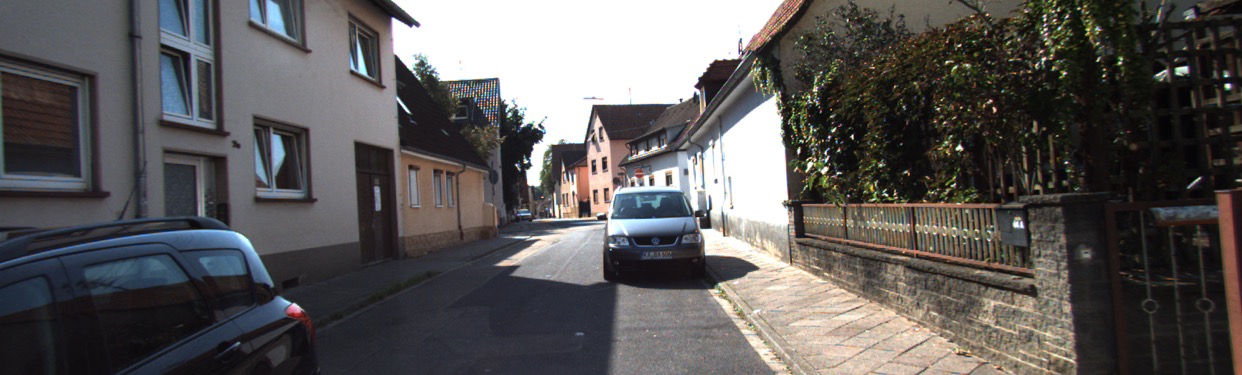}\vspace{0.1cm}
      \includegraphics[width=\linewidth]{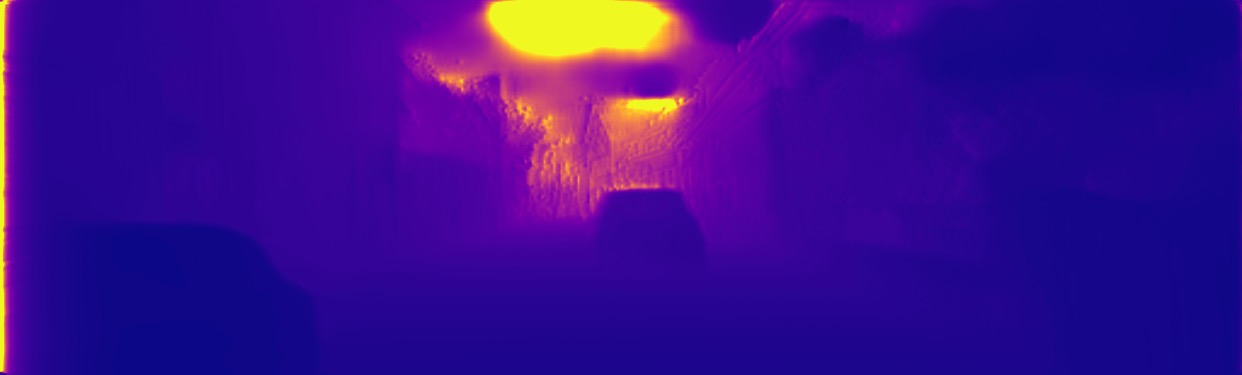}\vspace{0.1cm}
      \includegraphics[width=\linewidth]{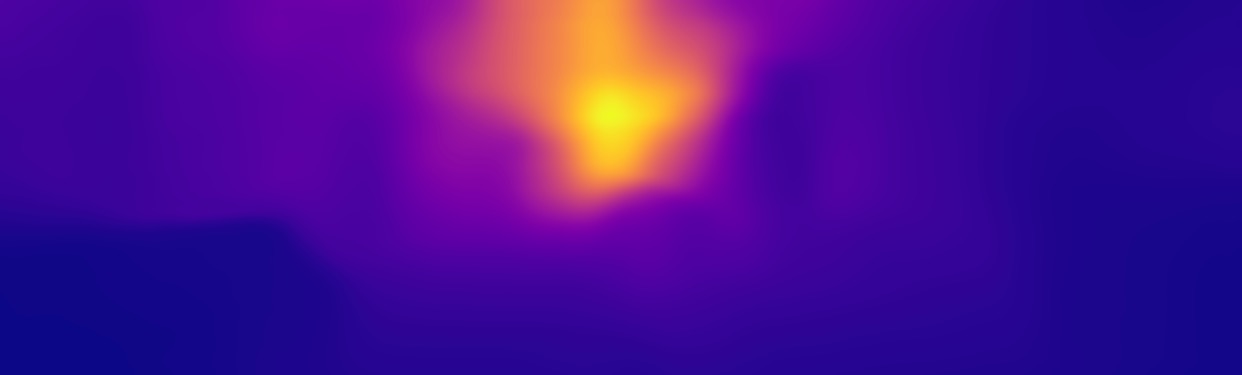}\vspace{0.1cm}
      \includegraphics[width=\linewidth]{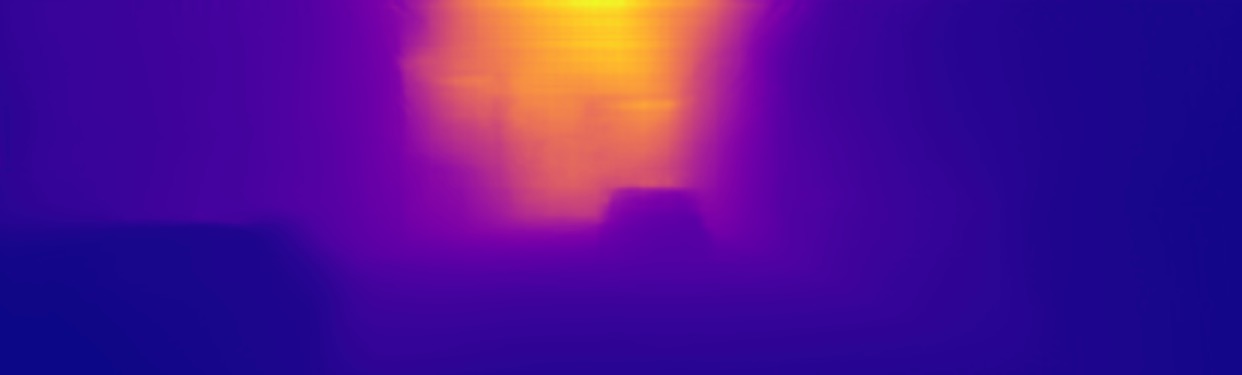}\vspace{0.1cm}
      \includegraphics[width=\linewidth]{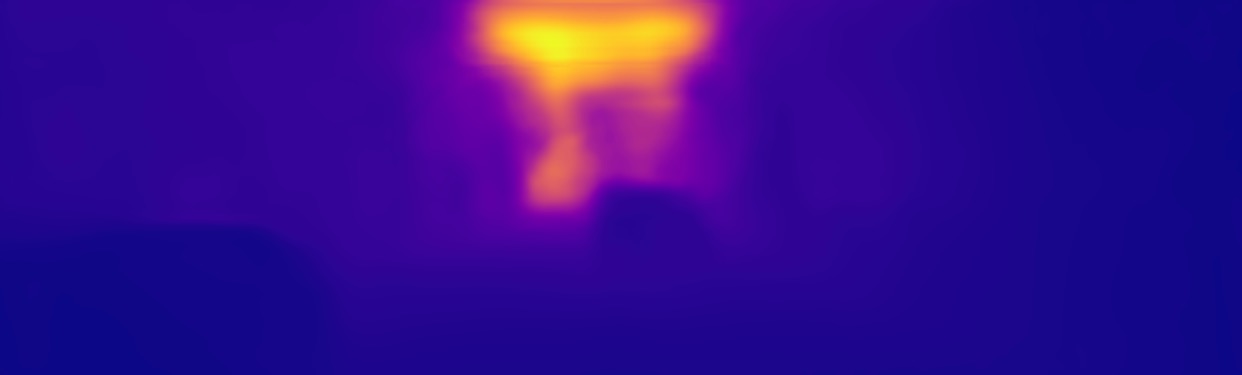}\vspace{0.1cm}
      \includegraphics[width=\linewidth]{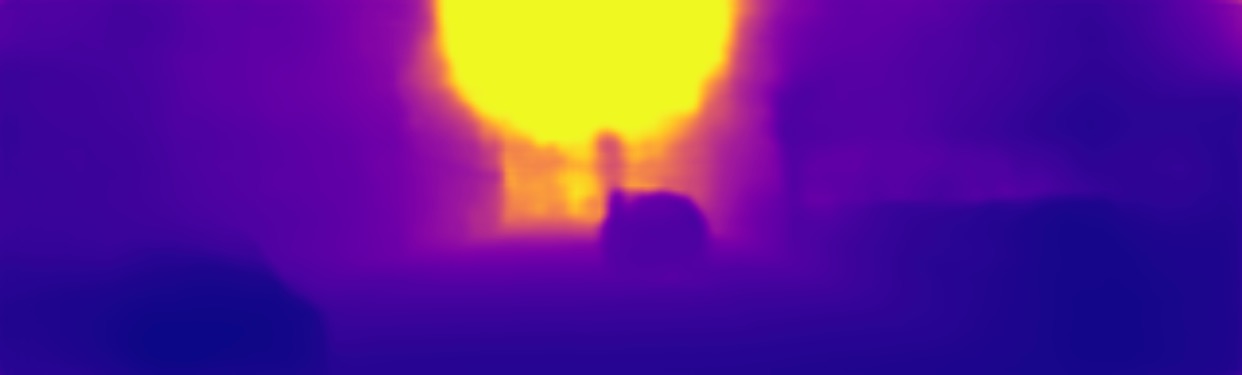}\vspace{0.1cm}
      \includegraphics[width=\linewidth]{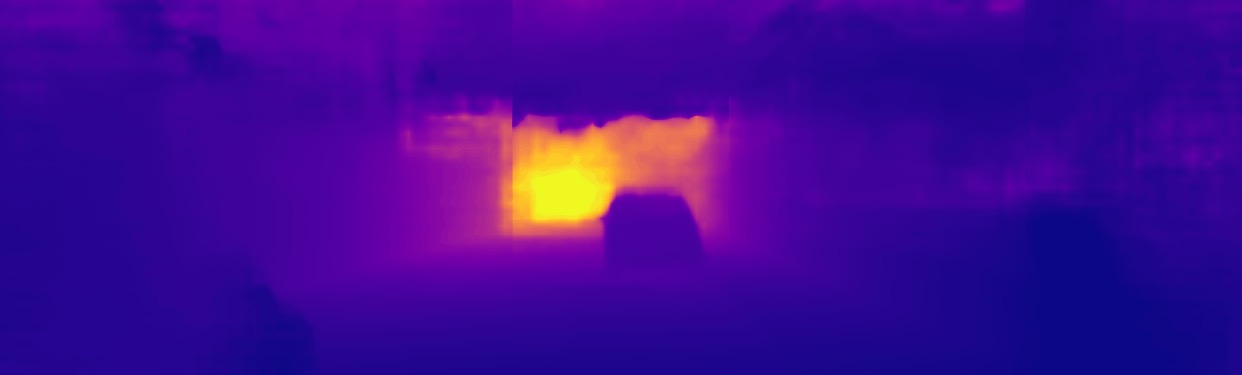}\vspace{0.1cm}
      \includegraphics[width=\linewidth]{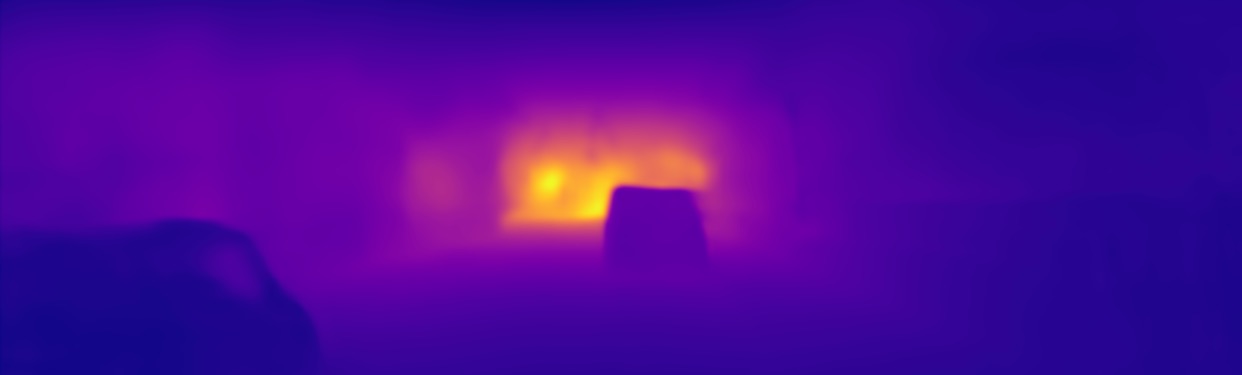}\vspace{0.1cm}
      \includegraphics[width=\linewidth]{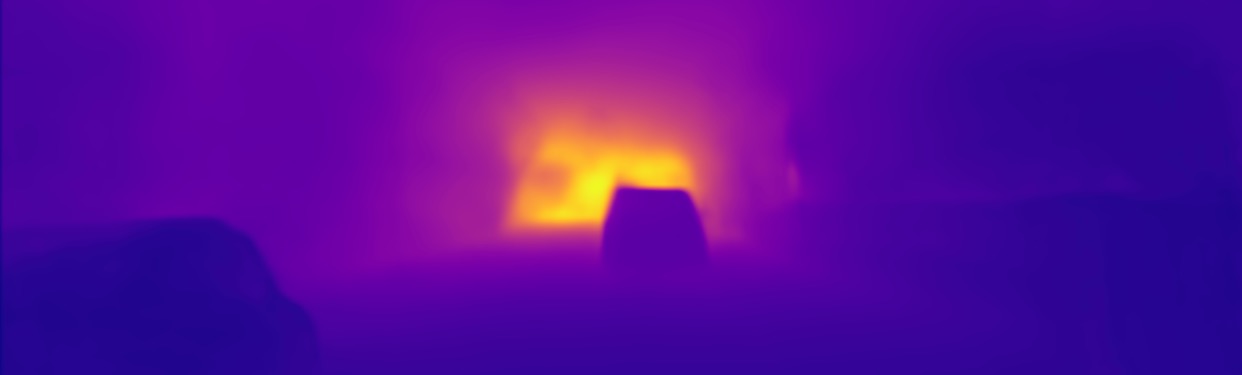}\vspace{0.1cm}
      \includegraphics[width=\linewidth]{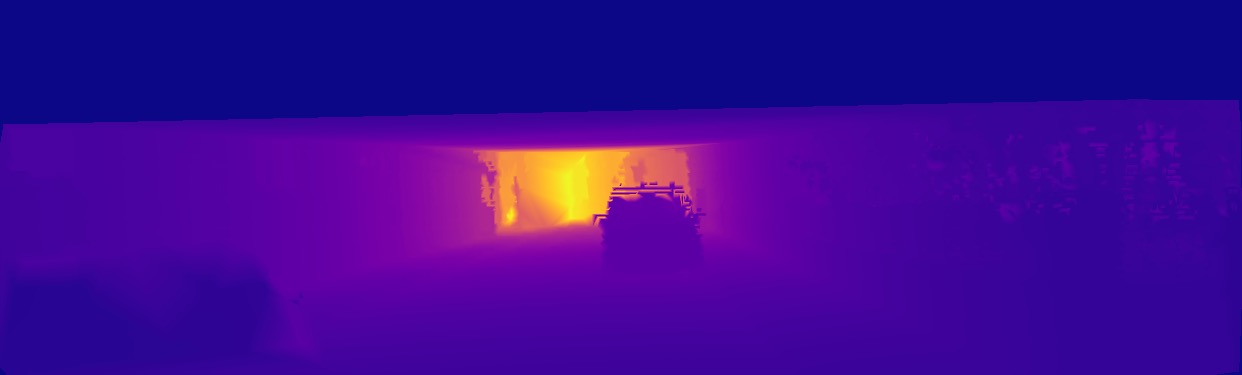}\vspace{0.1cm}
    \end{minipage}
    }\hspace{-0.3cm}
  \subfigure[]{
    \begin{minipage}[t]{0.23\linewidth}
      \includegraphics[width=\linewidth]{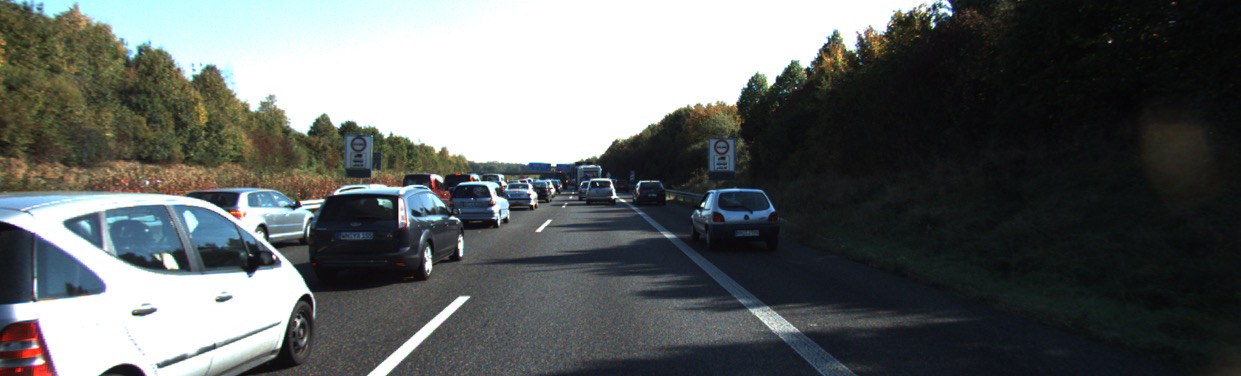}\vspace{0.096cm}
      \includegraphics[width=\linewidth]{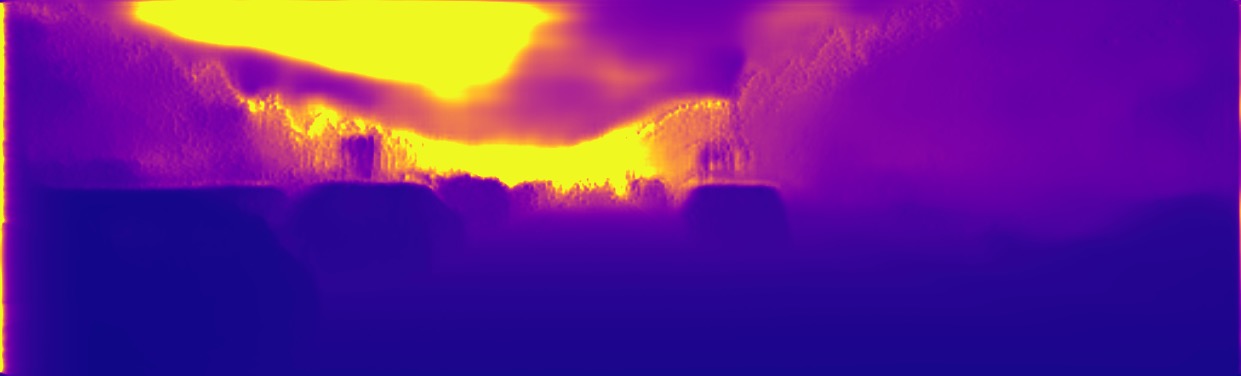}\vspace{0.096cm}
      \includegraphics[width=\linewidth]{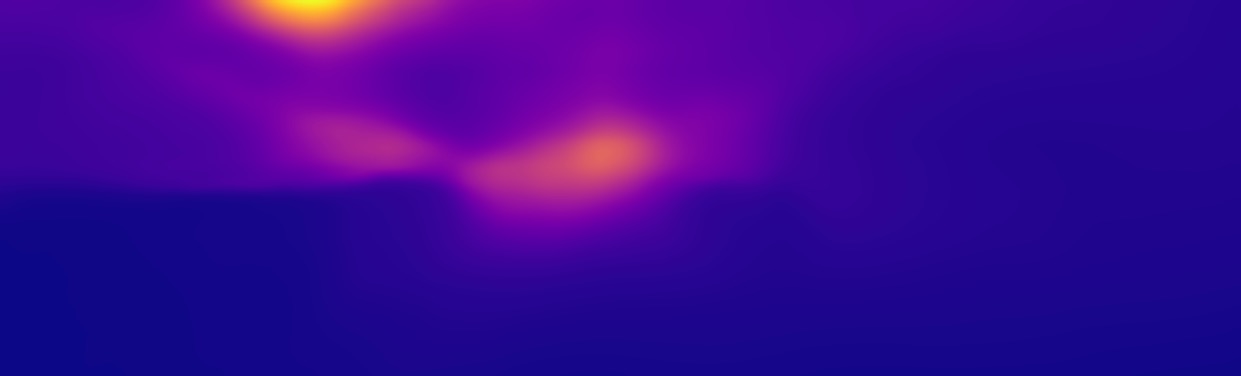}\vspace{0.096cm}
      \includegraphics[width=\linewidth]{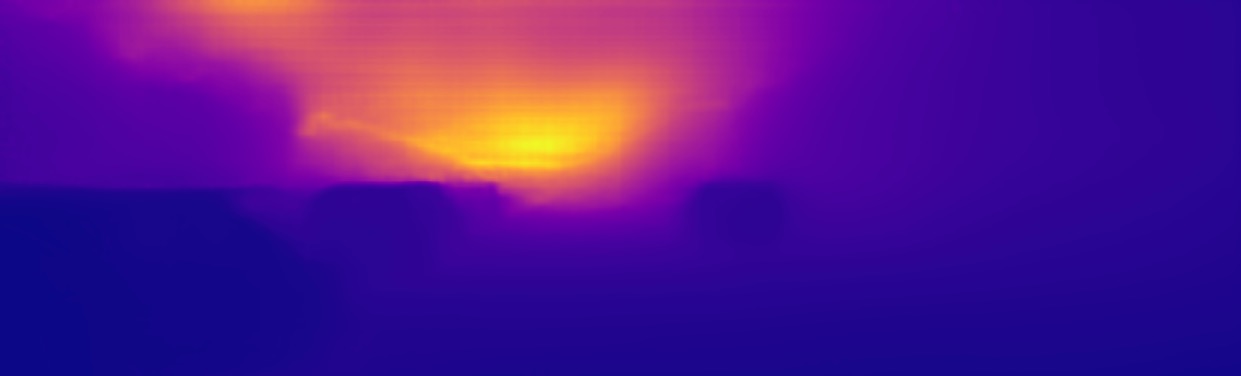}\vspace{0.096cm}
      \includegraphics[width=\linewidth]{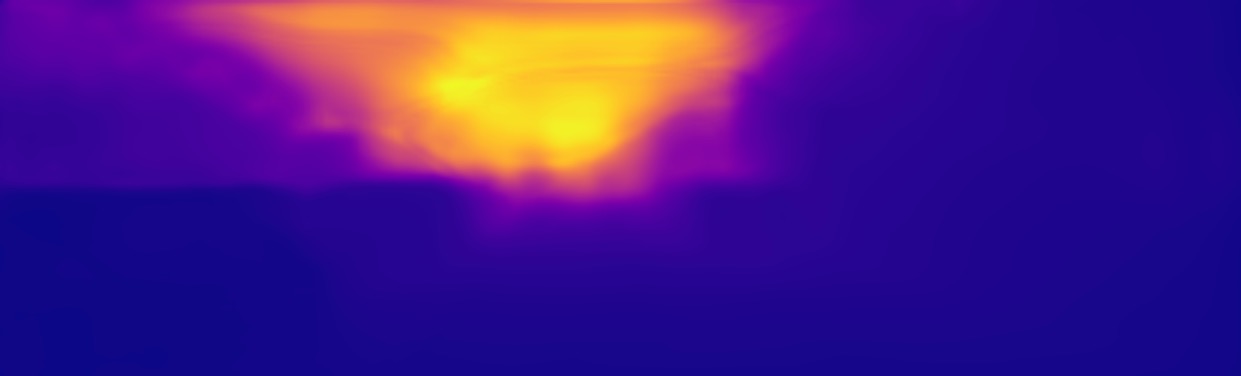}\vspace{0.096cm}
      \includegraphics[width=\linewidth]{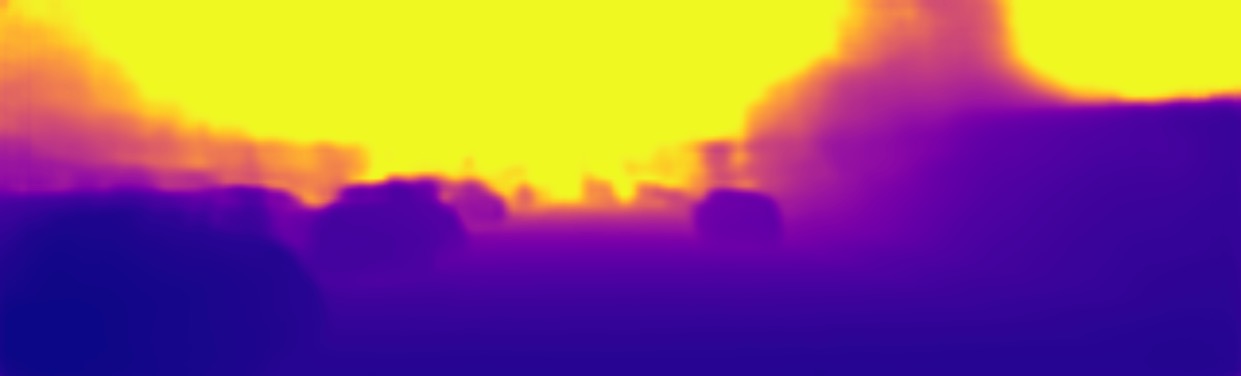}\vspace{0.096cm}
      \includegraphics[width=\linewidth]{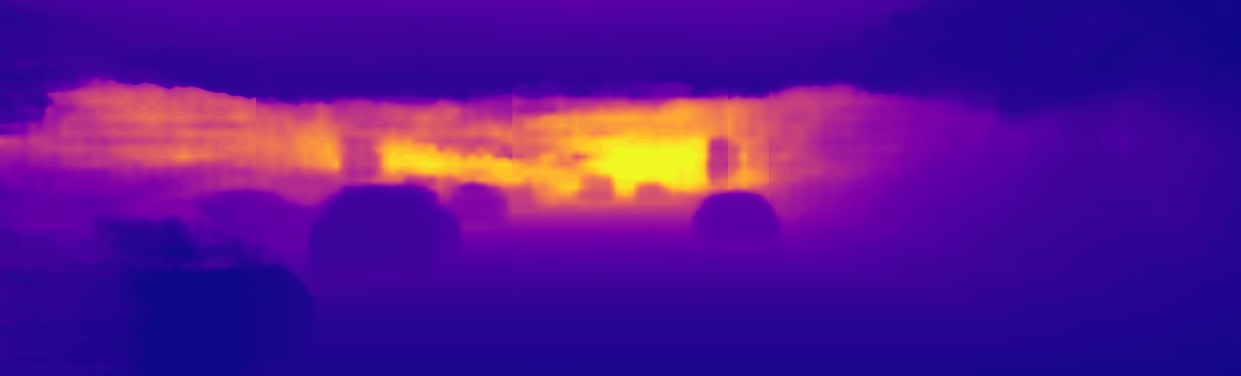}\vspace{0.096cm}
      \includegraphics[width=\linewidth]{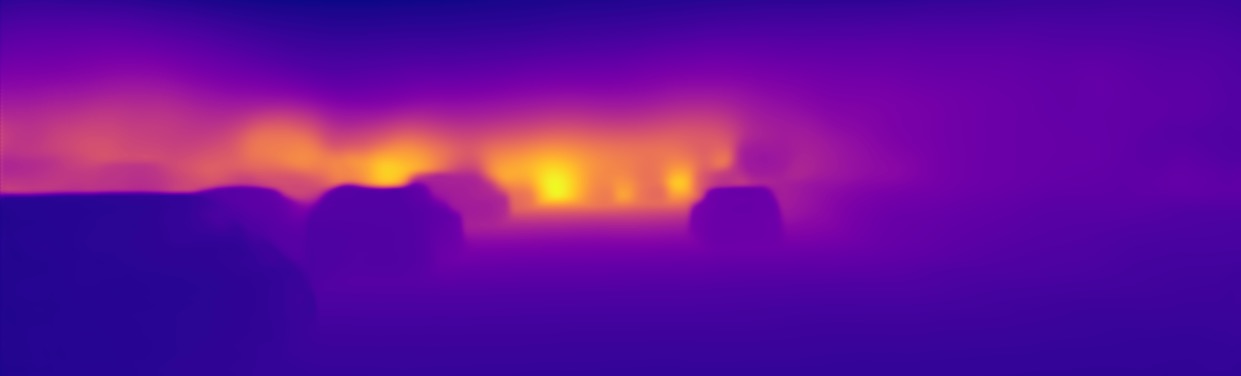}\vspace{0.096cm}
      \includegraphics[width=\linewidth]{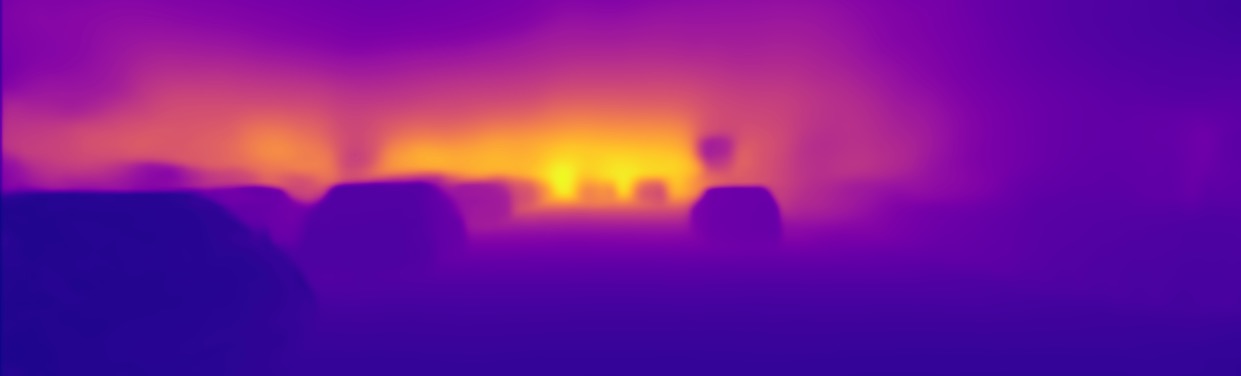}\vspace{0.096cm}
      \includegraphics[width=\linewidth]{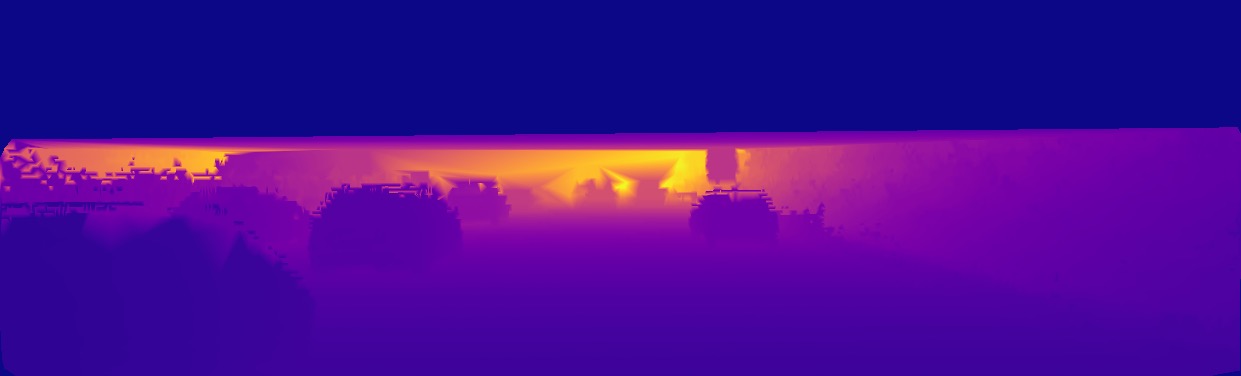}\vspace{0.096cm}
    \end{minipage}
    }\hspace{-0.3cm}
  \subfigure[]{
    \begin{minipage}[t]{0.23\linewidth}
      \includegraphics[width=\linewidth]{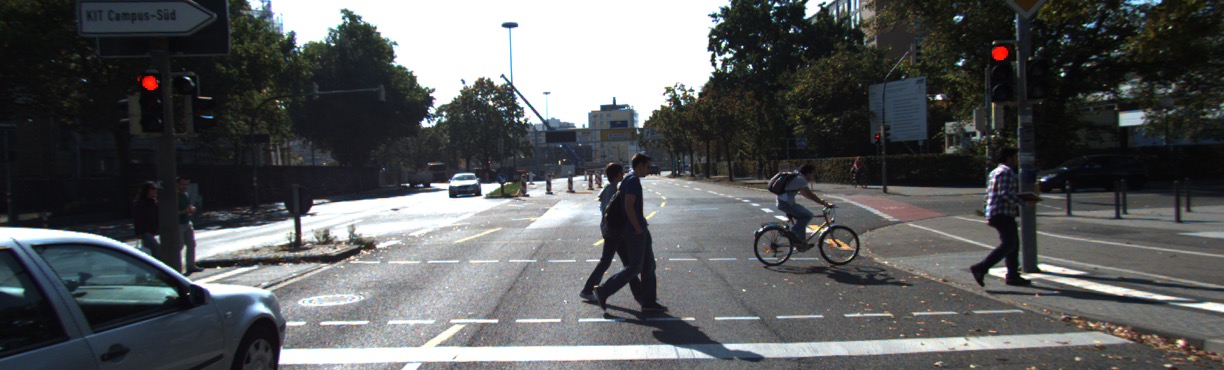}\vspace{0.1cm}
      \includegraphics[width=\linewidth]{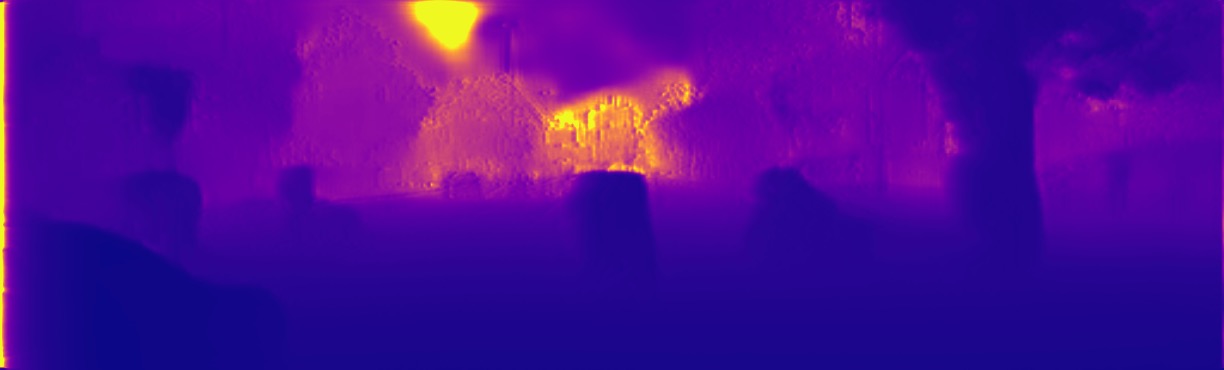}\vspace{0.1cm}
      \includegraphics[width=\linewidth]{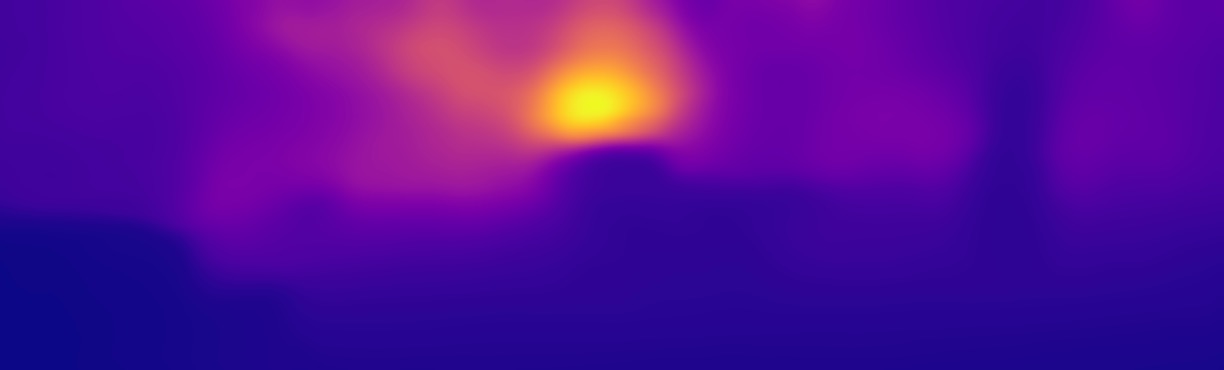}\vspace{0.1cm}
      \includegraphics[width=\linewidth]{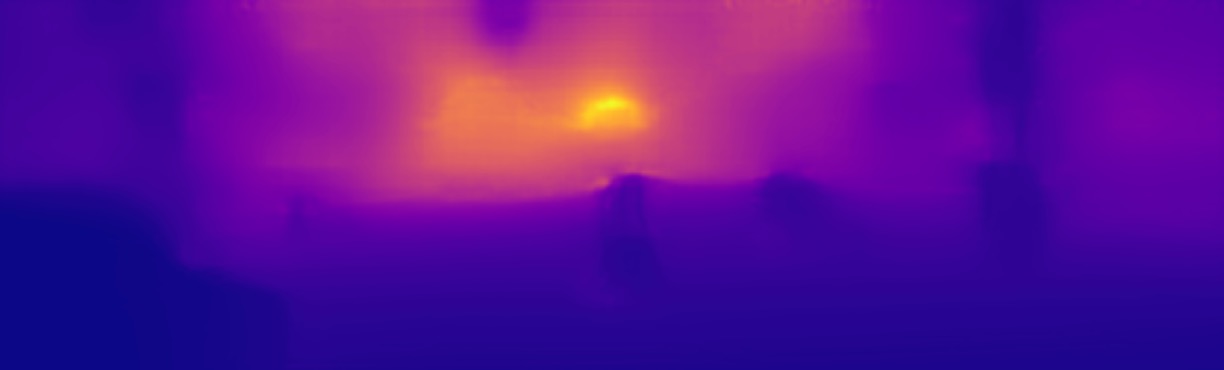}\vspace{0.1cm}
      \includegraphics[width=\linewidth]{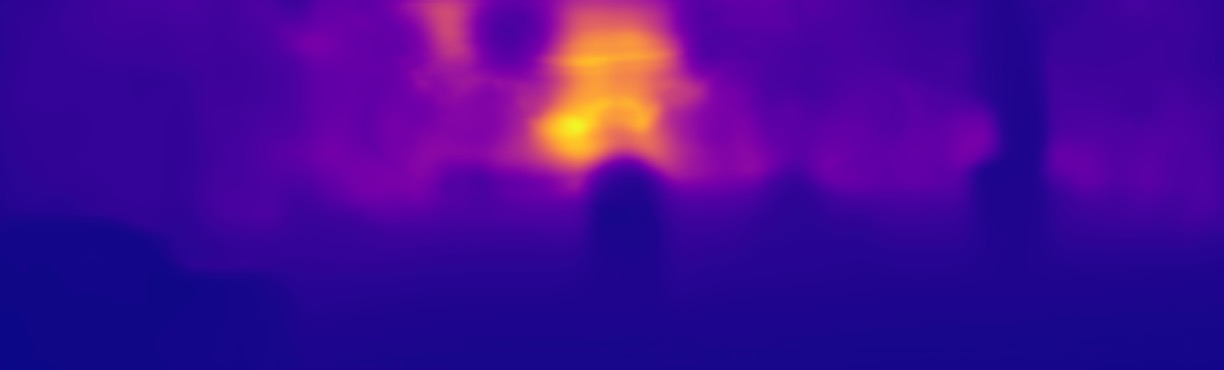}\vspace{0.1cm}
      \includegraphics[width=\linewidth]{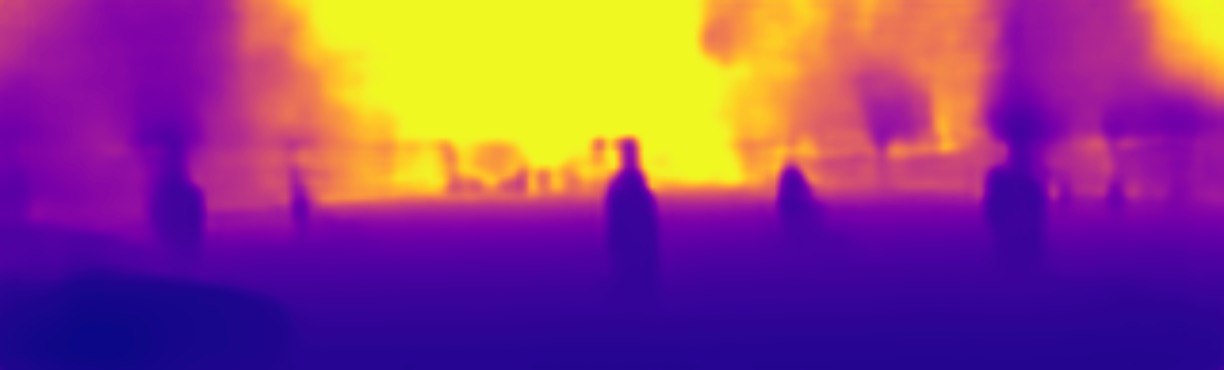}\vspace{0.1cm}
      \includegraphics[width=\linewidth]{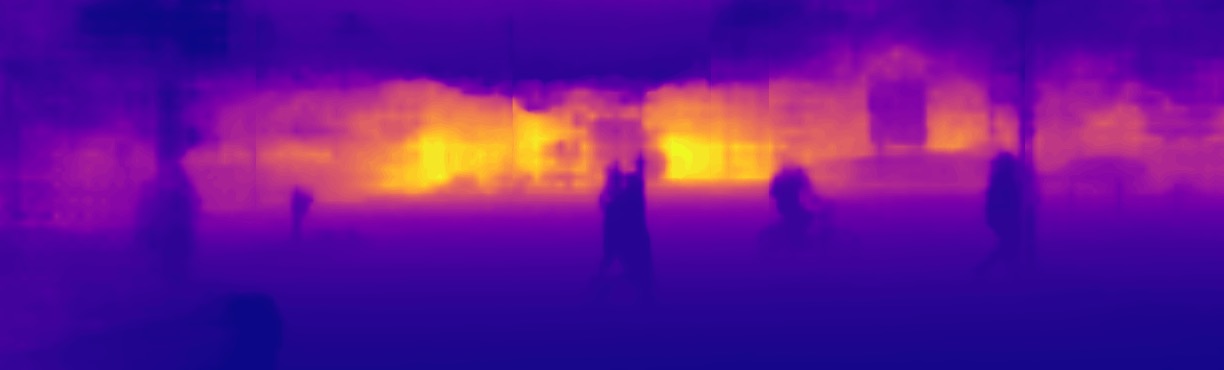}\vspace{0.1cm}
      \includegraphics[width=\linewidth]{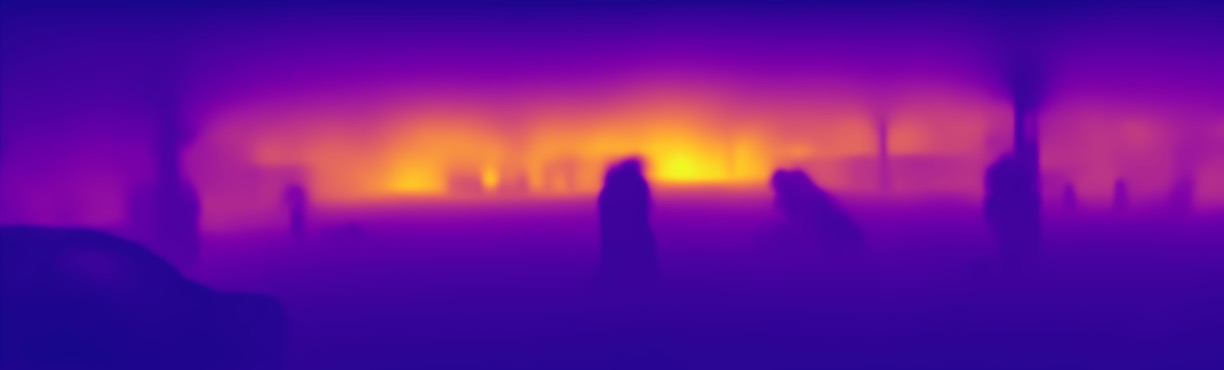}\vspace{0.1cm}
      \includegraphics[width=\linewidth]{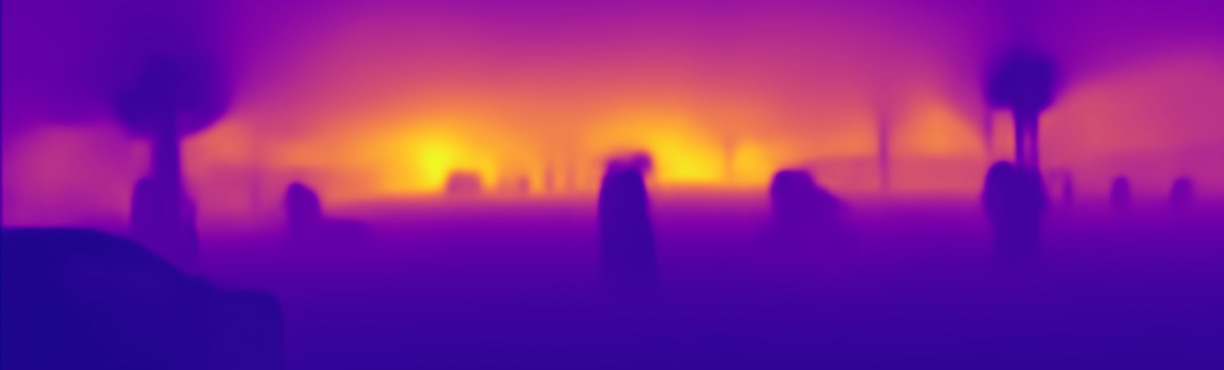}\vspace{0.1cm}
      \includegraphics[width=\linewidth]{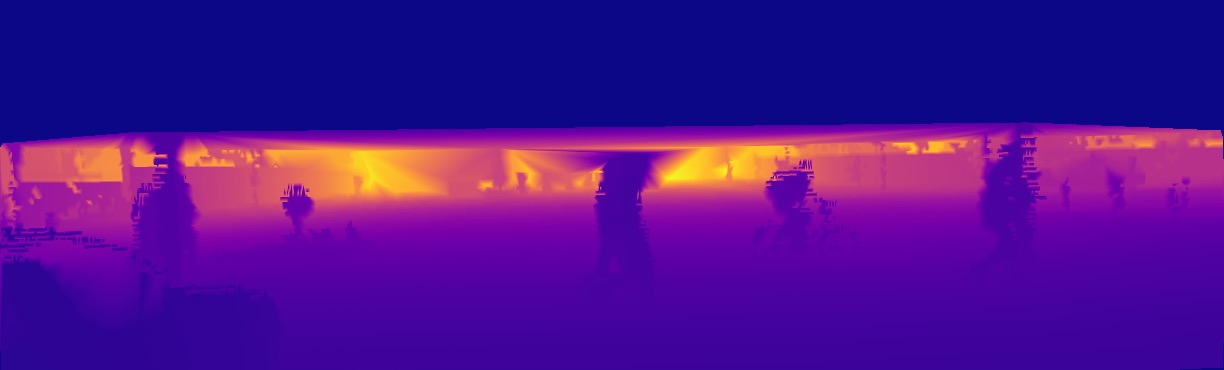}\vspace{0.1cm}
    \end{minipage}
    }\hspace{-0.3cm}
\caption{Visual comparison of predicted depth on the KITTI dataset~\cite{geiger2013vision}. Top to bottom: Video frames, depth images predicted by Godard~\emph{et al.}~\cite{godard2017unsupervised}, Zhou~\emph{et al.}~\cite{zhou2017unsupervised}, Wang~\emph{et al}.~\cite{wang2018learning}, Yin~\emph{et al.}~\cite{yin2018geonet}, Kuznietsov~\emph{et al}.~\cite{kuznietsov2017semi}, Fu~\emph{et al}.~\cite{fu2018deep} and our models~(Ours and Ours-CS+ft-K), and ground truth. We interpolate sparse ground-truth depth maps for the purpose of visualization only. Our method predicts depth for small-size or occluded objects~(e.g., thin poles and occluded cars on the bottom left of images) and provides a sharp depth transition without artifacts. (Best viewed in color.)}
  \vspace{-0.2cm}  
  \label{fig:result}
\end{figure*}

  \subsubsection{Comparison with the state of the art}

    We compare in Table~\ref{tabl:benchmark} our models with the state of the art on the test split of~\cite{eigen2014depth} in terms of prediction accuracy and runtime. We denote by ``K'', ``CS'', and ``I'' the KITTI~\cite{geiger2013vision,eigen2014depth}, Cityscapes~\cite{Cordts2016Cityscapes} and ImageNet~\cite{deng2009imagenet} datasets, respectively. Numbers in bold indicate the best performance and underscored ones are the second best among monocular depth prediction methods. Following the experimental protocol in~\cite{eigen2014depth}, we use standard metrics to measure depth prediction accuracy. The results for the comparison, except~\cite{eigen2014depth,liu2014discrete,cs2018depthnet}, have been obtained from models provided by the authors. The runtime is measured with a Nvidia GTX Titan X. From this table, we observe three things:~(1)~Our model trained on the KITTI dataset~(``Ours") achieves comparable or better performance than others in terms of depth prediction accuracy. In particular, it gives results comparable to~\cite{kuznietsov2017semi,fu2018deep}, even without using ResNet features~\cite{he2016deep} trained for ImageNet classification~\cite{kuznietsov2017semi,fu2018deep}, and exploiting stereo images for training~\cite{kuznietsov2017semi}; (2)~Our method benefits from using additional training samples. We fine-tune our model trained with the Cityscapes~\cite{Cordts2016Cityscapes} using the KITTI dataset~(``Ours-CS+ft-K"), boosting the performance and outperforming the state of the art;~(3) Our models show a good trade-off between runtime and depth prediction accuracy. They outperform other state-of-the-art methods, expect~\cite{fu2018deep}, in terms of accuracy with a small loss of speed. Our models are slightly outperformed by Fu~\emph{et al.}~\cite{fu2018deep} in terms of accuracy, but with significantly faster overall speed~(0.13 vs 1.08 seconds).  

\begin{table}[t]
  \renewcommand{\arraystretch}{1.0}
\caption{Quantitive comparison with the state of the art on the test split provided by~\cite{eigen2014depth} in terms of the average TDT.}
  \label{tabl:TDT}
  \Large
  \centering
  \resizebox{0.95\linewidth}{!}{
  \begin{tabular}{l c c c c}
    \hline
    \hline
     & lower is better & \multicolumn{3}{c}{\centering higher is better} \\ 
    \cmidrule(lr){2-2} \cmidrule(lr){3-5}
    {\centering Method} & TDT & TDT $<1$ & TDT $<2$ & TDT $<3$ \\
    \hline
    Godard~\emph{et al.}~\cite{godard2017unsupervised} & 2.964 & 0.759 & 0.856 & 0.898 \\
    Zhou~\emph{et al.}~\cite{zhou2017unsupervised} & 1.578 & 0.786 & 0.893 & 0.935 \\
    Wang~\emph{et al.}~\cite{wang2018learning} & 1.251 & 0.809 & 0.914 & 0.951 \\
    Yin~\emph{et al.}~\cite{yin2018geonet} & 1.651 & 0.791 & 0.894 & 0.932 \\
    Kuznietsov~\emph{et al.}~\cite{kuznietsov2017semi} & 1.335 & 0.805 & 0.907 & 0.947 \\
    Fu~\emph{et al.}~\cite{fu2018deep} & 1.049 & 0.827 & 0.932 & \underline{0.966} \\
    Ours & \underline{0.940} & \underline{0.835} & \underline{0.951} & \textbf{0.979} \\
    Ours-CS+ft-K & \textbf{0.896} & \textbf{0.848} & \textbf{0.952} & \textbf{0.979} \\
    \hline
    Ground truth & 0.712 & 0.924 & 0.982 & 0.989 \\
    \hline
    \hline
  \end{tabular}
  }
\vspace{-0.3cm}
\end{table}

\begin{figure*}[t]
  \centering
  \renewcommand*{\thesubfigure}{}
  \subfigure[Godard~\emph{et al.}~\cite{godard2017unsupervised}]{
    \begin{minipage}[t]{0.23\linewidth}
      \includegraphics[width=\linewidth]{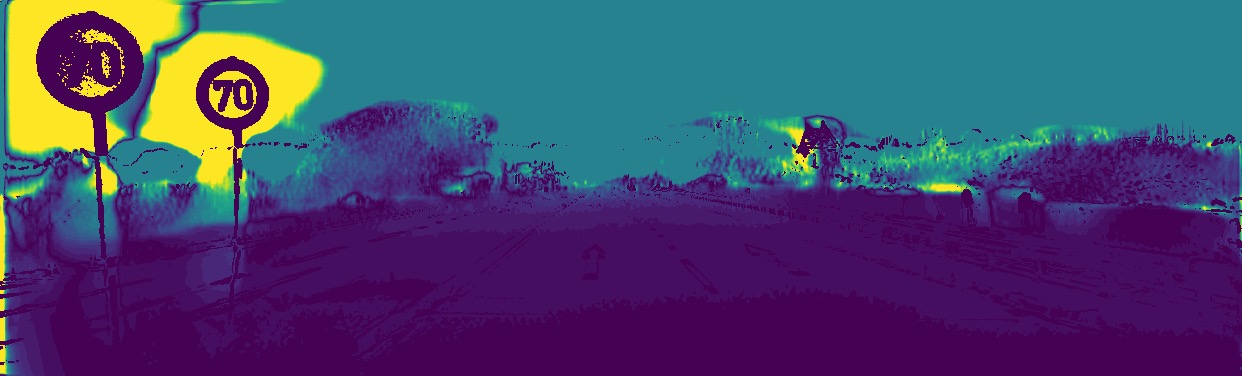}\vspace{0.1cm}
      \includegraphics[width=\linewidth]{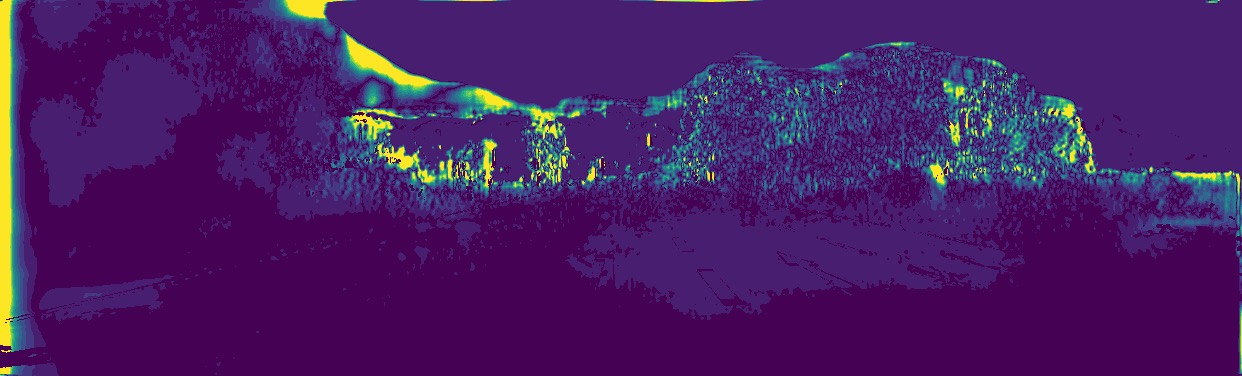}\vspace{0.1cm}
    \end{minipage}
    }\hspace{-0.3cm}
  \subfigure[Zhou~\emph{et al.}~\cite{zhou2017unsupervised}]{
    \begin{minipage}[t]{0.23\linewidth}
      \includegraphics[width=\linewidth]{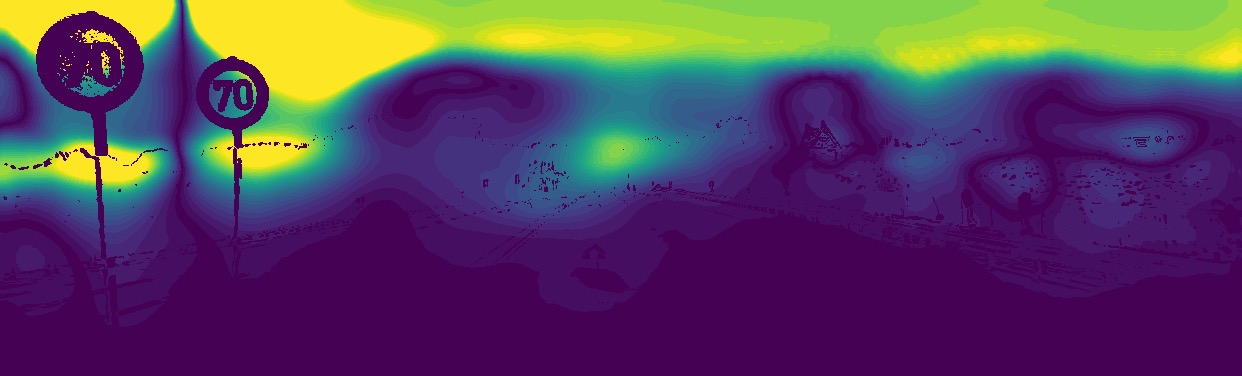}\vspace{0.1cm}
      \includegraphics[width=\linewidth]{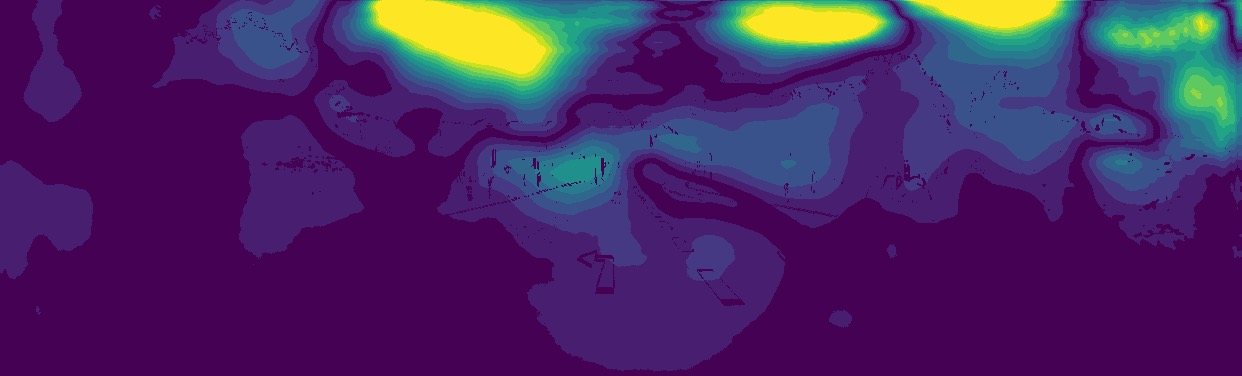}\vspace{0.1cm}
    \end{minipage}
    }\hspace{-0.3cm}
  \subfigure[Wang~\emph{et al.}~\cite{wang2018learning}]{
    \begin{minipage}[t]{0.23\linewidth}
      \includegraphics[width=\linewidth]{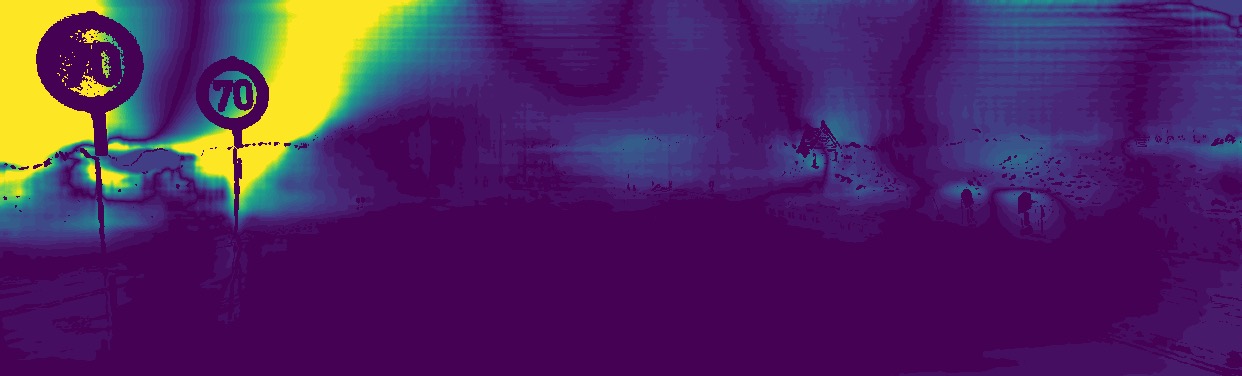}\vspace{0.1cm}
      \includegraphics[width=\linewidth]{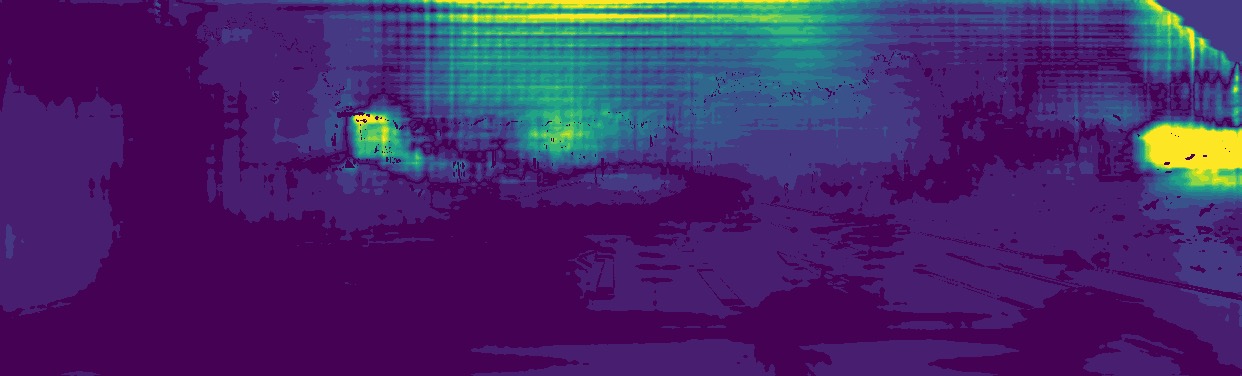}\vspace{0.1cm}
    \end{minipage}
    }\hspace{-0.3cm}
  \subfigure[Yin~\emph{et al.}~\cite{yin2018geonet}]{
    \begin{minipage}[t]{0.23\linewidth}
      \includegraphics[width=\linewidth]{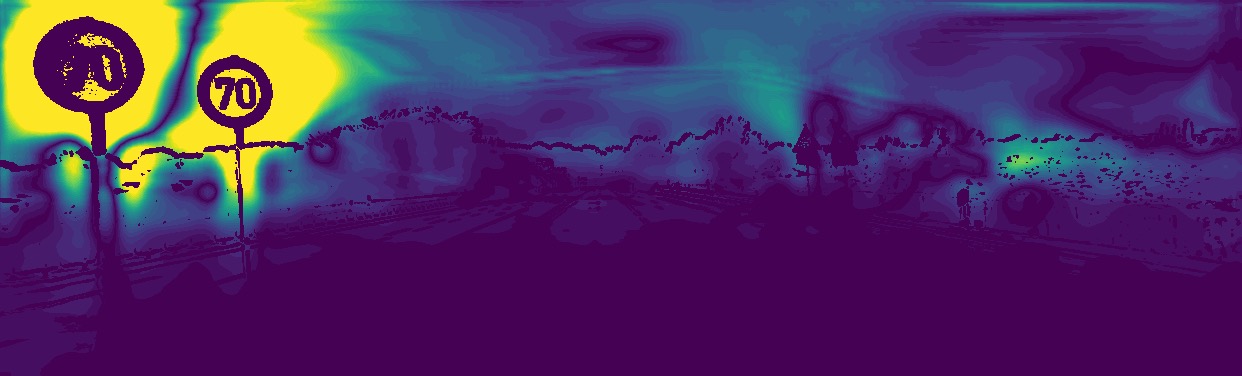}\vspace{0.1cm}
      \includegraphics[width=\linewidth]{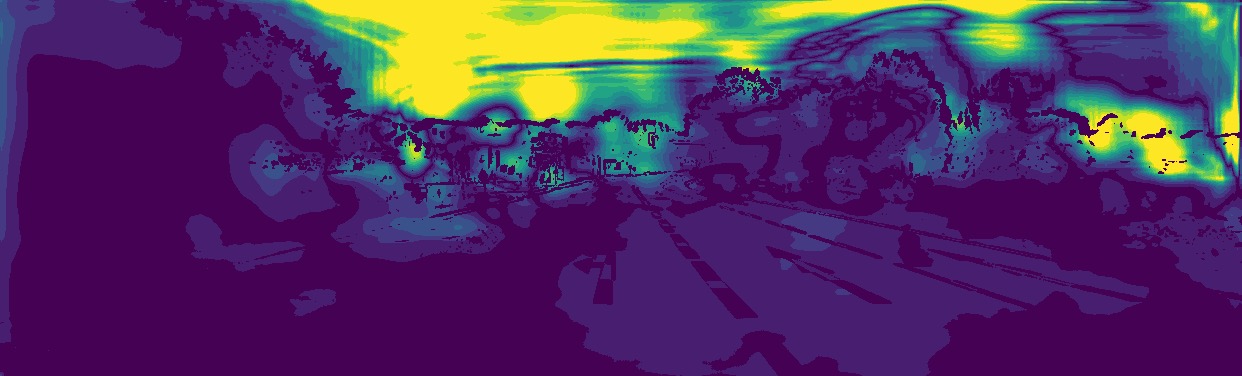}\vspace{0.1cm}
    \end{minipage}
    \hspace{-0.25cm}
    }\hspace{-0.3cm}\\
  \vspace{-0.2cm}
  \subfigure[Kuznietsov~\emph{et al.}~\cite{kuznietsov2017semi}]{
    \hspace{-0.17cm}
    \begin{minipage}[t]{0.23\linewidth}
      \includegraphics[width=\linewidth]{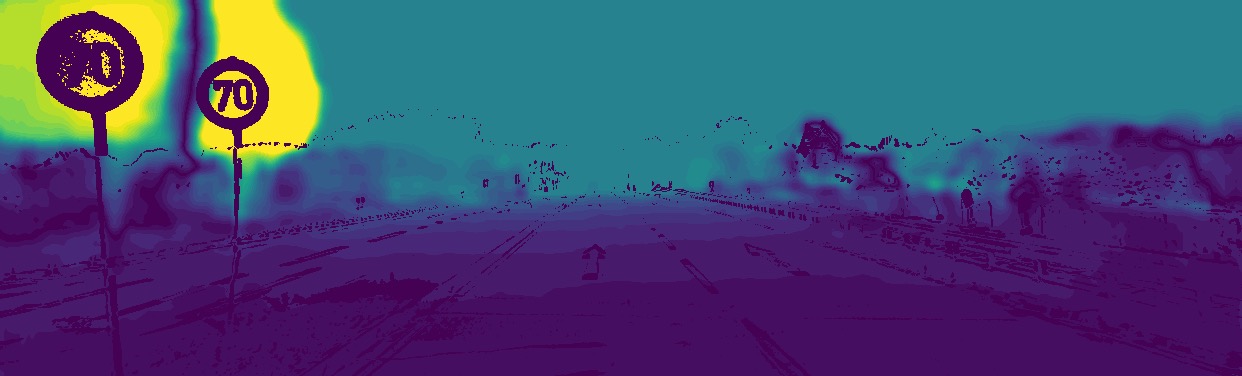}\vspace{0.1cm}
      \includegraphics[width=\linewidth]{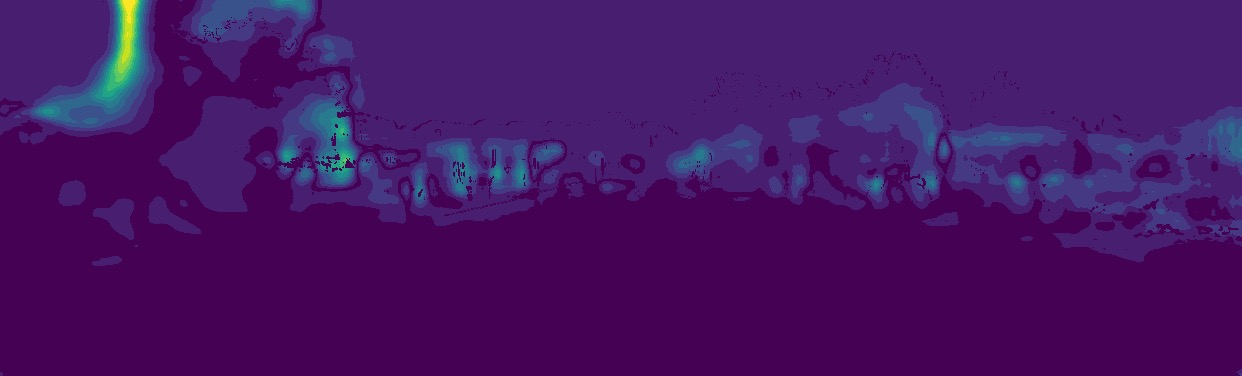}\vspace{0.1cm}
    \end{minipage}
    }\hspace{-0.3cm}
  \subfigure[Fu~\emph{et al.}~\cite{fu2018deep}]{
    \begin{minipage}[t]{0.23\linewidth}
      \includegraphics[width=\linewidth]{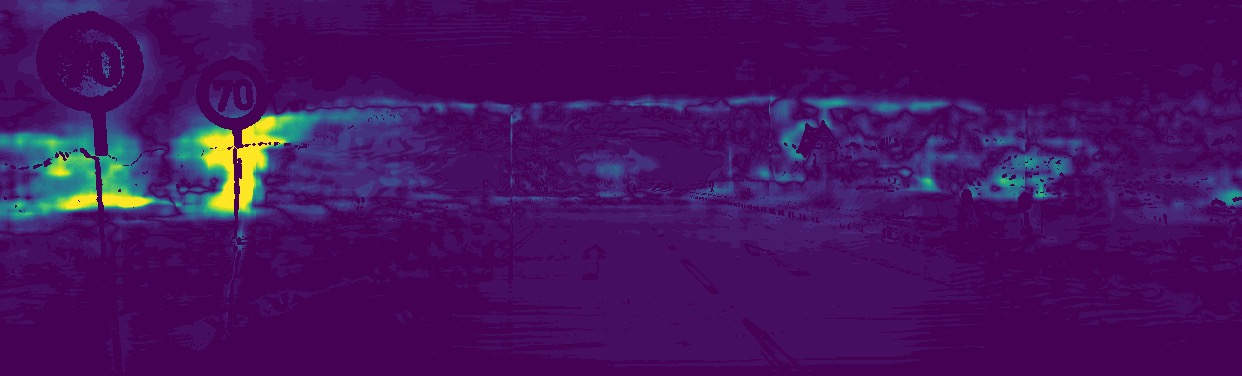}\vspace{0.1cm}
      \includegraphics[width=\linewidth]{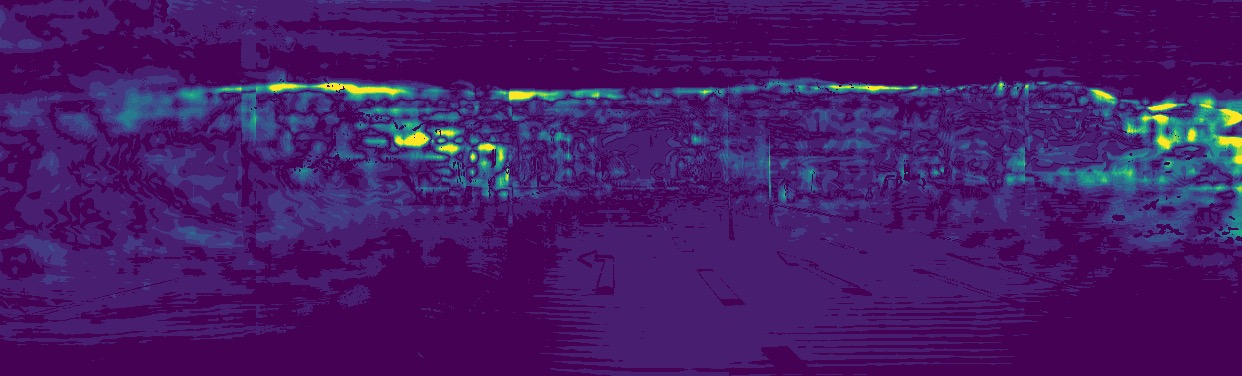}\vspace{0.1cm}
    \end{minipage}
    }\hspace{-0.3cm}
  \subfigure[Ours-CS+ft-K]{
    \begin{minipage}[t]{0.23\linewidth}
      \includegraphics[width=\linewidth]{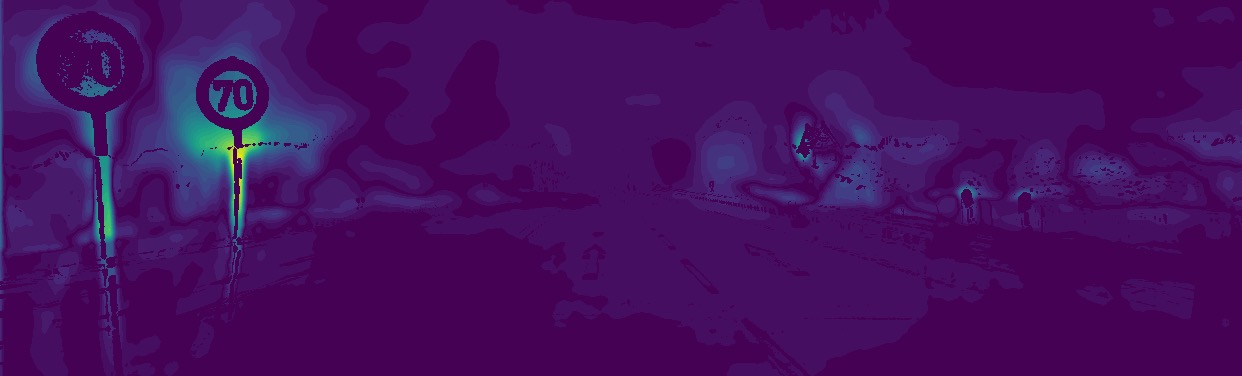}\vspace{0.1cm}
      \includegraphics[width=\linewidth]{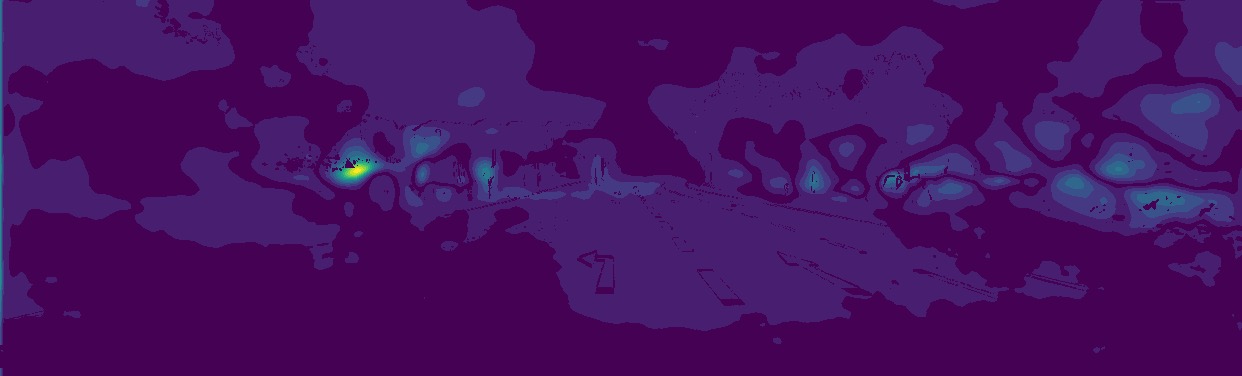}\vspace{0.1cm}
    \end{minipage}
    }\hspace{-0.3cm}
  \subfigure[Ground truth]{
    \begin{minipage}[t]{0.23\linewidth}
      \includegraphics[width=\linewidth]{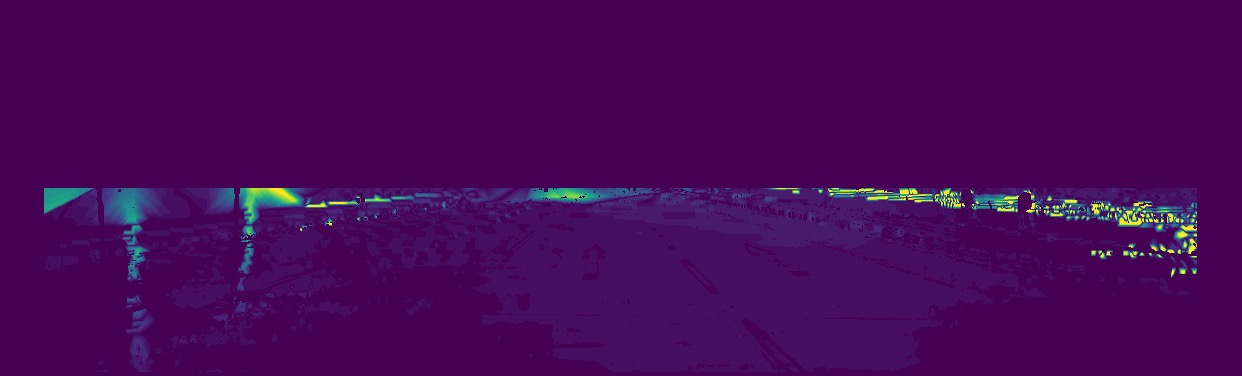}\vspace{0.1cm}
      \includegraphics[width=\linewidth]{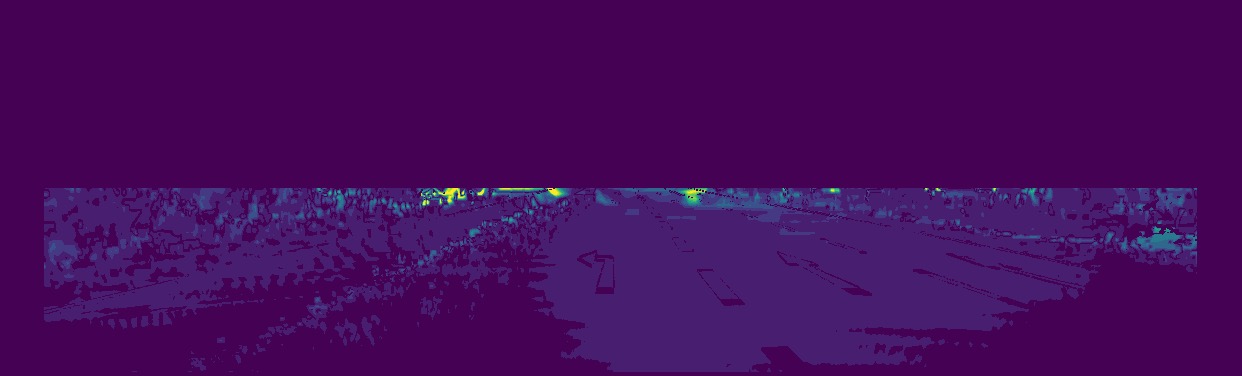}\vspace{0.1cm}
    \end{minipage}
    }\hspace{-0.3cm}
 \vspace{-0.2cm}
\caption{Visual comparison of pixel-wise TDT scores. Two examples are shown for each method. The TDT scores are color-coded (blue: low, yellow: high). Our model shows lower TDT scores than the state of the art~\cite{godard2017unsupervised,zhou2017unsupervised,wang2018learning,yin2018geonet,kuznietsov2017semi,fu2018deep}, especially for the regions near objects, demonstrating that it gives temporally consistent results. (Best viewed in color.)}
  \vspace{-0.3cm}  
  \label{fig:temp_result}
\end{figure*}

	\begin{figure*}[t]
	  \centering
	  \renewcommand*{\thesubfigure}{}
	  \subfigure[(a)]{
	    \begin{minipage}[t]{0.23\linewidth}
	      \includegraphics[height=0.33\linewidth, width=\linewidth]{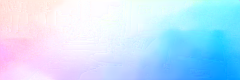}\vspace{0.065cm}
	      \includegraphics[height=0.33\linewidth, width=\linewidth]{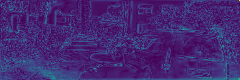}\vspace{0.3cm}
	    \end{minipage}
	    }\hspace{-0.3cm}
	  \subfigure[(b)]{
	    \begin{minipage}[t]{0.23\linewidth}
	      \includegraphics[height=0.33\linewidth, width=\linewidth]{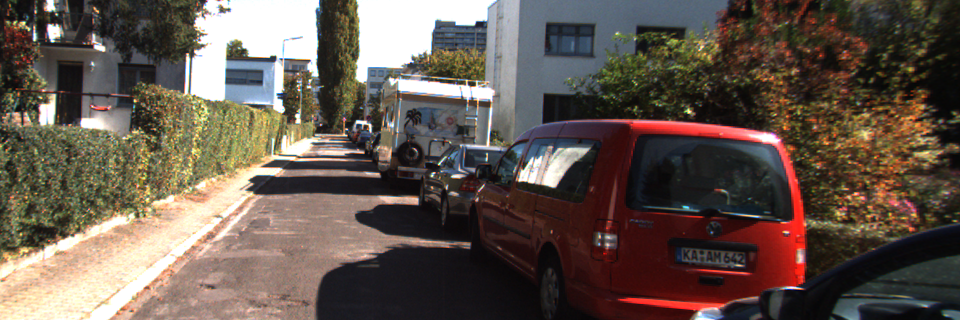}\vspace{0.065cm}
	      \includegraphics[height=0.33\linewidth, width=\linewidth]{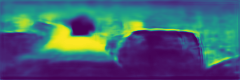}\vspace{0.3cm}
	    \end{minipage}
	    }\hspace{-0.3cm}
	  \subfigure[(c)]{
	    \begin{minipage}[t]{0.23\linewidth}
	      \includegraphics[height=0.33\linewidth, width=\linewidth]{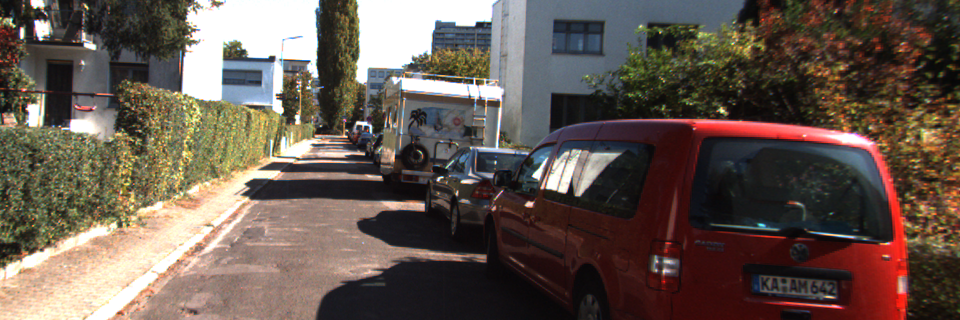}\vspace{0.065cm}
	      \includegraphics[height=0.33\linewidth, width=\linewidth]{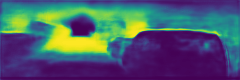}\vspace{0.3cm}
	    \end{minipage}
	    }\hspace{-0.3cm}
	  \subfigure[(d)]{
	    \begin{minipage}[t]{0.23\linewidth}
	      \includegraphics[height=0.33\linewidth, width=\linewidth]{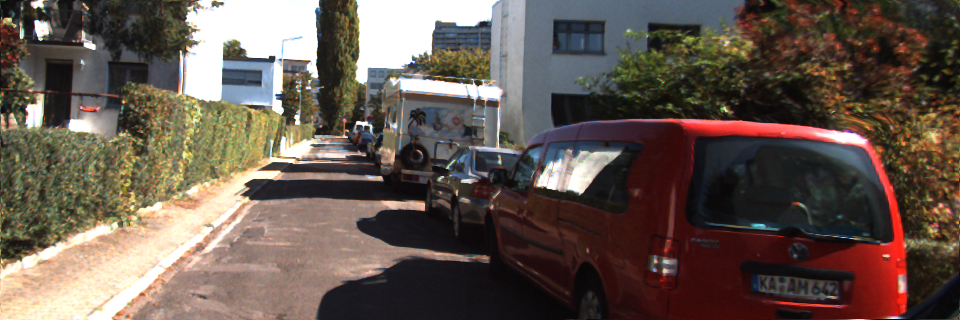}\vspace{0.065cm}
	      \includegraphics[height=0.33\linewidth, width=\linewidth]{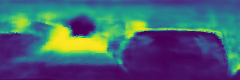}\vspace{0.3cm}
	    \end{minipage}
	    }\hspace{-0.13cm}
	  \vspace{-0.3cm}
	  \caption{Examples of a refined flow field and warping results. (a) Top to bottom: A refined flow and its difference from the input optical flow. (b-c) Top to bottom: Video frames and hidden states at time~$t-1$ and $t$, respectively. (d) A video frame and a hidden state aligned w.r.t. time~$t$ by warping using the refined flow. The refined flow captures structure details, particularly around moving objects, allowing to provide a sharp depth transition. It also aligns both video frames and hidden states well, making it possible for our model to give temporally consistent results without flickering artifacts. (Best viewed in color.)}
	  \vspace{-0.3cm}  
	  \label{fig:refinedflow}
	\end{figure*}

	\begin{figure}[t]
	  \centering
	  \vspace{0.2cm}
	  \renewcommand*{\thesubfigure}{}
	  \subfigure[(a) Cityscapes dataset]{
	    \begin{minipage}[t]{0.95\linewidth}
	    \vspace{-0.2cm}
	      \includegraphics[width=\linewidth]{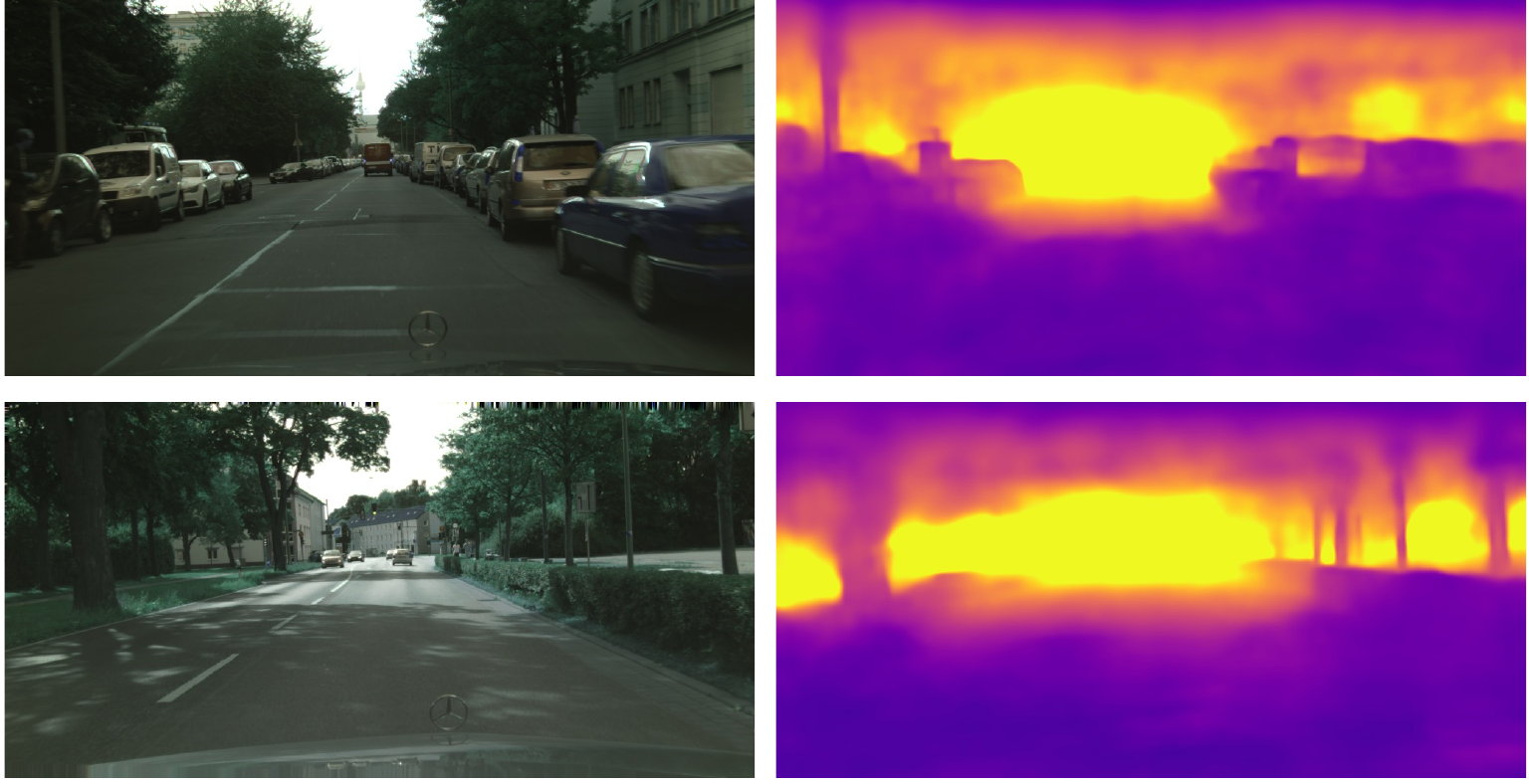}\vspace{0.065cm}
	    \end{minipage}
	    }\hspace{0.9cm}\vspace{-0.2cm}
	  
	  \subfigure[(b) NYU dataset]{
	   \hspace{-0.3cm}
	    \begin{minipage}[t]{0.95\linewidth}
	    \vspace{-0.2cm}
	    \includegraphics[width=\linewidth]{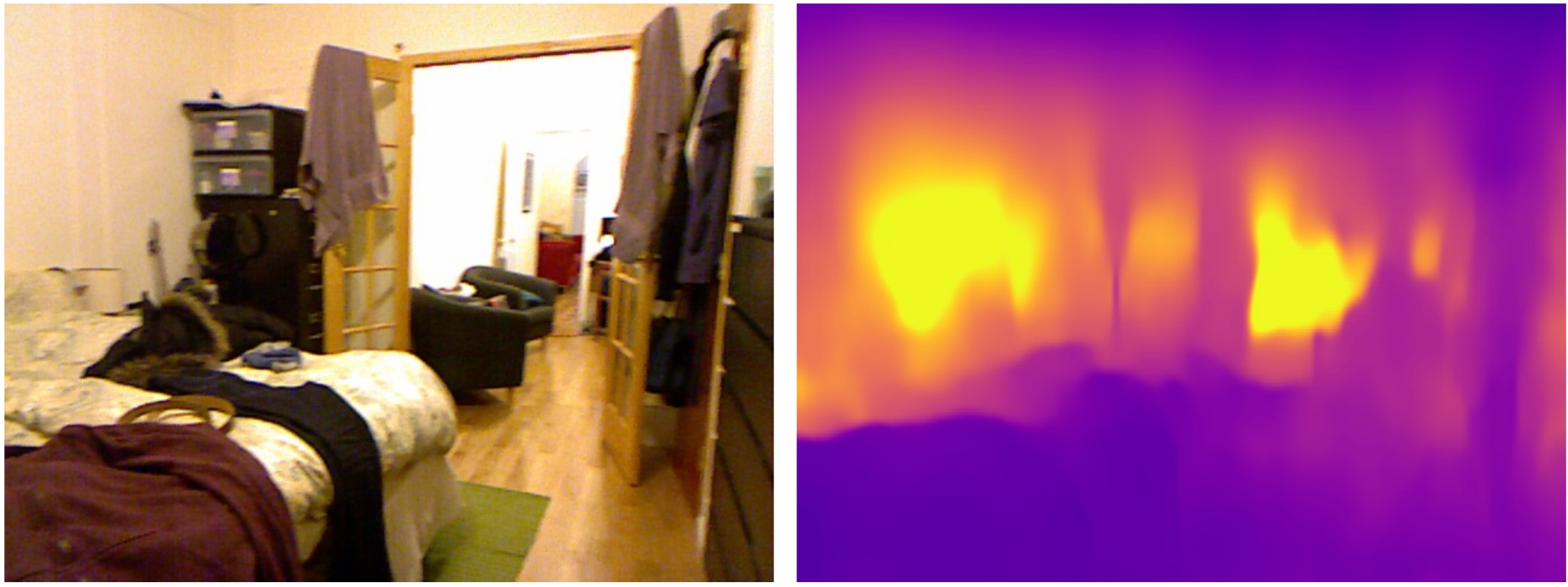}\vspace{0.065cm}
	    \end{minipage}
	    }\hspace{-0.3cm}
	  \vspace{-0.2cm}
	  \caption{Examples of predicted depth by our model on (b)the Cityscapes~\cite{Cordts2016Cityscapes} and (b)the NYU datasets~\cite{Silberman:ECCV12}. We apply our model trained with the KITTI~\cite{eigen2014depth}. The examples demonstrate that our model performs well on other images outside the training dataset.}
	  \vspace{-0.2cm}  
	  \label{fig:gen_CS}
	\end{figure}

We show in Fig.~\ref{fig:temp} an example of the TDT comparison of the state of the art and our models in the KITTI dataset~\cite{geiger2013vision}. Although Zhou~\emph{et al.}~\cite{zhou2017unsupervised} and Wang~\emph{et al.}~\cite{wang2018learning} use a video sequence as a supervisory signal similar to ours, they do not consider temporal coherence in the video, producing temporally inconsistent results. Kuznietsov~\emph{et al.}~\cite{kuznietsov2017semi} and Fu~\emph{et al.}~\cite{fu2018deep} give results comparable to ours in terms of depth prediction accuracy as shown in~Table~\ref{tabl:benchmark}, but their TDT scores are far from the ground truth. On the contrary, our models produce temporally stable and consistent results, with lower errors than the state of the art. In Table~\ref{tabl:TDT}, we show the average TDT scores on the test split of~\cite{eigen2014depth} and compare our models with the state of the art in terms of temporal consistency. Numbers in bold indicate the best performance and underscored ones are the second best. Our method outperforms the state of the art including~\cite{kuznietsov2017semi,fu2018deep} by a significant margin. For comparison, the scores computed with ground-truth depth are 0.712 for TDT, and 0.924, 0.982, 0.989 for TDT$<1$, TDT$<2$, TDT$<3$, respectively. To this end, we interpolate sparse ground-truth depth maps and discard values at highly sparse regions~(\emph{e.g.}, upper parts of images) using masks provided by~\cite{garg2016unsupervised}. Note that the better ability to give temporally consistent results by our method does not come from the use of ground-truth depth. The supervised learning approach~\cite{kuznietsov2017semi} shows much worse results than the unsupervised one~\cite{wang2018learning}, indicating that using ground truth does not always give temporally consistent results.

    \subsubsection{Qualitative results}
We show in Fig.~\ref{fig:result} a visual comparison of depth prediction results on the KITTI dataset~\cite{eigen2014depth}. We can see that our models predict a fine-grained depth~(\emph{e.g.},~for distant objects and poles) and provide a sharp depth transition without artifacts. For comparison, Fu~\emph{et al.}~\cite{fu2018deep} shows grid artifacts often caused by dilated convolutions~\cite{yu2017dilated}. We can also see that our models are highly robust to occlusion compared to other methods. For example, they predict depth from occluded cars on the bottom left of images while others are limited to handle such objects. Figure~\ref{fig:temp_result} visualizes pixel-wise TDT scores. We show temporal differences~$\lVert D^t(p) - \bar{D}^t(p) \rVert_1$, weighted by the confidence map~$C^t(p)$, between predicted depth maps. It shows that our model gives temporally consistent results, especially for regions having large displacements~(\emph{e.g.},~traffic signs), resulting in less flickering artifacts.

    \subsubsection{Refined optical flow}
In Fig.~\ref{fig:refinedflow}(a), we show an example of the refined flow field and its difference from the input flow. We can see that the flow refine network modifies the input flow, particularly around moving objects, making it possible to capture fine details while preserving edges and object boundaries. Our model uses the refined flow to align video frames and hidden states in the visual memory. We show video frames and hidden states at time~$t-1$ and $t$ in Figs.~\ref{fig:refinedflow}(b-c), respectively. Warping results w.r.t. time~$t$ using the refined flow are shown in Fig.~\ref{fig:refinedflow}(d). By comparing Figs.~\ref{fig:refinedflow}(c) and (d), we can see that the refined flow aligns both the video frame and the hidden state well, which enables our model to aggregate temporally aligned features and to prevent flickering artifacts.  

	\subsubsection{Generalization to other dataset}
We test our model trained with the KITTI~\cite{eigen2014depth} on the Cityscapes~\cite{Cordts2016Cityscapes} and the NYU~\cite{Silberman:ECCV12} datasets to demonstrate its generalization ability. Examples shown in Fig.~\ref{fig:gen_CS} demonstrate that our model generalizes well to other images outside the training dataset. Particularly, it infers both a geometric layout in a scene and object instances~(\emph{e.g.},~cars and trees in Fig.~\ref{fig:gen_CS}(a) and a bed in Fig.~\ref{fig:gen_CS}(b)) well. Note that, for the Cityscapes and the NYU datasets, all previous works we are aware of~(\emph{e.g.}, \cite{zhou2017unsupervised,wang2018learning,godard2017unsupervised,kuznietsov2017semi,fu2018deep,yin2018geonet}) offer qualitative results only.

\section{Conclusion}

We have presented a recurrent network for monocular depth prediction that gives temporally consistent results while preserving depth boundaries. Particularly, we have introduced a flow-guided memory module that selectively retains hidden states aligned along motion trajectories, enforcing a long-term temporal consistency to prediction results. We have also presented a flow refine network that outputs dense flow fields specific to our task. We have shown that the refined flow aligns both video frames and hidden states, preventing flickering artifacts. We have demonstrated that our method outperforms the state of the art by a large margin in terms of temporal consistency, shows a good trade-off between depth prediction accuracy and runtime, and performs well on other images outside training datasets. 

\ifCLASSOPTIONcaptionsoff
  \newpage
\fi

\bibliographystyle{IEEEtran}
\bibliography{egbib}

\begin{thebibliography}{10}
\providecommand{\url}[1]{#1}
\csname url@samestyle\endcsname
\providecommand{\newblock}{\relax}
\providecommand{\bibinfo}[2]{#2}
\providecommand{\BIBentrySTDinterwordspacing}{\spaceskip=0pt\relax}
\providecommand{\BIBentryALTinterwordstretchfactor}{4}
\providecommand{\BIBentryALTinterwordspacing}{\spaceskip=\fontdimen2\font plus
\BIBentryALTinterwordstretchfactor\fontdimen3\font minus
  \fontdimen4\font\relax}
\providecommand{\BIBforeignlanguage}[2]{{%
\expandafter\ifx\csname l@#1\endcsname\relax
\typeout{** WARNING: IEEEtran.bst: No hyphenation pattern has been}%
\typeout{** loaded for the language `#1'. Using the pattern for}%
\typeout{** the default language instead.}%
\else
\language=\csname l@#1\endcsname
\fi
#2}}
\providecommand{\BIBdecl}{\relax}
\BIBdecl

\bibitem{chen2017coherent}
D.~Chen, J.~Liao, L.~Yuan, N.~Yu, and G.~Hua, ``Coherent online video style
  transfer,'' in \emph{Proc. Intl. Conf. Computer Vision}, 2017.

\bibitem{keller2011benefits}
C.~G. Keller, M.~Enzweiler, M.~Rohrbach, D.~F. Llorca, C.~Schnorr, and D.~M.
  Gavrila, ``The benefits of dense stereo for pedestrian detection,''
  \emph{IEEE Trans. Intell. Transp. Syst.}, vol.~12, no.~4, pp. 1096--1106,
  2011.

\bibitem{li2018traffic}
L.~Li, B.~Qian, J.~Lian, W.~Zheng, and Y.~Zhou, ``Traffic scene segmentation
  based on rgb-d image and deep learning,'' \emph{IEEE Trans. Intell. Transp.
  Syst.}, vol.~19, no.~5, pp. 1664--1669, 2018.

\bibitem{wu2017geometry}
P.~Wu, Y.~Liu, M.~Ye, Z.~Xu, and Y.~Zheng, ``Geometry guided multi-scale depth
  map fusion via graph optimization,'' \emph{IEEE Trans. Image Process},
  vol.~26, no.~3, pp. 1315--1329, 2017.

\bibitem{nguyen2017robust}
V.~D. Nguyen, H.~Van~Nguyen, and J.~W. Jeon, ``Robust stereo data cost with a
  learning strategy,'' \emph{IEEE Trans. Intell. Transp. Syst.}, vol.~18,
  no.~2, pp. 248--258, 2017.

\bibitem{van2006real}
W.~Van Der~Mark and D.~M. Gavrila, ``Real-time dense stereo for intelligent
  vehicles,'' \emph{IEEE Trans. Intell. Transp. Syst.}, vol.~7, no.~1, pp.
  38--50, 2006.

\bibitem{li2008binocular}
S.~Li, ``Binocular spherical stereo,'' \emph{IEEE Trans. Intell. Transp.
  Syst.}, vol.~9, no.~4, pp. 589--600, 2008.

\bibitem{muresan2017mutlipatch}
M.~P. Muresan, S.~Nedevschi, and R.~Danescu, ``A multi patch warping approach
  for improved stereo block matching,'' in \emph{Proc. Int. Conf. Comput. Vis.
  Theory App.}, 2017.

\bibitem{hirschmuller2007stereo}
H.~Hirschmuller, ``Stereo processing by semiglobal matching and mutual
  information,'' \emph{IEEE Trans. Pattern Anal. Mach. Intell.}, vol.~30,
  no.~2, pp. 328--341, 2007.

\bibitem{miclea2018real}
V.-C. Miclea, L.~Miclea, and S.~Nedevschi, ``Real-time stereo reconstruction
  failure detection and correction using deep learning,'' in \emph{IEEE Int.
  Conf. on Intell. Transp. Syst.}, 2018, pp. 1095--1102.

\bibitem{spangenberg2014large}
R.~Spangenberg, T.~Langner, S.~Adfeldt, and R.~Rojas, ``Large scale semi-global
  matching on the cpu,'' in \emph{Proc. IEEE Intell. Veh. Symp.}, 2014, pp.
  195--201.

\bibitem{eigen2014depth}
D.~Eigen, C.~Puhrsch, and R.~Fergus, ``Depth map prediction from a single image
  using a multi-scale deep network,'' in \emph{Proc. Int. Conf. Adv. Neural
  Inf. Process. Syst.}, 2014, pp. 2366--2374.

\bibitem{liu2014discrete}
M.~Liu, M.~Salzmann, and X.~He, ``Discrete-continuous depth estimation from a
  single image,'' in \emph{Proc. IEEE Conf. Comput. Vis. Pattern Recog.}, 2014,
  pp. 716--723.

\bibitem{zhou2017unsupervised}
T.~Zhou, M.~Brown, N.~Snavely, and D.~G. Lowe, ``Unsupervised learning of depth
  and ego-motion from video,'' in \emph{Proc. IEEE Conf. Comput. Vis. Pattern
  Recog.}, vol.~2, no.~6, 2017, p.~7.

\bibitem{wang2018learning}
C.~Wang, J.~M. Buenaposada, R.~Zhu, and S.~Lucey, ``Learning depth from
  monocular videos using direct methods,'' in \emph{Proc. IEEE Conf. Comput.
  Vis. Pattern Recog.}, 2018, pp. 2022--2030.

\bibitem{godard2017unsupervised}
C.~Godard, O.~Mac~Aodha, and G.~J. Brostow, ``Unsupervised monocular depth
  estimation with left-right consistency,'' in \emph{Proc. IEEE Conf. Comput.
  Vis. Pattern Recog.}, vol.~2, no.~6, 2017, p.~7.

\bibitem{kuznietsov2017semi}
Y.~Kuznietsov, J.~St{\"u}ckler, and B.~Leibe, ``Semi-supervised deep learning
  for monocular depth map prediction,'' in \emph{Proc. IEEE Conf. Comput. Vis.
  Pattern Recog.}, 2017, pp. 6647--6655.

\bibitem{cs2018depthnet}
A.~CS~Kumar, S.~M. Bhandarkar, and M.~Prasad, ``Depth{N}et: A recurrent neural
  network architecture for monocular depth prediction,'' in \emph{Proc. IEEE
  Conf. Comput. Vis. Pattern Recog. Workshop}, 2018, pp. 283--291.

\bibitem{fu2018deep}
H.~Fu, M.~Gong, C.~Wang, K.~Batmanghelich, and D.~Tao, ``Deep ordinal
  regression network for monocular depth estimation,'' in \emph{Proc. IEEE
  Conf. Comput. Vis. Pattern Recog.}, 2018, pp. 2002--2011.

\bibitem{rogers1979motion}
B.~Rogers and M.~Graham, ``Motion parallax as an independent cue for depth
  perception,'' \emph{Perception}, vol.~8, no.~2, pp. 125--134, 1979.

\bibitem{du2018recurrent}
W.~Du, Y.~Wang, and Y.~Qiao, ``Recurrent spatial-temporal attention network for
  action recognition in videos,'' \emph{IEEE Trans. Image Process}, vol.~27,
  no.~3, pp. 1347--1360, 2018.

\bibitem{zhang2017context}
B.~Zhang, D.~Xiong, J.~Su, and H.~Duan, ``A context-aware recurrent encoder for
  neural machine translation,'' \emph{ACM Trans. Audio, Speech, Language
  Process.}, vol.~25, no.~12, pp. 2424--2432, 2017.

\bibitem{shi2017deep}
X.~Shi, Z.~Gao, L.~Lausen, H.~Wang, D.-Y. Yeung, W.-k. Wong, and W.-c. Woo,
  ``Deep learning for precipitation nowcasting: A benchmark and a new model,''
  in \emph{Proc. Int. Conf. Adv. Neural Inf. Process. Syst.}, 2017, pp.
  5617--5627.

\bibitem{ballas2015delving}
N.~Ballas, L.~Yao, C.~Pal, and A.~Courville, ``Delving deeper into
  convolutional networks for learning video representations,'' \emph{Proc. Int.
  Conf. Learning Representations}, 2016.

\bibitem{simonyan2014two}
K.~Simonyan and A.~Zisserman, ``Two-stream convolutional networks for action
  recognition in videos,'' in \emph{Proc. Int. Conf. Adv. Neural Inf. Process.
  Syst.}, 2014, pp. 568--576.

\bibitem{geiger2013vision}
A.~Geiger, P.~Lenz, C.~Stiller, and R.~Urtasun, ``Vision meets robotics: The
  {KITTI} dataset,'' \emph{Int. J. Rob. Res.}, vol.~32, no.~11, pp. 1231--1237,
  2013.

\bibitem{lowe2004distinctive}
D.~G. Lowe, ``Distinctive image features from scale-invariant keypoints,''
  \emph{Int. J. Comput. Vis.}, vol.~60, no.~2, pp. 91--110, 2004.

\bibitem{dalal2005histograms}
N.~Dalal and B.~Triggs, ``Histograms of oriented gradients for human
  detection,'' in \emph{Proc. IEEE Conf. Comput. Vis. Pattern Recog.}, 2005.

\bibitem{saxena2006learning}
A.~Saxena, S.~H. Chung, and A.~Y. Ng, ``Learning depth from single monocular
  images,'' in \emph{Proc. Int. Conf. Adv. Neural Inf. Process. Syst.}, 2006,
  pp. 1161--1168.

\bibitem{karsch2014depth}
K.~Karsch, C.~Liu, and S.~B. Kang, ``Depth transfer: Depth extraction from
  video using non-parametric sampling,'' \emph{IEEE Trans. Pattern Anal. Mach.
  Intell.}, vol.~36, no.~11, pp. 2144--2158, 2014.

\bibitem{liu2011sift}
C.~Liu, J.~Yuen, and A.~Torralba, ``{SIFT} {F}low: Dense correspondence across
  scenes and its applications,'' \emph{IEEE Trans. Pattern Anal. Mach.
  Intell.}, vol.~33, no.~5, pp. 978--994, 2011.

\bibitem{liu2015deep}
F.~Liu, C.~Shen, and G.~Lin, ``Deep convolutional neural fields for depth
  estimation from a single image,'' in \emph{Proc. IEEE Conf. Comput. Vis.
  Pattern Recog.}, 2015, pp. 5162--5170.

\bibitem{yin2018geonet}
Z.~Yin and J.~Shi, ``Geo{N}et: Unsupervised learning of dense depth, optical
  flow and camera pose,'' in \emph{Proc. IEEE Conf. Comput. Vis. Pattern
  Recog.}, vol.~2, 2018.

\bibitem{steinbrucker2011real}
F.~Steinbr{\"u}cker, J.~Sturm, and D.~Cremers, ``Real-time visual odometry from
  dense {RGB}-{D} images,'' in \emph{Proc. Int. Conf. Comput. Vis. Workshops},
  2011, pp. 719--722.

\bibitem{hopfield1982neural}
J.~J. Hopfield, ``Neural networks and physical systems with emergent collective
  computational abilities,'' \emph{Proc. Natl. Acad. Sci. U.S.A.}, vol.~79,
  no.~8, pp. 2554--2558, 1982.

\bibitem{rumelhart1986learning}
D.~E. Rumelhart, G.~E. Hinton, and R.~J. Williams, ``Learning representations
  by back-propagating errors,'' \emph{Nature}, vol. 323, no. 6088, p. 533,
  1986.

\bibitem{cho2014learning}
K.~Cho, B.~Van~Merri{\"e}nboer, C.~Gulcehre, D.~Bahdanau, F.~Bougares,
  H.~Schwenk, and Y.~Bengio, ``Learning phrase representations using {RNN}
  encoder-decoder for statistical machine translation,'' \emph{Conf. Empir.
  Meth. Nat. Lang. Proc.}, 2014.

\bibitem{hochreiter1997long}
S.~Hochreiter and J.~Schmidhuber, ``Long short-term memory,'' \emph{Neural
  computation}, vol.~9, no.~8, pp. 1735--1780, 1997.

\bibitem{srivastava2015unsupervised}
N.~Srivastava, E.~Mansimov, and R.~Salakhudinov, ``Unsupervised learning of
  video representations using {LSTM}s,'' in \emph{Proc. Int. Conf. Mach.
  Learn}, 2015, pp. 843--852.

\bibitem{donahue2015long}
J.~Donahue, L.~Anne~Hendricks, S.~Guadarrama, M.~Rohrbach, S.~Venugopalan,
  K.~Saenko, and T.~Darrell, ``Long-term recurrent convolutional networks for
  visual recognition and description,'' in \emph{Proc. IEEE Conf. Comput. Vis.
  Pattern Recog.}, 2015, pp. 2625--2634.

\bibitem{wang2018capturing}
X.~Wang, R.~Jiang, L.~Li, Y.~Lin, X.~Zheng, and F.-Y. Wang, ``Capturing
  car-following behaviors by deep learning,'' \emph{IEEE Trans. Intell. Transp.
  Syst.}, vol.~19, no.~3, pp. 910--920.

\bibitem{xingjian2015convolutional}
S.~Xingjian, Z.~Chen, H.~Wang, D.-Y. Yeung, W.-K. Wong, and W.-c. Woo,
  ``Convolutional {LSTM} network: A machine learning approach for precipitation
  nowcasting,'' in \emph{Proc. Int. Conf. Adv. Neural Inf. Process. Syst.},
  2015, pp. 802--810.

\bibitem{tokmakov2017learning}
P.~Tokmakov, K.~Alahari, and C.~Schmid, ``Learning video object segmentation
  with visual memory,'' in \emph{Proc. IEEE Int. Conf. Comput. Vis.}, 2017, pp.
  4491--4500.

\bibitem{jie2018left}
Z.~Jie, P.~Wang, Y.~Ling, B.~Zhao, Y.~Wei, J.~Feng, and W.~Liu, ``Left-right
  comparative recurrent model for stereo matching,'' in \emph{Proc. IEEE Conf.
  Comput. Vis. Pattern Recog.}, 2018, pp. 3838--3846.

\bibitem{dai2017deformable}
J.~Dai, H.~Qi, Y.~Xiong, Y.~Li, G.~Zhang, H.~Hu, and Y.~Wei, ``Deformable
  convolutional networks,'' \emph{Proc. IEEE Int. Conf. Computer Vision}, 2017.

\bibitem{lang2012practical}
M.~Lang, O.~Wang, T.~Aydin, A.~Smolic, and M.~Gross, ``Practical temporal
  consistency for image-based graphics applications,'' \emph{ACM Trans. Graph},
  vol.~31, no.~4, p.~34, 2012.

\bibitem{aydin2014temporally}
T.~O. Aydin, N.~Stefanoski, S.~Croci, M.~Gross, and A.~Smolic, ``Temporally
  coherent local tone mapping of hdr video,'' \emph{ACM Trans. Graph}, vol.~33,
  no.~6, p. 196, 2014.

\bibitem{bonneel2013example}
N.~Bonneel, K.~Sunkavalli, S.~Paris, and H.~Pfister, ``Example-based video
  color grading.'' \emph{ACM Trans. Graph.}, vol.~32, no.~4, pp. 39--1, 2013.

\bibitem{lai2018learning}
W.-S. Lai, J.-B. Huang, O.~Wang, E.~Shechtman, E.~Yumer, and M.-H. Yang,
  ``Learning blind video temporal consistency,'' \emph{Proc. Eur. Conf. Comput.
  Vis.}, 2018.

\bibitem{gadde2017semantic}
R.~Gadde, V.~Jampani, and P.~V. Gehler, ``Semantic video {CNN}s through
  representation warping,'' \emph{Proc. IEEE Int. Conf. Comput. Vis.}, 2017.

\bibitem{wang2016temporal}
L.~Wang, Y.~Xiong, Z.~Wang, Y.~Qiao, D.~Lin, X.~Tang, and L.~Van~Gool,
  ``Temporal segment networks: Towards good practices for deep action
  recognition,'' in \emph{Proc. Eur. Conf. Comput. Vis.}, 2016, pp. 20--36.

\bibitem{yu2017dilated}
F.~Yu, V.~Koltun, and T.~A. Funkhouser, ``Dilated residual networks.'' in
  \emph{Proc. IEEE Conf. Comput. Vis. Pattern Recog.}, vol.~2, 2017, p.~3.

\bibitem{yu2015multi}
F.~Yu and V.~Koltun, ``Multi-scale context aggregation by dilated
  convolutions,'' \emph{Int. Conf. Learning Representations}, 2016.

\bibitem{he2016deep}
K.~He, X.~Zhang, S.~Ren, and J.~Sun, ``Deep residual learning for image
  recognition,'' in \emph{Proc. IEEE Conf. Comput. Vis. Pattern Recog.}, 2016,
  pp. 770--778.

\bibitem{jaderberg2015spatial}
M.~Jaderberg, K.~Simonyan, A.~Zisserman \emph{et~al.}, ``Spatial transformer
  networks,'' in \emph{Proc. Int. Conf. Adv. Neural Inf. Process. Syst.}, 2015,
  pp. 2017--2025.

\bibitem{zhao2017loss}
H.~Zhao, O.~Gallo, I.~Frosio, and J.~Kautz, ``Loss functions for image
  restoration with neural networks,'' \emph{IEEE Trans. Comput. Imaging},
  vol.~3, no.~1, pp. 47--57, 2017.

\bibitem{garg2016unsupervised}
R.~Garg, V.~K. BG, G.~Carneiro, and I.~Reid, ``Unsupervised {CNN} for single
  view depth estimation: Geometry to the rescue,'' in \emph{Proc. Eur. Conf.
  Comput. Vis.}, 2016, pp. 740--756.

\bibitem{Cordts2016Cityscapes}
M.~Cordts, M.~Omran, S.~Ramos, T.~Rehfeld, M.~Enzweiler, R.~Benenson,
  U.~Franke, S.~Roth, and B.~Schiele, ``The {C}ityscapes dataset for semantic
  urban scene understanding,'' in \emph{Proc. IEEE Conf. Comput. Vis. Pattern
  Recog.}, 2016.

\bibitem{kingma2014adam}
D.~P. Kingma and J.~Ba, ``Adam: A method for stochastic optimization,''
  \emph{Int. Conf. Learning Representations}, 2015.

\bibitem{wang2018occlusion}
Y.~Wang, Y.~Yang, Z.~Yang, L.~Zhao, and W.~Xu, ``Occlusion aware unsupervised
  learning of optical flow,'' in \emph{Proc. IEEE Conf. Comput. Vis. Pattern
  Recog.}, 2018, pp. 4884--4893.

\bibitem{kroeger2016fast}
T.~Kroeger, R.~Timofte, D.~Dai, and L.~Van~Gool, ``Fast optical flow using
  dense inverse search,'' in \emph{Proc. Eur. Conf. Comput. Vis.}, 2016, pp.
  471--488.

\bibitem{abadi2016tensorflow}
M.~Abadi, P.~Barham, J.~Chen, Z.~Chen, A.~Davis, J.~Dean, M.~Devin,
  S.~Ghemawat, G.~Irving, M.~Isard \emph{et~al.}, ``Tensorflow: a system for
  large-scale machine learning.'' in \emph{Proc. USENIX Symp. Oper. Syst. Des.
  Implement.}, vol.~16, 2016, pp. 265--283.

\bibitem{krizhevsky2012imagenet}
A.~Krizhevsky, I.~Sutskever, and G.~E. Hinton, ``Image{N}et classification with
  deep convolutional neural networks,'' in \emph{Proc. Int. Conf. Adv. Neural
  Inf. Process. Syst.}, 2012, pp. 1097--1105.

\bibitem{mayer2016large}
N.~Mayer, E.~Ilg, P.~Hausser, P.~Fischer, D.~Cremers, A.~Dosovitskiy, and
  T.~Brox, ``A large dataset to train convolutional networks for disparity,
  optical flow, and scene flow estimation,'' in \emph{Proc. IEEE Conf. Comput.
  Vis. Pattern Recog.}, 2016, pp. 4040--4048.

\bibitem{sun2018pwc}
D.~Sun, X.~Yang, M.-Y. Liu, and J.~Kautz, ``{PWC}-{N}et: {CNN}s for optical
  flow using pyramid, warping, and cost volume,'' in \emph{Proc. IEEE Conf.
  Comput. Vis. Pattern Recog.}, 2018, pp. 8934--8943.

\bibitem{deng2009imagenet}
J.~Deng, W.~Dong, R.~Socher, L.-J. Li, K.~Li, and L.~Fei-Fei, ``Image{N}et: A
  large-scale hierarchical image database,'' in \emph{Proc. IEEE Conf. Comput.
  Vis. Pattern Recog.}, 2009, pp. 248--255.

\bibitem{Silberman:ECCV12}
P.~K. Nathan~Silberman, Derek~Hoiem and R.~Fergus, ``Indoor segmentation and
  support inference from r{G}{B}{D} images,'' in \emph{Proc. Eur. Conf. Comput.
  Vis.}, 2012.

\end{thebibliography}

\begin{IEEEbiography}
[{\includegraphics[width=1in,height=1.25in,clip]{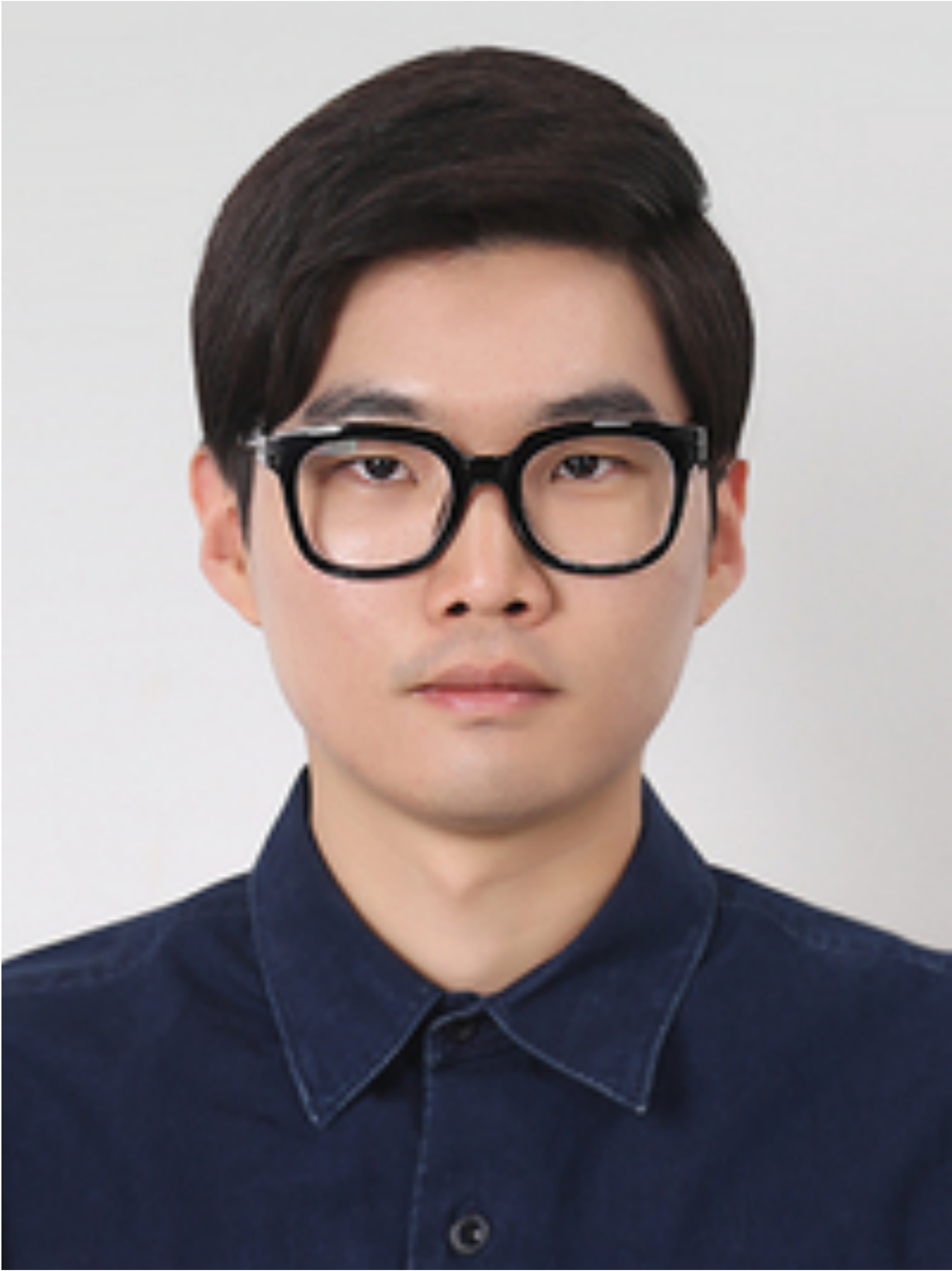}}]
{Chanho Eom} received the B.S. degree in electrical and electronic engineering from Yonsei University, Seoul, South Korea, in 2017, where he is currently pursuing the joint M.S. and Ph.D. degrees in electrical and electronic engineering. His current research interests include computer vision and machine learning both in theory and applications.
\end{IEEEbiography}
\vspace{-0.7cm}
\begin{IEEEbiography}
[{\includegraphics[width=1in,height=1.25in,clip]{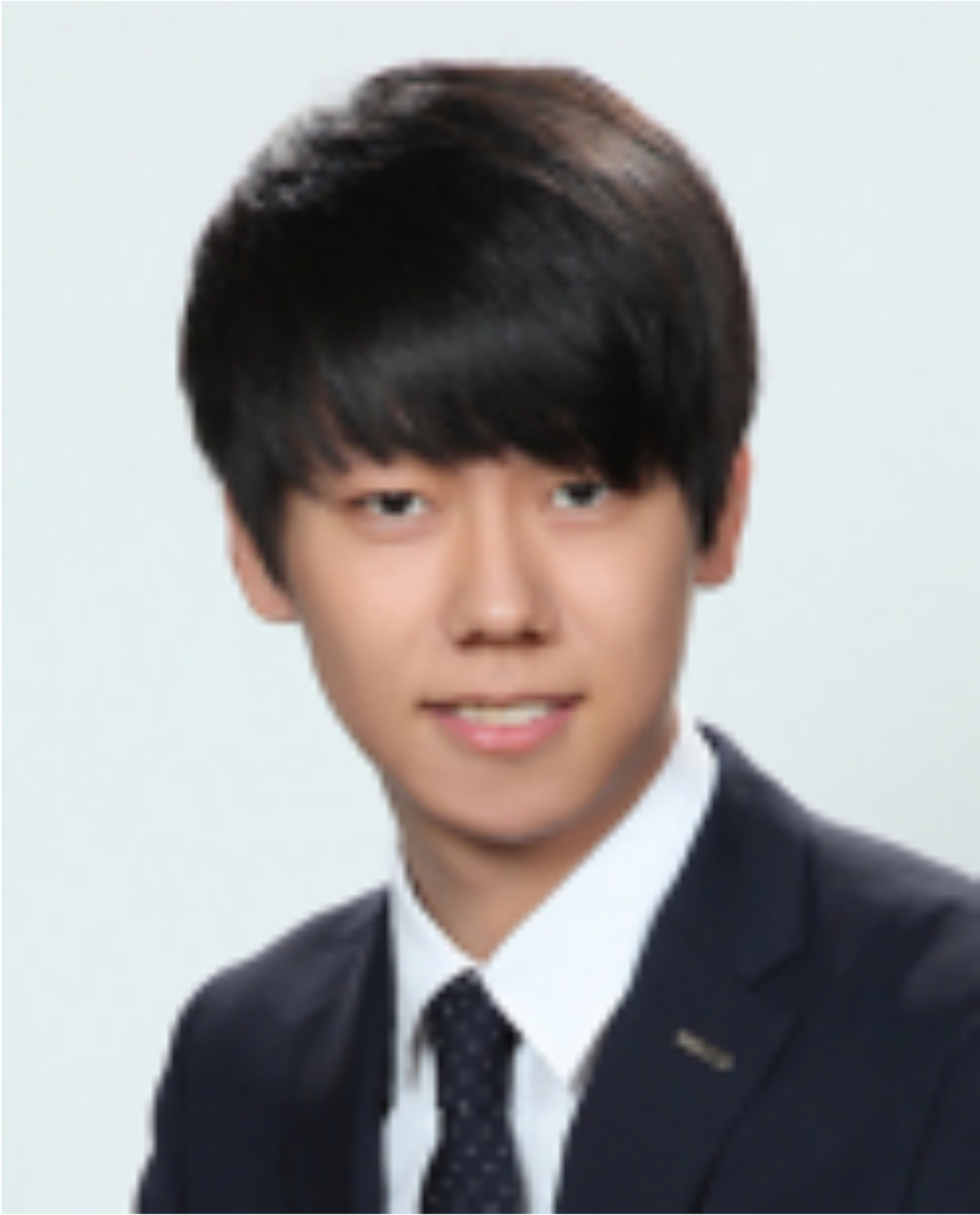}}]
{Hyunjong Park} received the B.S. degree in electrical and electronic engineering from Yonsei University, Seoul, South Korea, in 2017, where he is currently pursuing the joint M.S. and Ph.D. degrees in electrical and electronic engineering. His current research interests include optical flow and depth prediction to solve computer vision tasks.
\end{IEEEbiography}
\vspace{-0.7cm}
\begin{IEEEbiography}
[{\includegraphics[width=1in,height=1.25in,clip]{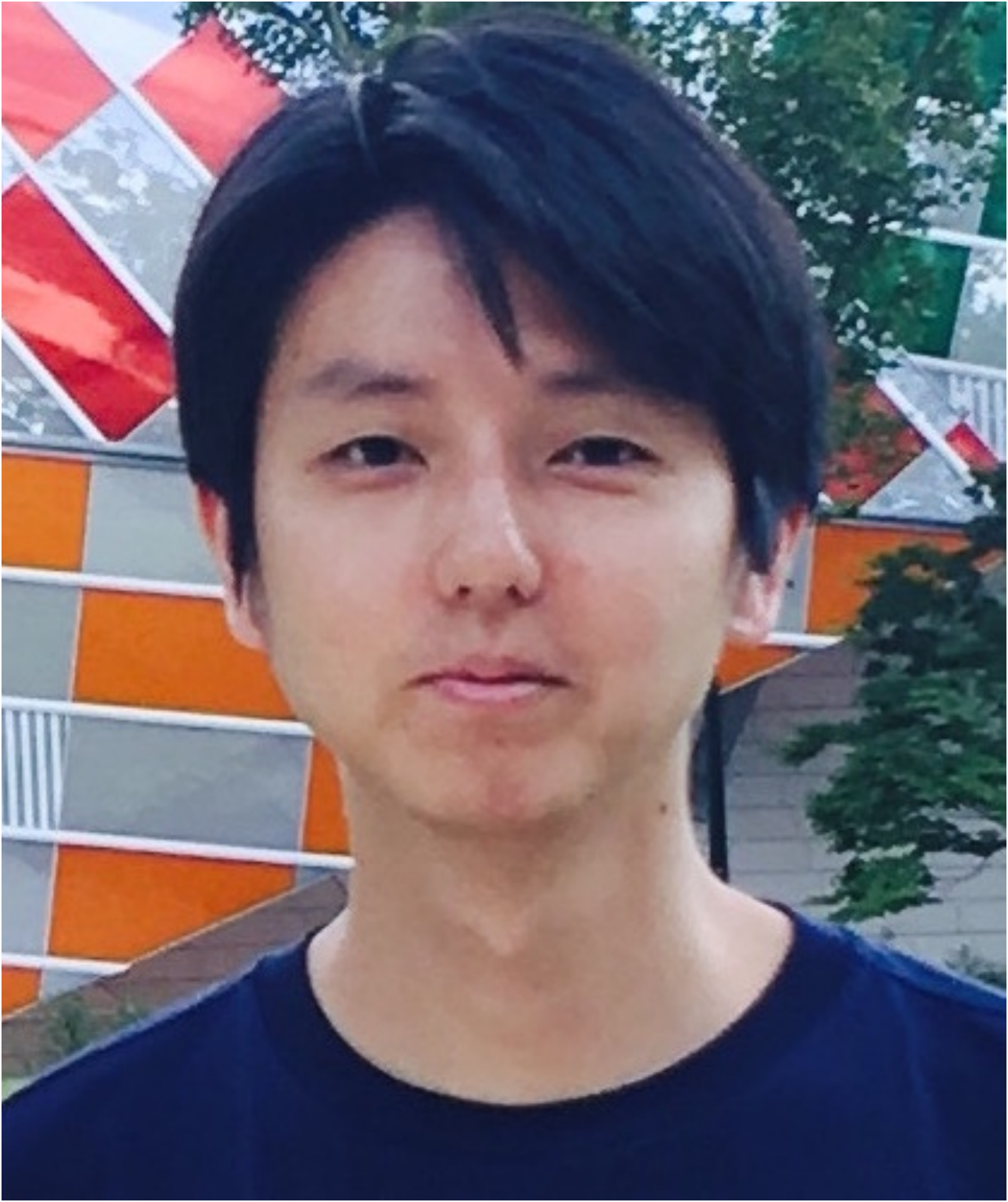}}]
{Bumsub Ham} is an an Assistant Professor of Electrical and Electronic Engineering at Yonsei University in Seoul, Korea. He received the B.S. and Ph.D. degrees in Electrical and Electronic Engineering from Yonsei University in 2008 and 2013, respectively. From 2014 to 2016, he was Post-Doctoral Research Fellow with Willow Team of INRIA Rocquencourt, {\'E}cole Normale Sup{\'e}rieure de Paris, and Centre National de la Recherche Scientifique. His research interests include computer vision, computational photography, and machine learning, in particular, regularization and  matching, both in theory and applications.
\end{IEEEbiography}

\end{document}